\documentclass{article}

    \PassOptionsToPackage{numbers, compress}{natbib}

    \usepackage[final]{neurips_2025}

\usepackage{graphicx}
\usepackage{multirow}
\usepackage{amsmath}
\usepackage{array} 
\usepackage{bbm}
\usepackage{bm}
\usepackage{threeparttable}
\usepackage{pdfpages}
\usepackage{geometry}
\usepackage{lipsum}
\usepackage{float}
\usepackage{subcaption}
\usepackage{xcolor}
\usepackage{colortbl} %

\usepackage[utf8]{inputenc} %
\usepackage[T1]{fontenc}    %
\usepackage{hyperref}       %
\usepackage{url}            %
\usepackage{booktabs}       %
\usepackage{amsfonts}       %
\usepackage{nicefrac}       %
\usepackage{microtype}      %
\usepackage{url}

\title{ Orientation Matters: Making 3D Generative Models Orientation-Aligned}

\newcommand{\bI}{\mathbf{I}}

\newcommand{\cN}{\mathcal{N}}

\makeatletter
\DeclareRobustCommand\onedot{\futurelet\@let@token\@onedot}
\def\@onedot{\ifx\@let@token.\else.\null\fi\xspace}

\makeatother

\newcommand{\figref}[1]{Figure~\ref{#1}}
\newcommand{\secref}[1]{Section~\ref{#1}}

\renewcommand{\eqref}[1]{Eq.~\ref{#1}}
\newcommand{\tabref}[1]{Table~\ref{#1}}

\newcommand{\boldparagraph}[1]{\vspace{0.2cm}\noindent{\bf #1:} }

\newif\ifcomment
\commenttrue
\ifcomment
	\newcommand{\ag}[1]{ \noindent {\color{red} {\bf Andreas:} {#1}} }
	\newcommand{\yl}[1]{ \noindent {\color{cyan} {\bf YL:} {#1}} }

\else
	\newcommand{\ag}[1]{}
	\newcommand{\yl}[1]{}
\fi

\newcolumntype{P}[1]{>{\centering\arraybackslash}m{#1}}

\newcommand{\tabred}{\cellcolor[HTML]{F47983}}
\newcommand{\taborange}{\cellcolor[HTML]{FFC773}}
\newcommand{\tabyellow}{\cellcolor[HTML]{faff72}}

\makeatletter
\def\blfootnote{\gdef\@thefnmark{}\@footnotetext}
\makeatother

\begin{document}

\author{
  \textnormal{Yichong Lu}$^{1,2\text{*}}$
  \and
  Yuzhuo Tian$^{1\text{*}}$
  \and
  Zijin Jiang$^{1\text{*}}$
  \and
  Yikun Zhao$^{1}$
  \and
  Yuanbo Yang$^{1,2}$
  \and
  Hao Ouyang$^{2}$
  \and
  Haoji Hu$^{1}$
  \and
  Huimin Yu$^{1}$
  \and
  Yujun Shen$^{2}$
  \and
  Yiyi Liao\text{$^{1\text{†}}$}\\
  \vspace{-1em}
  \and
  $^{1}$Zhejiang University 
  \quad 
  $^{2}$Ant Group \\
   \\
  \href{https://xdimlab.github.io/Orientation_Matters/}{\textcolor[rgb]{0,0,1}{\text{https://xdimlab.github.io/Orientation\_Matters}}}
}

\blfootnote{$^\text{*}$Equal contribution. $^\text{†}$Corresponding author.}

\maketitle

\begin{abstract} \label{abstract}

Humans intuitively perceive object shape and orientation from a single image, guided by strong priors about canonical poses. However, existing 3D generative models often produce misaligned results due to inconsistent training data, limiting their usability in downstream tasks. To address this gap, we introduce the task of orientation-aligned 3D object generation: producing 3D objects from single images with consistent orientations across categories. 
To facilitate this, we construct Objaverse-OA, a dataset of 14,832 orientation-aligned 3D models spanning 1,008 categories. 
Leveraging Objaverse-OA, we fine-tune two representative 3D generative models based on multi-view diffusion and 3D variational autoencoder frameworks to produce aligned objects that generalize well to unseen objects across various categories. Experimental results demonstrate the superiority of our method over post-hoc alignment approaches. Furthermore, we showcase downstream applications enabled by our aligned object generation, including zero-shot object orientation estimation via analysis-by-synthesis and efficient arrow-based object rotation manipulation.

\end{abstract}
    
\section{Introduction} \label{introduction}

Humans possess a remarkable ability to imagine the 3D structure of an object (``synthesis'') and infer its properties, such as orientation (``analysis''), from a single image.
We intuitively recognize the correct orientation of a car on the road, know how to grasp a cup by its handle, or decide how to place a chair in a room.
In cognitive science, this ability is linked to the concept of \textit{object constancy}—the capacity to mentally reconstruct a canonicalized object despite changes in viewpoint or other variations~\cite{Murray2002ShapePR, Lawson1999AchievingVO}.
This synthesis of an internal object representation enables analysis of its pose, a process often described as analysis by synthesis.

Replicating this perceptual capability has long been a central pursuit in computer vision. Recent progress in generative AI has significantly advanced single-view 3D object generation, making it both accessible and effective. This advancement in ``synthesis'' offers a rapid and adaptable means of constructing 3D models from everyday images, enabling a wide range of applications in AR/VR content creation, robotic simulation, etc.

Despite these strides, a critical challenge remains: \textit{orientation alignment}. Existing methods often neglect this aspect, largely due to inconsistencies in large-scale 3D datasets~\cite{Deitke2022ObjaverseAU, Xiang2024Structured3L}. As a result, generated models frequently lack standardized canonical orientations—chairs may face different directions, mugs may appear tipped over, and vehicles may be misaligned. 
This disparity between human perceptual consistency and the output of generative models poses a significant obstacle for downstream tasks such as analysis-by-synthesis, e.g., orientation estimation.
To bridge this perceptual gap, it is essential that 3D generative models not only reconstruct object geometry but also generate objects in consistent, semantically meaningful orientations.

In this paper, we introduce a novel task: orientation-aligned 3D object generation—the generation of a 3D object from a single image such that its orientation is consistently aligned both within and across categories, in accordance with common-sense priors. 
While no existing methods explicitly address this task, a possible workaround involves first generating a 3D object and then applying pose estimation to align its orientation. However, achieving robust and generalizable orientation alignment across diverse object categories remains a significant challenge.
Most 3D pose estimation approaches~\cite{Wen2023FoundationPoseU6, Peng2018PVNetPV, Sun2022OnePoseOO} focus on predicting \textbf{relative} poses with respect to predefined 3D CAD models or multi-view reference images.  
In contrast, our task demands \textbf{absolute} orientation prediction — aligning objects to a canonical coordinate frame that corresponds to intuitive understandings of front, top, and upright directions.
While some prior works explore absolute pose estimation~\cite{Wang2019NormalizedOC, Chen2020CategoryLO}, they are typically constrained to a limited set of categories, primarily due to the substantial manual effort required to curate orientation-aligned training data. Recent efforts to scale such methods have relied either on labor-intensive human annotations~\cite{Krishnan2024OmniNOCSAU, Ma2024ImageNet3DTG} or on the use of Vision-Language Models (VLMs)~\cite{Wang2024OrientAL}. However, the former approach remains category-restricted, while the latter is subject to the inherent inaccuracies of VLM-based orientation predictions.
Moreover, these strategies necessitate complex post-processing after 3D generation, which not only introduces potential inaccuracies but also reduces user-friendliness and practical applicability.

\begin{figure}[t]
    \centering
    \includegraphics[width=\linewidth]{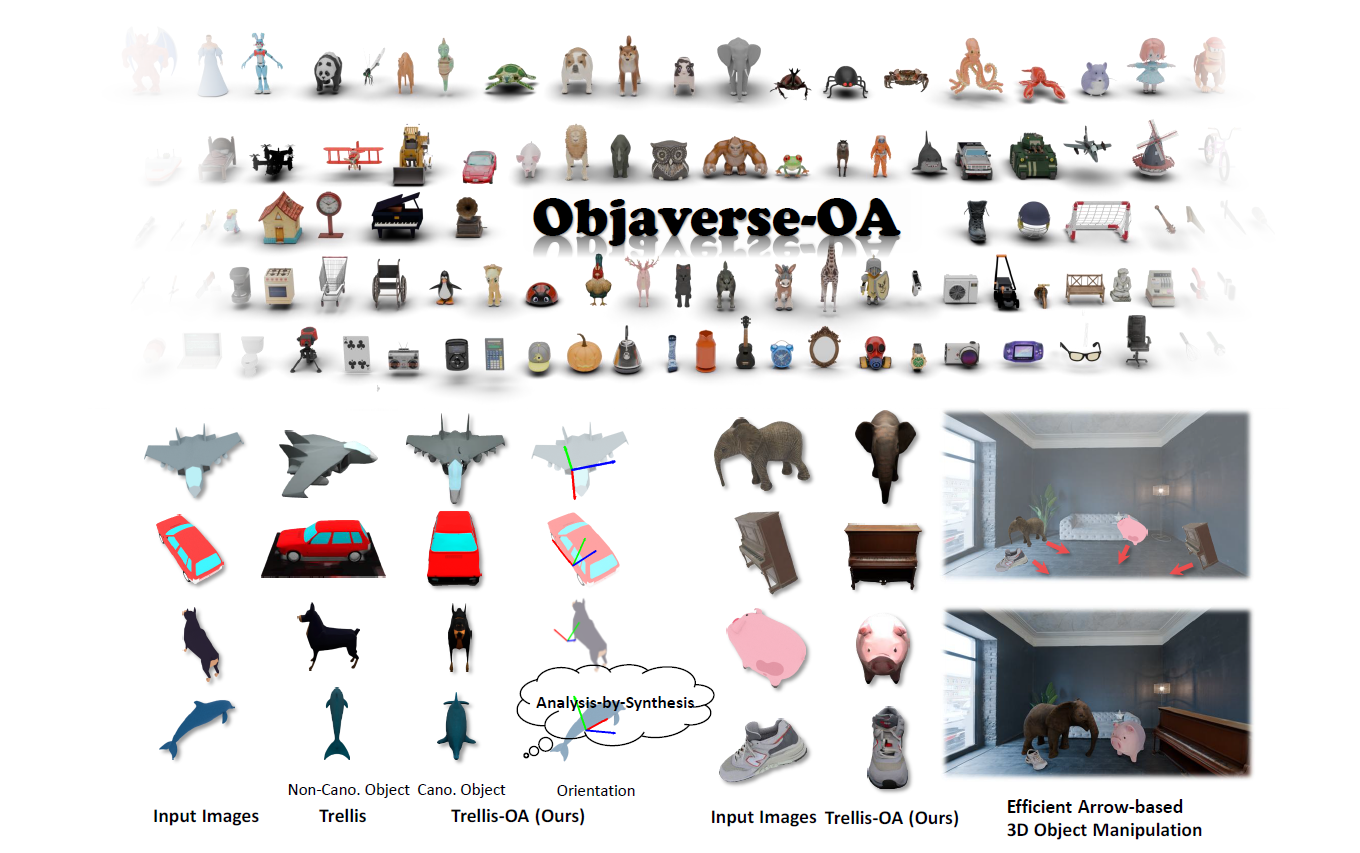} \\
    \vspace{-1em}
    \caption{\textbf{Objaverse-OA for Orientation-Aligned Generation}. We construct a new dataset named Objaverse-OA, which contains orientation-aligned 3D models across 1008 categories (top). Using Objaverse-OA, we make existing 3D generative models orientation-aligned, which can further be used for zero-shot model-free orientation estimation (bottom left) and efficient arrow-based 3D object rotation manipulation (bottom right).} 
    \label{fig:teaser}
    
\end{figure}

To address this, we propose to learn canonical 3D object generation by directly fine-tuning pre-trained 3D generative models, thereby avoiding the limitations of such two-stage pipelines. Our key insight is that a sufficiently diverse set of orientation-aligned 3D models can effectively adapt existing generative models from producing arbitrarily oriented outputs to generating objects with consistent, canonical orientations, while maintaining strong generalization capabilities to unseen objects across various categories. To support this, we introduce Objaverse-OA, a new dataset comprising 14,832 3D models spanning 1,008 categories, each aligned to a consistent, common-sense orientation. Leveraging Objaverse-OA, we fine-tune two representative 3D generative models~\cite{Xiang2024Structured3L, Long2023Wonder3DSI} to produce Trellis-OA and Wonder3D-OA, allowing for generating well-aligned 3D objects across a broad spectrum of categories, including those not included in the fine-tuning set.

We further demonstrate the utility of orientation-aligned 3D generative models through two downstream applications: zero-shot 3D object orientation estimation and efficient arrow-based object rotation manipulation. For orientation estimation, our canonically generated 3D models serve as templates to estimate object poses from single images, generalizing well across categories. Moreover, we develop a user-friendly interface for object rotation manipulation in the augmented reality applications and 3D software, allowing users to specify the desired orientation via drawing an arrow, thereby facilitating precise placement without tedious pose adjustments.

Our contributions are as follows: 
1) We introduce the novel task of orientation-aligned 3D object generation across a wide range of categories.
2) We construct Objaverse-OA, the largest orientation-aligned 3D dataset in terms of category coverage.
3) We fine-tune existing 3D generative models on Objaverse-OA to enable canonical object generation with robust generalization to unseen objects across various categories. Experimental results across multiple datasets demonstrate that our method achieves superior orientation alignment compared to existing baselines. 
4) We showcase the practical benefits of our orientation-aligned models in two key applications: zero-shot orientation estimation and efficient object rotation manipulation via intuitive user interaction.

\section{Related Work}

\noindent \textbf{3D Generative Models: }
Early 3D generation methods~\cite{Wu2016LearningAP, Schwarz2020GRAFGR, Chan2021EfficientG3, Pavllo2022ShapePA, Chen2024Learning3G} typically employed GANs~\cite{Treat2018GENERATIVEAN} to model 3D distribution, while these methods generate orientation-aligned objects, they are limited to a single category.
The recent breakthroughs in 2D diffusion models~\cite{Croitoru2022DiffusionMI, Ho2020DenoisingDP} provide new solutions for 3D generation. Pioneering works DreamFusion~\cite{Poole2022DreamFusionTU} and SJC~\cite{Wang2022ScoreJC} propose to generate 3D models by distilling from a 2D text-to-image generation model. However, these methods and their follow-ups~\cite {Liang2023LucidDreamerTH, Lin2022Magic3DHT, Tang2023MakeIt3DH3, Tang2023DreamGaussianGG, Wang2023ProlificDreamerHA} always suffer from low efficiency and multi-face problems due to per-shape optimization and lack of explicit 3D supervision. Recently, methods based on multi-view diffusion models~\cite{Long2023Wonder3DSI, Liu2023SyncDreamerGM, Wang2023ImageDreamIM, Shi2023Zero123AS, Szymanowicz2023ViewsetD, Shi2023MVDreamMD,Tang2024LGMLM, Xu2024GRMLG} have succeeded in efficiently producing multi-view consistent images via 3D attention. More recently, \cite{Meng2024LT3SDLT, Xiang2024Structured3L, Zhao2025Hunyuan3D2S, Zhang2024CLAYAC, Xu2023DMV3DDM} utilize 3D latent spaces to further improve the geometry quality of the 2D-assisted approaches. However, despite huge progress on quality and efficiency, they all produce 3D models with uncanonical orientations due to orientation misalignment in their 3D training data, like Objaverse~\cite{Deitke2022ObjaverseAU}, Objaverse-XL~\cite{Deitke2023ObjaverseXLAU}, and TRELLIS-500K~\cite{Xiang2024Structured3L}.

\noindent \textbf{Object Orientation Estimation: }
One feasible approach to align 3D model orientations is to render the 3D models from a fixed camera and estimate the object orientations in the renderings. Although image-based 3D object pose estimation has been widely researched, most methods~\cite{Wen2023FoundationPoseU6, Peng2018PVNetPV, Sun2022OnePoseOO, Lee2025Any6DM6, Liu2025HIPPoHI, Yu2025BoxDreamerDB} focus on predicting relative poses based on known 3D CAD models or reference images. Since the orientation-aligned 3D CAD models and reference images are not available, they cannot estimate the object orientations aligned with common sense. Category-level object pose estimation methods~\cite{Wang2019NormalizedOC, Chen2020CategoryLO, Wei2023RGBbasedCO, Lin2022CategoryLevel6O, Fan2022ObjectLD, Tian2020ShapePD} address this problem by generating 3D shape priors from the input images. However, most of them are limited to category-level due to heavy labor effort in constructing orientation-aligned 3D datasets~\cite{Wang2019NormalizedOC, Chang2015ShapeNetAI, Krishnan2024OmniNOCSAU, Ma2024ImageNet3DTG}. In contrast, our method can generate orientation-aligned 3D objects across a large number of categories. Concurrently, Orient Anything~\cite{Wang2024OrientAL} realizes zero-shot object orientation estimation by automatically constructing large orientation-aligned 3D datasets using advanced Vision Language Models (VLMs). However, it still suffers from generalizability and accuracy due to a lack of training on the object orientation estimation task. Besides, post-processing after 3D generation is costly compared to directly generating orientation-aligned 3D models. Concurrently, ~\cite{Jin2025Oneshot3O} proposes an intra-category object pose canonicalization method and constructs Canonical Objaverse Dataset, which contains 3D objects with canonical poses within categories. However, our work focuses on inter-category object pose canonicalization.

\section{Objaverse-OA Dataset} \label{sec:objaverse-oa}

In this section, we introduce the construction of our dataset, Objaverse-OA. Dataset diversity plays a crucial role in achieving strong generalization capability. 
To the best of our knowledge, the existing orientation-aligned 3D dataset ~\cite{Ma2024ImageNet3DTG} with the largest category number includes only 200 categories and fewer than 2,000 3D objects. In contrast, our Objaverse-OA dataset contains 14,832 orientation-aligned 3D objects across 1008 categories, which will be made publicly available to the research community. To build this large-scale dataset while maintaining both efficiency and accuracy, we employ a hybrid pipeline that combines Vision-Language Model (VLM) pre-processing with manual correction.

\begin{figure}
    \centering
    \includegraphics[width=1\linewidth]{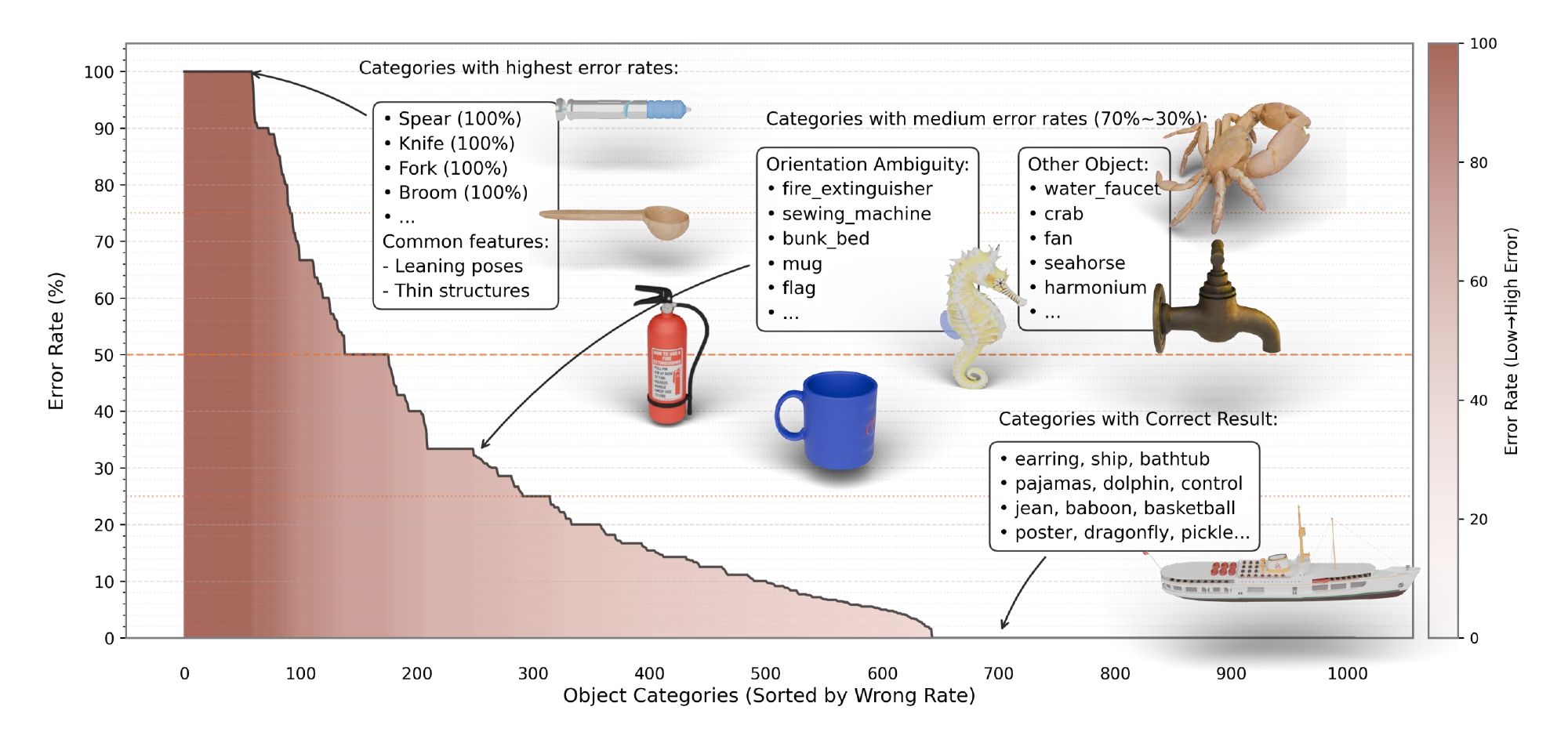}
    \caption{\textbf{VLM's Performance in Orientation Estimation}. We utilize our manually curated dataset as ground truth (GT) and show the error rate of VLM's estimation across different categories.
    We observe that (1) the VLM demonstrates particular difficulty in recognizing front-facing orientations for stick-like objects, and (2) a significant portion of recognition errors occur when processing objects with inherently unclear or ambiguous frontal views. These challenges highlight the necessity of our manual curation.}
    \label{fig:dataset_oa}
\end{figure}

\noindent \textbf{VLM pre-processing: }
As discovered by Orient Anything~\cite{Wang2024OrientAL}, advanced VLMs demonstrate the ability to recognize object front views without task-specific training. 
Since most models in Objaverse primarily vary in the horizontal (yaw) axis, we follow the strategy proposed in Orient Anything: we render each 3D object from four horizontal viewpoints—front, back, left, and right—and use a VLM to identify the correct front view. Based on the identified view, we then rotate the 3D model accordingly to align it to a canonical orientation. Our data processing begins with the Objaverse-LVIS dataset, and we use Gemini-2.0~\cite{gemini} as the VLM for view recognition. From a total of 46,219 3D models, Gemini successfully identifies front views for 20,664 objects. However, we observe that VLM-based recognition, while promising, still falls short of human-level accuracy due to the absence of fine-tuning and challenges associated with ambiguous or difficult cases. 
We illustrate the error rate of VLM's prediction across different categories in Fig. \ref{fig:dataset_oa}.
One of the challenging cases is stick-like objects, like spears, keys, and forks, since many of them are not aligned in roll and pitch angles in the Objaverse. 
Another challenge involves geometrically narrow or thin objects such as fish, bicycles, and water faucets.
In such cases, VLM struggles to identify the correct orientation, as it relies solely on visible front-view features without reasoning from side-view context, unlike humans.
What's more, objects with ambiguous front views, like teapots, extinguishers, and cups, can result in inconsistent orientation predictions. To address these issues, we introduce a manual correction step described below.

\noindent \textbf{Manual correction: }
Starting from the VLM-based alignment, we manually filter or correct the wrong recognition results of the VLM and canonicalize objects with the ambiguous front-view definition. Moreover, to preserve category diversity, we reintroduce objects that were incorrectly filtered out by the VLM, particularly in cases where a category has too few correctly aligned instances. As illustrated in the \figref{fig:dataset_oa}, we manually correct the orientations for about 600 object categories, especially the stick-like objects and objects with ambiguous orientation or unclear front view features. Note that some objects have orientation ambiguity, especially tools like spoons, mugs, and fire extinguishers. During the manual correction process, we refer to the object orientations defined in prior work, specifically ImageNet3D~\cite{Ma2024ImageNet3DTG}, for the categories it covers. For example, for spoons, which are included in ImageNet3D, we align their poses in our dataset accordingly. For ambiguous objects only included in our dataset, we define canonical poses based on semantic part structures, geometric features, and common knowledge, following the principles established by ImageNet3D and our supplementary material ~\secref{sec:dataset_curation}. What's more, we also filter objects with low geometry quality and scenes with multiple objects. Experiments show that the pre-trained 3D generative models can further improve geometry quality after fine-tuning on our dataset.

\section{Orientation-Aligned 3D Object Generation}

Based on our curated Objaverse-OA (\secref{sec:objaverse-oa}), we fine-tune existing 3D generative models to generate orientation-aligned objects. 
To demonstrate that orientation-aligned object generation benefits a variety of architectural backbones, we implement our approach on two widely used single-view image-to-3D reconstruction frameworks: a 3D VAE-based generative model (see \secref{sec:3d-vae}) and a multi-view diffusion model (see \secref{sec:mv-diffusion}).

\subsection{3D-VAE Based Generative Model} \label{sec:3d-vae}

We choose a state-of-the-art 3D-VAE-based generative model, Trellis~\cite{Xiang2024Structured3L}, as our base model, which can produce fine-grained geometry and appearance aligned with the input image.

\begin{figure}[t]
    \centering
    \includegraphics[width=\linewidth]{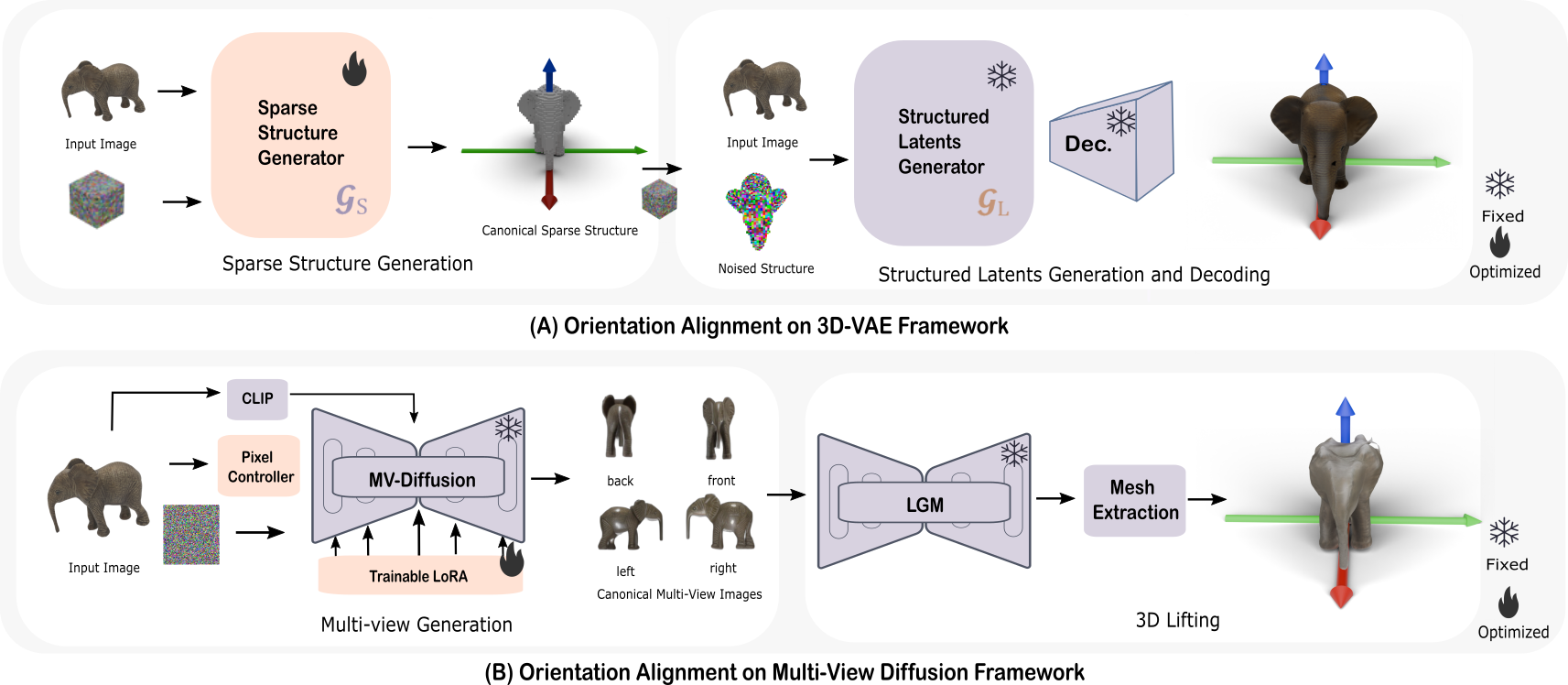}
    \caption{\textbf{Trellis-OA and Wonder3D-OA}.We fine-tune two representative methods: Trellis~\cite{Xiang2024Structured3L}, based on a 3D-VAE backbone (top), and Wonder3D~\cite{Long2023Wonder3DSI}, based on a multi-view diffusion backbone (bottom). For the 3D-VAE, we find that fine-tuning only the sparse structure generator is sufficient to produce orientation-aligned objects. For the multi-view diffusion model, we adopt LoRA as a lightweight domain adapter to enable the generation of orientation-aligned target images.}
    \vspace{-1em}
    \label{fig:pipline}
\end{figure}

\noindent \textbf{Preliminary: }As shown in the \figref{fig:pipline}, Trellis~\cite{Xiang2024Structured3L} adopts three modules for 3D asset generation during inference including sparse structure generator $\bm{\mathcal{G}}_{\mathrm{S}}$, structured latents generator $\bm{\mathcal{G}}_{\mathrm{L}}$, and 3D decoder $\bm{\mathcal{D}}$. Given the input image $\bI$, sparse structure generator $\bm{\mathcal{G}}_{\mathrm{S}}$ produces dense binary 3D grid $\boldsymbol{O} \in\{0,1\}^{N \times N \times N}$: $\boldsymbol{O} = \bm{\mathcal{G}}_{\mathrm{S}} (\bI, \bf{\bm{\varepsilon}_{3d}})$, where N is the length of the grid and $\bm{\varepsilon}_{3d}$ is the 3D noise sampled from $\cN$(0,1), which is further converted into active voxels $\{(\boldsymbol{\mathit{p}}_i)\}_{i=1}^N$ defined as sparse structure. After that, sparse latents generator $\bm{\mathcal{G}}_{\mathrm{S}}$ is used to generate structured latents $\boldsymbol{\mathit{z}} = \{(\boldsymbol{\mathit{z}}_i, \boldsymbol{\mathit{p}}_i)\}_{i=1}^N$: $\boldsymbol{\mathit{z}} = \bm{\mathcal{G}}_{\mathrm{L}} (\bI, \boldsymbol{\mathit{z}}_{noised})$, where $\boldsymbol{\mathit{z}}_{noised}$ is the noised sparse structure $\{(\boldsymbol{\bm{\varepsilon}}_i, \boldsymbol{\mathit{p}}_i)\}_{i=1}^N$. Finally, the 3D representation $\bm{\mathcal{M}}$ is obtained via 3D decoding: $\bm{\mathcal{M}} = \bm{\mathcal{D}}(\boldsymbol{\mathit{z}})$.

\noindent \textbf{Trellis-OA: }The original Trellis model~\cite{Xiang2024Structured3L} is unable to produce orientation-aligned 3D outputs due to the orientation inconsistencies present in its training data. To address this, we fine-tune Trellis using our Objaverse-OA dataset, resulting in Trellis-OA, which generates 3D objects with aligned orientations. Although Trellis comprises several modules for 3D generation, we find that fine-tuning only the sparse structure generator $\bm{\mathcal{G}}{\mathrm{S}}$ is sufficient for achieving orientation alignment. This is likely because Trellis inherently generates object poses randomly sampled from four orthogonal directions, and our aligned pose distribution resides within this range. As a result, both the pre-trained structured latent generator $\bm{\mathcal{G}}{\mathrm{L}}$ and the 3D decoder $\bm{\mathcal{D}}$ remain compatible with the aligned orientations and do not require additional fine-tuning. Specifically, our fine-tuned sparse structure generator $\bm{\mathcal{G}}_{\mathrm{S}}^{\prime}$ generates canonical sparse structure $\{(\boldsymbol{\mathit{p}}_i^{\prime})\}_{i=1}^N$, which shares aligned orientations. Afterwards, pre-trained structured latents generator $\bm{\mathcal{G}}_{\mathrm{L}}$ generates canonical structured latents and 3D decoder $\bm{\mathcal{D}}$ produces final canonical 3D models $\bm{\mathcal{M}}_{OA}$with orientation aligned.  
Our experiments demonstrate that we can efficiently fine-tune Trellis to an orientation-aligned one while preserving its 3D priors.

\subsection{Multi-view Diffusion Model} \label{sec:mv-diffusion}

We choose Wonder3D~\cite{Long2023Wonder3DSI} as our base multi-view diffusion model since it is one of the most representative works based on multi-view diffusion frameworks. Note that the recipe for Wonder3D can also be used by other multi-view diffusion based methods.

\noindent \textbf{Preliminary: }Multi-view diffusion models are typically fine-tuned from large-scale pre-trained text-to-image models~\cite{Rombach2021HighResolutionIS} and achieve multi-view consistent via 3D-aware attention mechanisms. Specifically, given an input image $\bI$, the multi-view diffusion model $\boldsymbol{\mathit{MV}}$ generates $N$ multi-view images $\{\bI_{mv}^{i}\}_{i=1}^M = \boldsymbol{\mathit{MV}}(\bI, \bm{\varepsilon})$, where $\bm{\varepsilon}\in \cN(0,1)$ is a sampled noise vector. Here, the poses of the ground truth of the multi-view images $\{\bI_{mv}^{i}\}_{i=1}^N$ are dependent on the input image $\bI$, where $\bI_{mv}^1$ is set to predict the input image $\bI$. After that, Wonder3D adopts an optimization method based on NeuS~\cite{Wang2021NeuSLN} to lift $\{\bI_{mv}\}$ to 3D representation $\bm{\mathcal{M}}$.

\noindent \textbf{Wonder3D-OA:} Wonder3D~\cite{Long2023Wonder3DSI} cannot produce orientation-aligned 3D models due to its misaligned training data and input-image related camera settings. To address this, we implement Wonder3D-OA, leveraging our orientation-aligned Objaverse-OA dataset. Given an input image $\bI$, Wonder3D-OA first generates orientation-aligned, multi-view consistent images $\bI_{OA}^{mv}$, which are then lifted to orientation-aligned 3D representations $\bm{\mathcal{M}}_{OA}$.
Our key change is in the supervision setup: instead of using input-view-dependent ground truth, we render six canonical views (front, front-left, front-right, left, right, back) from fixed camera poses based on Objaverse-OA. This provides consistent orientation references. We adopt LoRA~\cite{Hu2021LoRALA} as a lightweight domain adapter to fine-tune the pre-trained Wonder3D, preserving its learned 3D priors from non-canonical data while enabling alignment. Additionally, we make architectural adjustments for improved performance. 
Originally, Wonder3D aligns the input image with one of the predicted views to inject local features, which fails to work in our canonical camera setting. 
Inspired by ImageDream~\cite{Wang2023ImageDreamIM}, we instead employ a pixel injector to integrate local features into the multi-view diffusion model.  Specifically, Wonder3D employs a 3D dense self-attention mechanism with a shape of ($b_z$, 6, c, $h_l$, $w_l$) across six views within a transformer layer, where $b_z$ is the batch size, c is the number of feature channel, $h_l$ and $w_l$ are the image resolution. Our pixel controller modifies this to ($b_z$, 7, c, $h_l$, $w_l$), incorporating the input image as an additional view.
We further improve efficiency by replacing Wonder3D’s test-time optimization-based 3D lifting module (NeuS) with LGM~\cite{Tang2024LGMLM}, a recent model tailored for sparse-view 3D reconstruction. Since LGM is trained using 4-views as input, we feed the front, left, right, and back views into LGM for 3D lifting.
Finally, as our lifting module performs effectively without the need for normal maps, we omit fine-tuning the cross-domain attention module in Wonder3D, simplifying the pipeline without sacrificing quality.

\section{Downstream Applications}

To further demonstrate why making the 3D generative models orientation aligned is important, we implement two downstream applications, including zero-shot model-free object orientation estimation \secref{sec:ori_est} and efficient arrow-based object rotation manipulation \secref{sec:rot_mani}.

\begin{figure}
    \centering
    \includegraphics[width=0.99\linewidth]{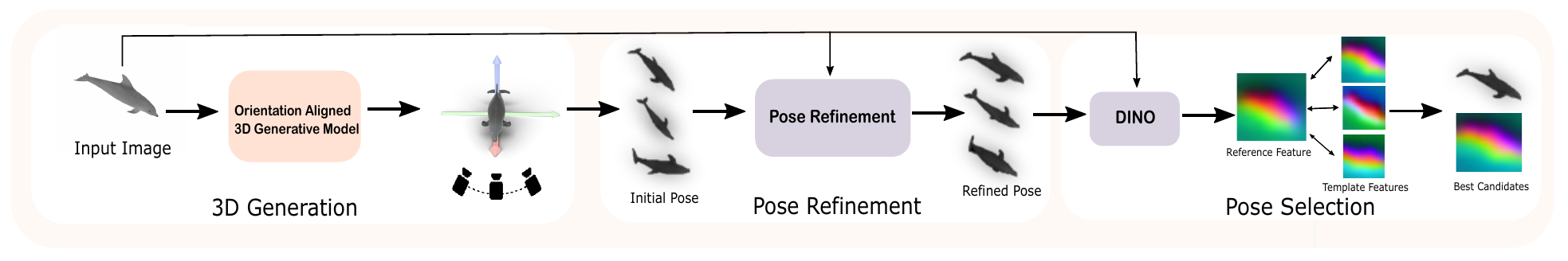}
    \caption{\textbf{Zero-Shot Orientation Estimation}. Our orientation-aligned 3D object acts as a template for pose estimation by rendering it from multiple views, refining each, and selecting the best-matching viewpoint. Note that we do not perform training for this downstream task, where the pose refinement module is directly from FoundationPose~\cite{Wen2023FoundationPoseU6}, and the pose selection module directly utilizes the pre-trained DINO feature extractor~\cite{Oquab2023DINOv2LR}. }
    \label{fig:orientation_pipeline}
    \vspace{-1.5em}
\end{figure}

\subsection{Zero-shot Model-free Object Orientation Estimation} \label{sec:ori_est}

One popular class of solutions to 3D object orientation estimation is based on analysis-by-synthesis~\cite{Wang2019NormalizedOC, Chen2020CategoryLO, Guo2022AVN}, but most of them are limited to category-level due to the difficulty in synthesizing orientation-aligned 3D models across categories. 
Another line of methods~\cite{Peng2018PVNetPV,Wen2023FoundationPoseU6} estimates poses relative to a given CAD model, limiting practical use since a corresponding model must be available for each input image. While one could replace CAD models with generated 3D shapes (e.g., from Trellis), the misaligned outputs of existing generative models result in unreliable pose references. In contrast, our method generates orientation-aligned 3D shapes directly from a single image, enabling the prediction of \textit{absolute poses} in a canonical frame by treating the generated 3D shape as a per-object template. 
Specifically, we build on a state-of-the-art template-based pose estimation method, FoundationPose~\cite{Wen2023FoundationPoseU6}, which has strong generalizability due to training on a large synthetic dataset. It renders templates from a fixed set of viewpoints, refines each pose, and then selects the best match. Although FoundationPose is trained with accurate CAD models and depth maps, we find that its pose refinement module remains effective even with our generated 3D shapes and without depth input. However, its pose selection performance degrades when templates differ slightly in geometry or appearance.
To address this, we retain the pose refinement module but replace the pose selection stage with a DINOv2~\cite{Oquab2023DINOv2LR}-feature-based similarity metric, as shown in \figref{fig:orientation_pipeline}. This is achieved by computing the L2 distance between DINOv2 patch feature maps of each refined rendering and the target image, and selecting the view with the highest similarity. Note that when estimating orientations of objects in scene-level images, we need to extract objects' masks via the image segmentation method~\cite{Kirillov2023SegmentA} before 3D generation.

\subsection{Efficient Arrow-based Object Rotation Manipulation} \label{sec:rot_mani}

Efficient manipulation of 3D model rotation within simulation systems is crucial, yet challenging, especially when models are initialized in non-canonical poses. This difficulty arises because users typically aim to orient a model toward a desired direction, but conventional 3D simulation systems only record the model's pose relative to its initial state. As a result, users must manually compensate for any misalignment in the initial pose, complicating the interaction process. In contrast, our generated orientation-aligned 3D models enable a more direct and user-friendly rotation manipulation approach due to their aligned initial orientations. With this alignment, users can simply draw an arrow indicating the desired forward-facing direction, without needing to consider the model’s original pose. This arrow-based interaction paradigm enhances usability and is compatible with both augmented reality (AR) applications and general-purpose 3D software.
In AR applications, users draw an arrow in the 2D image plane, which is subsequently lifted into 3D space using monocular depth estimation techniques (e.g., ~\cite{Piccinelli2024UniDepthUM}). The system then rotates the model so that its forward-facing axis aligns with the specified arrow, while ensuring the object remains grounded on the background plane. Note that because our models are normalized in scale, their size must still be specified, either through a large language model (LLM) or direct user input.
In general 3D software, users draw the arrow directly in 3D space. The system then applies Rodrigues’ rotation formula to align the model’s orientation with the user-specified direction. Additional implementation details are provided in the supplementary materials ~\secref{sec:method_implementation}.

\begin{figure}[htbp]
    \centering
    \includegraphics[width=0.99\linewidth]{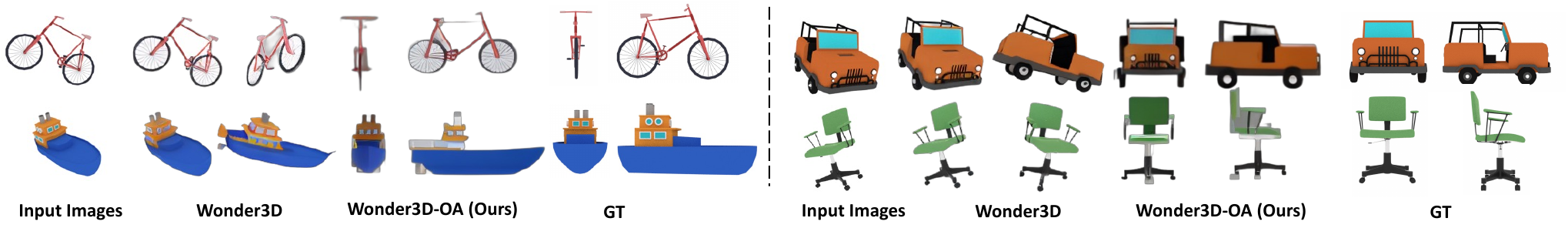}
    \caption{\textbf{Qualitative Results} on multi-view diffusion backbone, Wonder3D. For each input image, we show two views with the same index from the multi-view predictions.}
    \vspace{-0.5em}
    \label{fig:wonder3d}
\end{figure}

\begin{table*}[t]
\centering
\resizebox{.9\textwidth}{!} 
{
\begin{tabular}{l|ccc|ccc}
\toprule
& \multicolumn{3}{c|}{GSO~\cite{Downs2022GoogleSO}} & \multicolumn{3}{c}{Toys4k~\cite{Stojanov2021UsingST}} \\
& CD$\downarrow$ & LPIPS$\downarrow$ & CLIP$\uparrow$ & CD$\downarrow$& LPIPS$\downarrow$ & CLIP$\uparrow$\\
\midrule
Wonder3D & 0.0894 & 0.2799 & 76.37 & 0.0932 & 0.2859 & 87.10\\
Wonder3D + PCA & \tabyellow 0.0788 & \tabyellow 0.2554 &  \tabyellow 77.80 & \tabyellow 0.0858 & \tabyellow 0.2691 & 87.58\\
Wonder3D + VLM (Gemini-2.0)~\cite{gemini} &  0.0850 & 0.2752 & 76.30 &  0.0880 & 0.2804 &  87.53\\
Wonder3D + Orient Anything (ViT-L)~\cite{Wang2024OrientAL} & 0.1015 & 0.2600 & 77.50 & 0.1079 & 0.2699 &  \tabyellow 88.12\\
\midrule
Wonder3D-OA (ours w/o LGM) & \taborange 0.0609 & \taborange 0.2300 & \taborange 80.22 & \taborange 0.0571 & \taborange 0.2351 & \taborange 91.33\\
Wonder3D-OA (ours) & \tabred \textbf{0.0564} & \tabred \textbf{0.2270} & \tabred \textbf{80.30} & \tabred \textbf{0.0548} & \tabred \textbf{0.2317} & \tabred \textbf{92.09}\\
\bottomrule
\end{tabular}
}
\captionof{table}{\textbf{Quantitative Comparison} of geometry and appearance on Multi-view Diffusion backbone~\cite{Long2023Wonder3DSI}. We highlight the \colorbox[HTML]{F47983}{best}, \colorbox[HTML]{FFC773}{second-best}, and \colorbox[HTML]{faff72}{third-best} scores achieved on any metrics.}
\label{tab:quality_mv_diffusion}

\end{table*}

\section{Experiment} \label{experiment}

\subsection{Implementation Details}
\noindent \textbf{Dataset.} Orientation-aligned 3D generative models Trellis-OA and Wonder3D-OA are trained on our Objaverse-OA dataset, which is curated from Objaverse-LVIS~\cite{Deitke2022ObjaverseAU}. The base multi-view diffusion model is trained on Objaverse~\cite{Deitke2022ObjaverseAU}, and the base 3D-VAE-based model is trained on TRELLIS-500K~\cite{Xiang2024Structured3L}. To demonstrate the generalizability and accuracy of our method's orientation alignment ability, we evaluate on two unseen datasets, GSO~\cite{Downs2022GoogleSO} and Toys4k~\cite{Stojanov2021UsingST}. To further demonstrate the sim-to-real generalizability, we also evaluate on the real-world dataset Imagenet3D~\cite{Ma2024ImageNet3DTG}.

\noindent \textbf{Baselines.} For the task of aligned object generation, there are no existing baselines for this task. Therefore, we design baselines that perform this task in two stages: 1) object generation with misaligned orientations, and 2) orient them to aligned poses based on pose estimation using different variants: (i) Principal Component Analysis (PCA); (ii) advanced Vision Language Model (VLM) Gemini-2.0~\cite{gemini}; and (iii) zero-shot model-free orientation estimation method, Orient Anything~\cite{Wang2024OrientAL}.
For the task of zero-shot orientation estimation, we compare our method with
Orient Anything~\cite{Wang2024OrientAL} and FSDetView~\cite{Xiao2020FewShotOD}. Note that FSDetView doesn't support zero-shot estimation. Therefore, we evaluate its performance only on its supported categories.

\noindent \textbf{Metrics.} To evaluate the orientation alignment ability, we rotate reconstructed 3D models using different kinds of methods and calculate Chamfer Distance (CD), LPIPS~\cite{Zhang2018TheUE}, and CLIP~\cite{CLIP2021ICML} scores to measure the orientation alignment quality. To evaluate the performance of our zero-shot orientation estimation method, we calculate Acc@30 and orientation absolute error (Abs) according to the rotation error. We follow NOCS to calculate the rotation $\emph{e}_R$ defined by: 
$   \emph{e}_R = arccos \frac{Tr(\Tilde{R} \cdot R^T) - 1}{2}$,
where $Tr$ represents the trace of the matrix. Note that for stick-like objects, top and side directions typically have ambiguity. Therefore, we only calculate the rotation error in the front direction.

\noindent \textbf{Training and inference time.} To fine-tune Trellis-OA, we use a total batch size of 64 for training 30000 steps, which takes only about 10 hours on the cluster of 8 Nvidia Tesla A100 GPUs. To fine-tune Wonder3D-OA, we use a total batch size of 512 for training 40000 steps, which takes about 3 days on the cluster of 8 Nvidia Tesla A100 GPUs..  For the 3D object generation, Trellis-OA takes 32.93s and Wonder3D-OA takes 6.69s for multi-view generation, 2.53s for fused 3DGS generation, and 45.37s for mesh extraction. Given the generated 3D models, our pose refinement module takes 6.54s and the pose selection module takes 7.47s, which are comparable with the SOTA template-based pose estimation method FoundationPose~\cite{Wen2023FoundationPoseU6}. 

\subsection{Orientation-Aligned Object Generation}

\begin{table*}[t]
\centering
\resizebox{.9\textwidth}{!} 
{
\begin{tabular}{l|ccc|ccc}
\toprule
& \multicolumn{3}{c|}{GSO~\cite{Downs2022GoogleSO}} & \multicolumn{3}{c}{Toys4k~\cite{Stojanov2021UsingST}} \\
& CD$\downarrow$ & LPIPS$\downarrow$ & CLIP$\uparrow$ & CD$\downarrow$& LPIPS$\downarrow$ & CLIP$\uparrow$\\
\midrule
Trellis & 0.0770 & 0.2502 & 82.88 & 0.0951 & 0.2722 & 92.31\\
Trellis + PCA &  0.0798 & 0.2771 & 78.84 &  0.0911 & 0.2758 & 88.09\\
Trellis + VLM (Gemini-2.0)~\cite{gemini} &  \taborange 0.0421 & \tabred \textbf{0.1998} & \tabred \textbf{89.97} & \tabyellow 0.0564 & \tabyellow 0.2137 & \taborange 95.19\\
Trellis + Orient Anything (ViT-L)~\cite{Wang2024OrientAL} & 0.0604 & \tabyellow 0.2193 & \taborange 88.61 & 0.0725 & 0.2313 & \tabyellow 94.81\\
\midrule
Trellis-OA (ours small) & \tabyellow 0.0448 & 0.2224 & 82.46 & \taborange 0.0465 & \taborange 0.2055 & 93.74\\
Trellis-OA (ours) & \tabred \textbf{0.0407} & \taborange 0.2118 & \tabyellow 88.41 & \tabred \textbf{0.0393} & \tabred \textbf{0.1932} & \tabred \textbf{95.71}\\
\bottomrule
\end{tabular}
}
\captionof{table}{\textbf{Quantitative Comparison} of geometry and appearance on 3D-VAE backbone~\cite{Xiang2024Structured3L}.}
\label{tab:quality_3d_vae}
\end{table*}

\begin{figure}[t]
    \centering
    \includegraphics[width=0.99\linewidth]{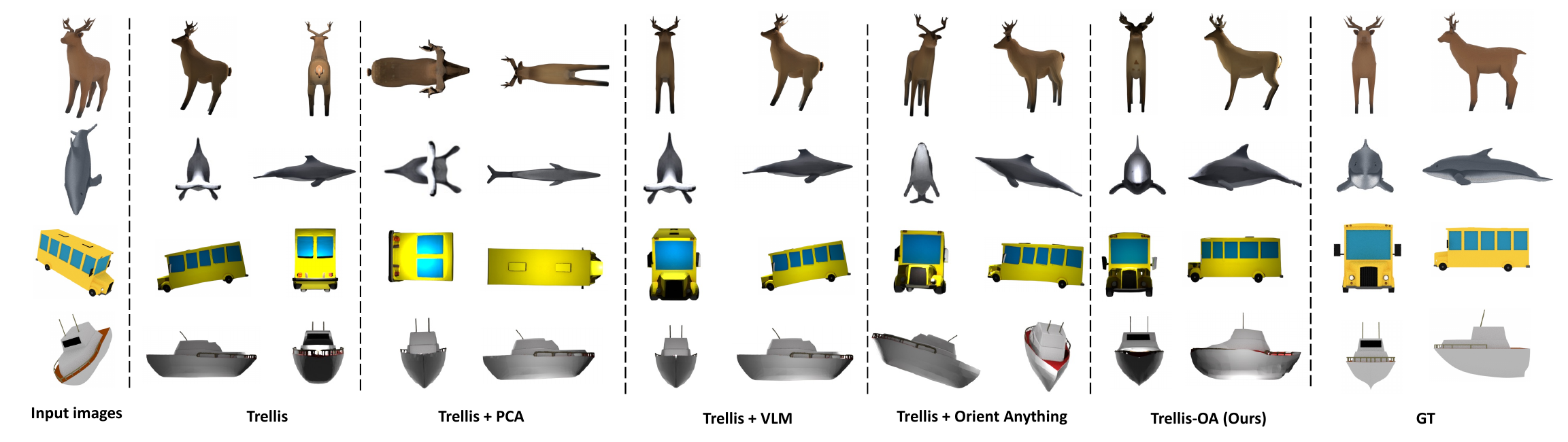}
    \caption{\textbf{Qualitative Results} based on the 3D-VAE backbone, Trellis. For each input image, we render the reconstructed object from two consistent views. Note that our method consistently generates objects in their canonical space. }
    \vspace{-0.5em}
    \label{fig:3d-vae}
\end{figure}

\begin{figure}[!h]
    \small
    \centering
    \begin{minipage}{0.55\linewidth}
        \centering
        \includegraphics[width=\linewidth]{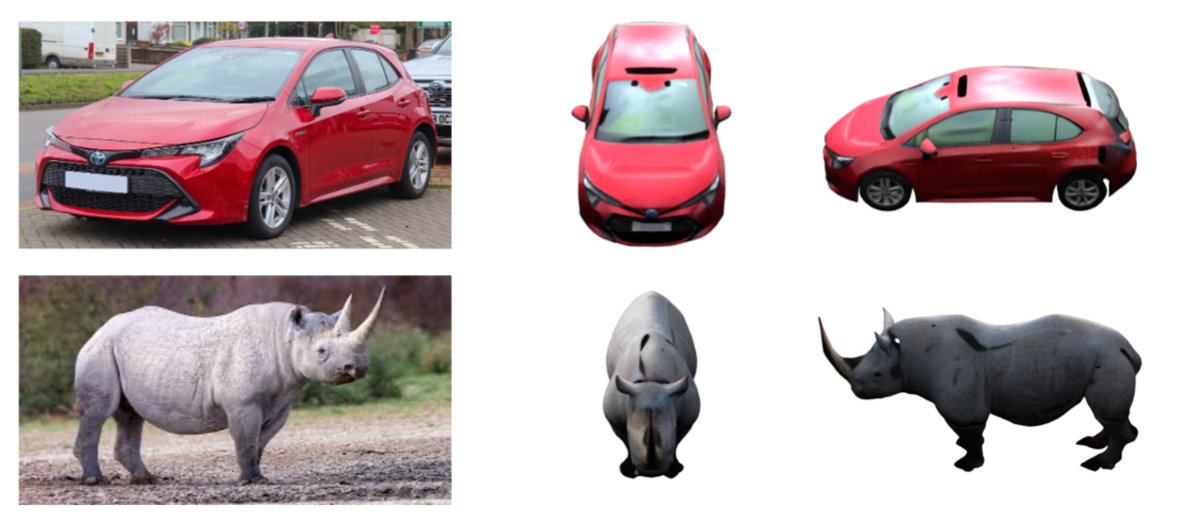}
        \vspace{-2em}
        \caption{\textbf{Qualitative Results} of Trellis-OA on real world internet images.}
        \label{fig:real_world}
    \end{minipage}
    \hfill
    \begin{minipage}{0.41\linewidth}
        \centering
        \includegraphics[width=\linewidth]{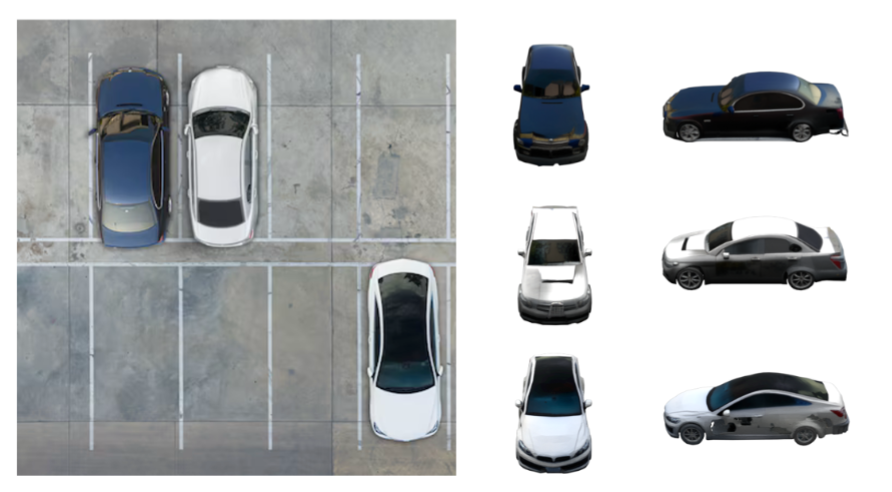}
        \caption{\textbf{Qualitative Results} of Trellis-OA on multi-objects images.}
        \label{fig:multi-objects}
    \end{minipage}
    \vspace{-1em}
\end{figure}

\boldparagraph{Comparison with baselines} 
We first evaluate the performance of orientation-aligned object generation.
As shown in the \tabref{tab:quality_mv_diffusion} and \tabref{tab:quality_3d_vae}, our fine-tuned models, Wonder3D-OA and Trellis-OA, surpass most two-stage baselines in geometry and appearance on both GSO and Toys4k datasets. Note that our method is suboptimal in appearance on the GSO dataset compared to the VLM baseline. It is probably because the GSO dataset has more objects with irregular geometry and appearance, which is challenging for our fine-tuned generator to handle. 
We present more qualitative results in the \figref{fig:wonder3d} and \figref{fig:3d-vae}. As shown in \figref{fig:3d-vae}, PCA cannot handle objects with different shape features and has difficulty in distinguishing the direction of estimated principal axes. VLM has trouble recognizing objects with unclear front-view features, like dolphins and boats. Orient Anything is still not that accurate, and typically rotates objects into leaning poses. In contrast, our method accurately aligns the orientation of the generated 3D models without obvious geometry and appearance degradation. Besides, our method can handle real world internet images and multi-objects images as shown in \figref{fig:real_world} and \figref{fig:multi-objects}. Furthermore, due to the high geometry quality of our Objaverse-OA, Trellis-OA eliminates the production of plane-like geometries. Please refer to the supplementary materials ~\secref{sec:more_results} for more qualitative results.

\boldparagraph{Ablation Study}
To assess the importance of using an orientation-aligned dataset with diverse categories, we conduct an ablation study by fine-tuning 3D generative models on a reduced set of 100 categories with 5720 objects, see (ours small) in \tabref{tab:quality_mv_diffusion} and \tabref{tab:quality_3d_vae}. The results show that reducing the number of training categories degrades both geometric and visual quality, highlighting the value of large category diversity in our Objaverse-OA dataset.

\subsection{Zero-Shot Object Orientation Estimation}

\newcolumntype{C}[1]{>{\centering\arraybackslash}p{#1}}

\begin{table*}[h]
\centering
\resizebox{.9\textwidth}{!} 
{
\begin{tabular}{C{4cm}|C{2.5cm}C{2.5cm}|C{2.5cm}C{2.5cm}}
\toprule
& \multicolumn{2}{c|}{Toys4k~\cite{Stojanov2021UsingST}} & \multicolumn{2}{c}{Stick-like Obj. from ImageNet3D~\cite{Ma2024ImageNet3DTG}} \\
& Acc@30$\uparrow$ & Abs$\downarrow$ & Acc@30$\uparrow$ & Abs$\downarrow$\\
\midrule
FSDetView~\cite{Xiao2020FewShotOD} (Few-shot) & 20.90 & 91.66 & \tabyellow 10.29 &  84.25 \\
Orient Anything~\cite{Wang2024OrientAL} (Vit-S) & 42.05 & \tabyellow 52.72 & 2.63 &  81.70 \\
Orient Anything~\cite{Wang2024OrientAL} (Vit-L) & \tabred \textbf{63.18} & \tabred \textbf{36.37} & 9.8 & \tabyellow 78.19\\
Ours (Vit-S) & \tabyellow 51.15 &  53.94 & \taborange 60.78 & \taborange 40.12\\
Ours (ViT-L) & \taborange 52.87 & \taborange 46.76 & \tabred \textbf{62.25} & \tabred \textbf{34.20}\\
\bottomrule
\end{tabular}
}
\captionof{table}{\textbf{Quantitative Comparison} of zero-shot orientation estimation on Toys4k~\cite{Stojanov2021UsingST} and stick-like objects from ImageNet3D~\cite{Ma2024ImageNet3DTG}.}

\label{tab:orientation_est}
\end{table*}

\begin{figure}[h]
    \small
    \centering
    \begin{minipage}{0.48\linewidth}
        \centering
        \includegraphics[width=\linewidth]{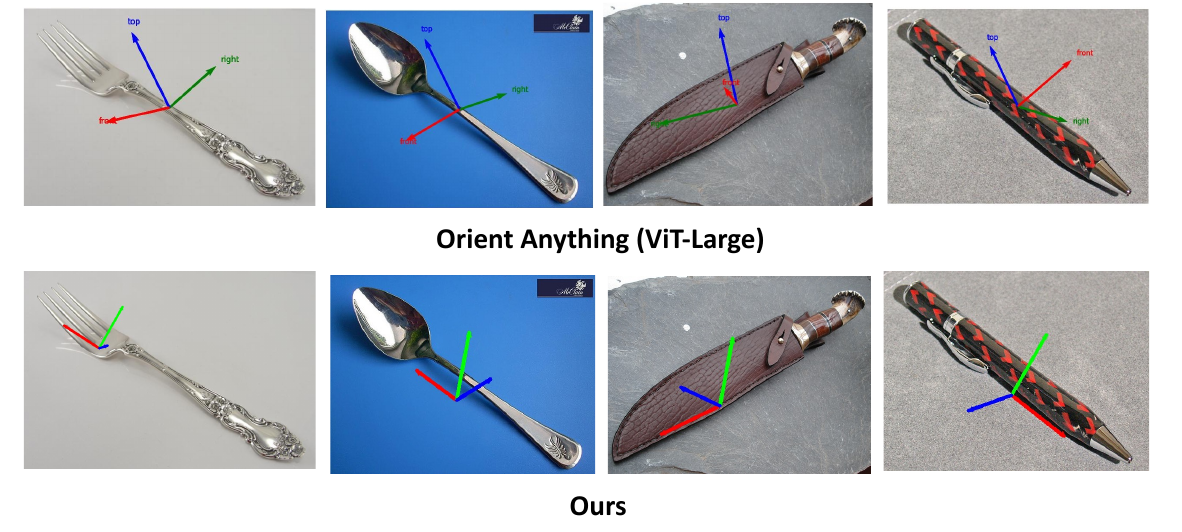}
        \caption{\textbf{Orientation Estimation Comparison} on stick-like objects from ImageNet3D~\cite{Ma2024ImageNet3DTG}.}
        \label{fig:image1}
    \end{minipage}
    \hfill
    \begin{minipage}{0.48\linewidth}
        \centering
        \includegraphics[width=\linewidth]{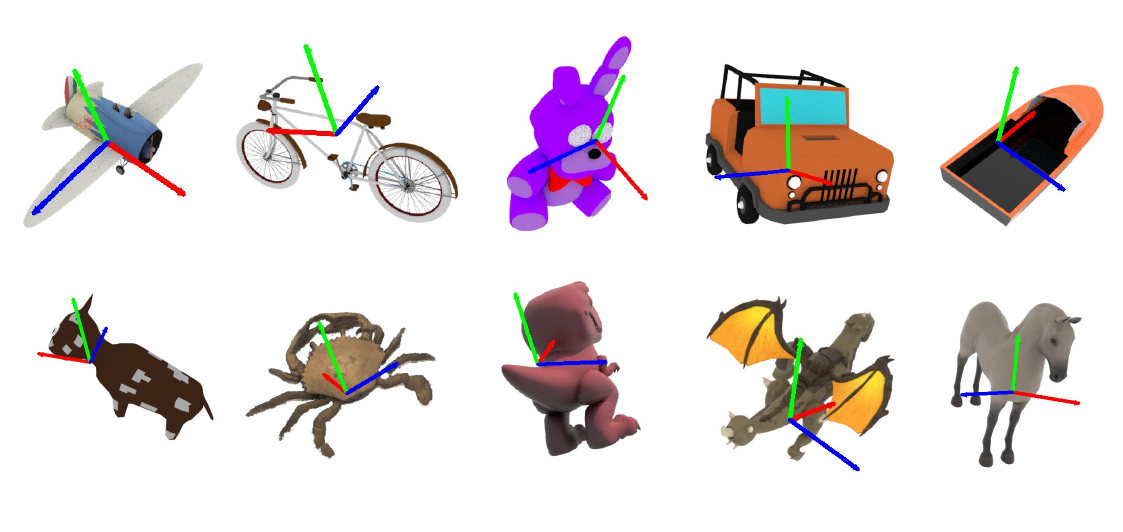}
        \vspace{-1.5em}
        \caption{\textbf{Orientation Estimation Results} of our method on Toys4k~\cite{Stojanov2021UsingST}.}
        \label{fig:image2}
    \end{minipage}
\end{figure}

\boldparagraph{Comparison with baselines} 
As shown in \tabref{tab:orientation_est}, without specific training, our method is already comparable with the SOTA orientation estimation method~\cite{Wang2024OrientAL} with the ViT-large architecture and surpasses \cite{Wang2024OrientAL} with the ViT-small architecture by a large margin. Besides, our method can handle challenging stick-like objects collected from the real-world dataset while baselines all fail to work, which further demonstrates our utility in coping with long-tailed situations.

\subsection{Efficient Arrow-based Object Rotation Manipulation}

\begin{figure}[H]
    \centering
    \begin{minipage}{0.49\textwidth}
    \includegraphics[width=\linewidth]{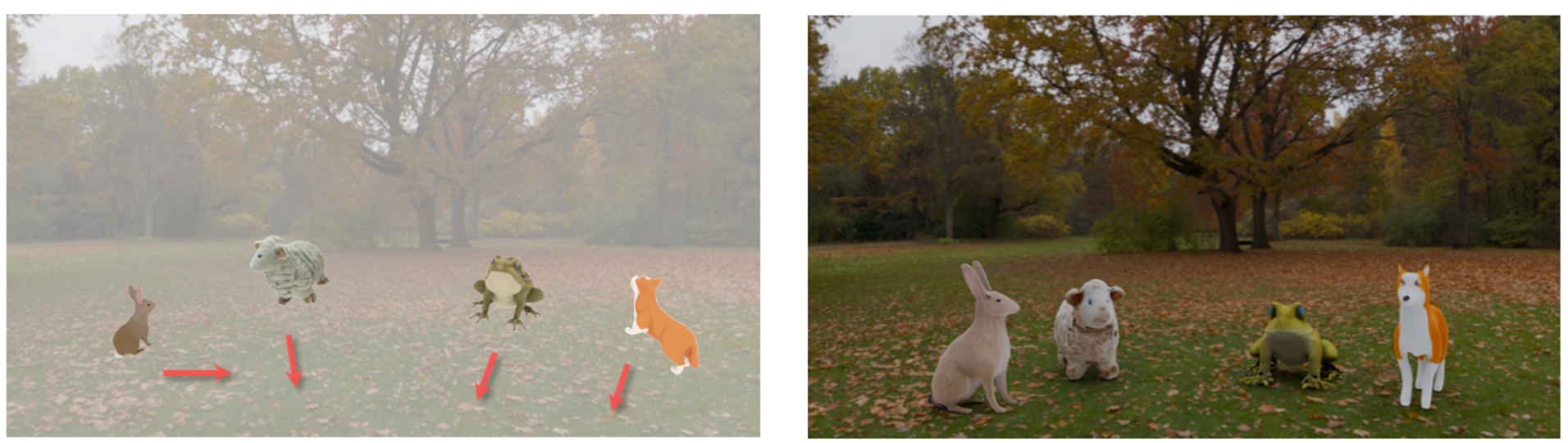}
    \end{minipage}
    \hfill
    \begin{minipage}{0.49\textwidth}
    \includegraphics[width=\linewidth]{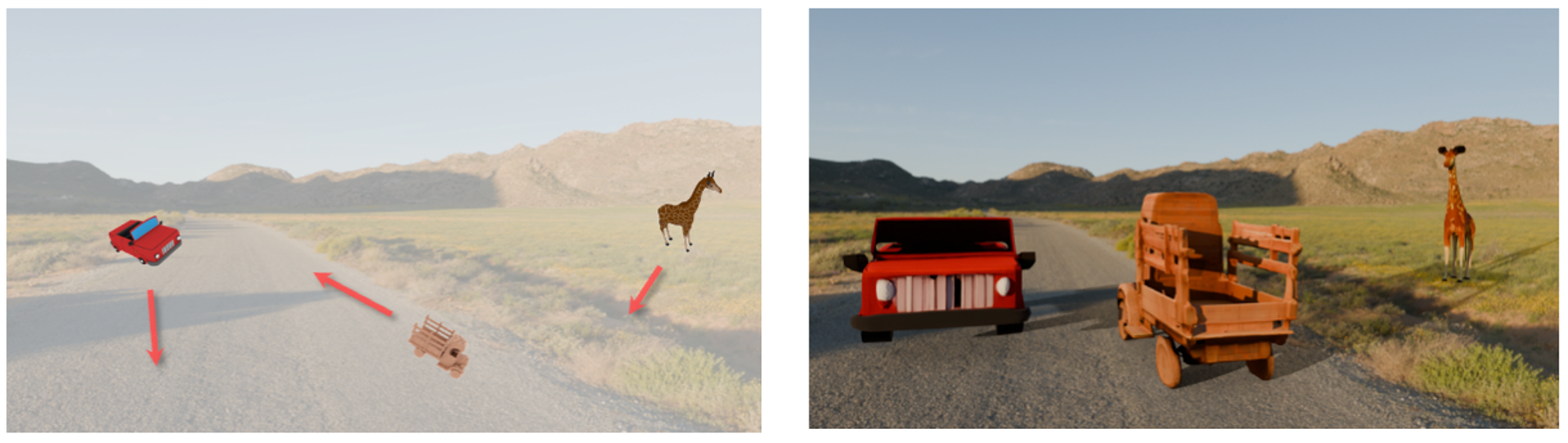}
    \end{minipage}
    \caption{\textbf{Qualitative results} of our efficient arrow-based object rotation manipulation method in augmented reality applications.}
    \label{fig:object_insertion}
\end{figure}

\figref{fig:teaser} and \figref{fig:object_insertion} illustrate our arrow-based object rotation manipulation method in augmented reality applications. Given an image of an object, we generate the corresponding object in its canonical pose using Trellis-OA, and then manipulate it in the reference image with the target orientation indicated by a user-specified arrow. This allows for efficient object manipulation without adjusting the object's orientation in a post-hoc fashion.  
We present more qualitative results of our method in the general 3D software in the video of supplementary materials, which demonstrate the usage of our approach.

\section{Conclusion and Limitation} \label{conclusion}

In this paper, we aim to align the orientations of the 3D generative models for downstream orientation estimation and efficient object rotation manipulation in 3D simulation systems. Towards this goal, we construct Objaverse-OA, a dataset covering orientation-aligned 3D models across the largest number of categories. Based on Objaverse-OA, we align the orientations of 3D generative models based on two popular 3D generation frameworks, including multi-view diffusion and 3D-VAE. Built upon the orientation-aligned 3D generative models, we develop a simple but effective orientation estimation approach following the analysis-by-synthesis paradigm and an efficient arrow-based object manipulation method. 
While our method achieves promising zero-shot orientation estimation without task-specific training, future work could further improve performance by training a dedicated model that leverages the generated 3D objects as templates.

\section{Acknowledgement} \label{sec:ack}

This work is supported by NSFC under grant 62441223 and 62202418, and partially by Ant Group.

{
    \small
    \bibliographystyle{ieeenat_fullname}
    \bibliography{bibliography, bibliography_long, bibliography_custom}

@inproceedings{CLIP2021ICML,
  author       = {Alec Radford and
                  Jong Wook Kim and
                  Chris Hallacy and
                  Aditya Ramesh and
                  Gabriel Goh and
                  Sandhini Agarwal and
                  Girish Sastry and
                  Amanda Askell and
                  Pamela Mishkin and
                  Jack Clark and
                  Gretchen Krueger and
                  Ilya Sutskever},
  editor       = {Marina Meila and
                  Tong Zhang},
  title        = {Learning Transferable Visual Models From Natural Language Supervision},
  booktitle    = {Proceedings of the 38th International Conference on Machine Learning,
                  {ICML} 2021, 18-24 July 2021, Virtual Event},
  series       = {Proceedings of Machine Learning Research},
  volume       = {139},
  pages        = {8748--8763},
  publisher    = {{PMLR}},
  year         = {2021},
}

@inproceedings{Treat2018GENERATIVEAN,
  title={GENERATIVE ADVERSARIAL NETS},
  author={Individualized Treat and Jinsung Yoon},
  year={2018},
  url={https://api.semanticscholar.org/CorpusID:10319744}
}

@inproceedings{Wu2016LearningAP,
  title={Learning a Probabilistic Latent Space of Object Shapes via 3D Generative-Adversarial Modeling},
  author={Jiajun Wu and Chengkai Zhang and Tianfan Xue and Bill Freeman and Joshua B. Tenenbaum},
  booktitle={Neural Information Processing Systems},
  year={2016},
  url={https://api.semanticscholar.org/CorpusID:3248075}
}

@article{Schwarz2020GRAFGR,
  title={GRAF: Generative Radiance Fields for 3D-Aware Image Synthesis},
  author={Katja Schwarz and Yiyi Liao and Michael Niemeyer and Andreas Geiger},
  journal={ArXiv},
  year={2020},
  volume={abs/2007.02442},
  url={https://api.semanticscholar.org/CorpusID:220364071}
}

@article{Chan2021EfficientG3,
  title={Efficient Geometry-aware 3D Generative Adversarial Networks},
  author={Eric Chan and Connor Z. Lin and Matthew Chan and Koki Nagano and Boxiao Pan and Shalini De Mello and Orazio Gallo and Leonidas J. Guibas and Jonathan Tremblay and S. Khamis and Tero Karras and Gordon Wetzstein},
  journal={2022 IEEE/CVF Conference on Computer Vision and Pattern Recognition (CVPR)},
  year={2021},
  pages={16102-16112},
  url={https://api.semanticscholar.org/CorpusID:245144673}
}

@article{Pavllo2022ShapePA,
  title={Shape, Pose, and Appearance from a Single Image via Bootstrapped Radiance Field Inversion},
  author={Dario Pavllo and David Joseph Tan and Marie-Julie Rakotosaona and Federico Tombari},
  journal={2023 IEEE/CVF Conference on Computer Vision and Pattern Recognition (CVPR)},
  year={2022},
  pages={4391-4401},
  url={https://api.semanticscholar.org/CorpusID:253734370}
}

@article{Croitoru2022DiffusionMI,
  title={Diffusion Models in Vision: A Survey},
  author={Florinel-Alin Croitoru and Vlad Hondru and Radu Tudor Ionescu and Mubarak Shah},
  journal={IEEE Transactions on Pattern Analysis and Machine Intelligence},
  year={2022},
  volume={45},
  pages={10850-10869},
  url={https://api.semanticscholar.org/CorpusID:252199918}
}

@article{Ho2020DenoisingDP,
  title={Denoising Diffusion Probabilistic Models},
  author={Jonathan Ho and Ajay Jain and P. Abbeel},
  journal={ArXiv},
  year={2020},
  volume={abs/2006.11239},
  url={https://api.semanticscholar.org/CorpusID:219955663}
}

@article{Poole2022DreamFusionTU,
  title={DreamFusion: Text-to-3D using 2D Diffusion},
  author={Ben Poole and Ajay Jain and Jonathan T. Barron and Ben Mildenhall},
  journal={ArXiv},
  year={2022},
  volume={abs/2209.14988},
  url={https://api.semanticscholar.org/CorpusID:252596091}
}

@article{Wang2022ScoreJC,
  title={Score Jacobian Chaining: Lifting Pretrained 2D Diffusion Models for 3D Generation},
  author={Haochen Wang and Xiaodan Du and Jiahao Li and Raymond A. Yeh and Gregory Shakhnarovich},
  journal={2023 IEEE/CVF Conference on Computer Vision and Pattern Recognition (CVPR)},
  year={2022},
  pages={12619-12629},
  url={https://api.semanticscholar.org/CorpusID:254125253}
}

@article{Long2023Wonder3DSI,
  title={Wonder3D: Single Image to 3D Using Cross-Domain Diffusion},
  author={Xiaoxiao Long and Yuanchen Guo and Cheng Lin and Yuan Liu and Zhiyang Dou and Lingjie Liu and Yuexin Ma and Song-Hai Zhang and Marc Habermann and Christian Theobalt and Wenping Wang},
  journal={2024 IEEE/CVF Conference on Computer Vision and Pattern Recognition (CVPR)},
  year={2023},
  pages={9970-9980},
  url={https://api.semanticscholar.org/CorpusID:264436465}
}

@article{Liu2023SyncDreamerGM,
  title={SyncDreamer: Generating Multiview-consistent Images from a Single-view Image},
  author={Yuan Liu and Chu-Hsing Lin and Zijiao Zeng and Xiaoxiao Long and Lingjie Liu and Taku Komura and Wenping Wang},
  journal={ArXiv},
  year={2023},
  volume={abs/2309.03453},
  url={https://api.semanticscholar.org/CorpusID:261582503}
}

@article{Wang2023ImageDreamIM,
  title={ImageDream: Image-Prompt Multi-view Diffusion for 3D Generation},
  author={Peng Wang and Yichun Shi},
  journal={ArXiv},
  year={2023},
  volume={abs/2312.02201},
  url={https://api.semanticscholar.org/CorpusID:265659122}
}

@article{Shi2023Zero123AS,
  title={Zero123++: a Single Image to Consistent Multi-view Diffusion Base Model},
  author={Ruoxi Shi and Hansheng Chen and Zhuoyang Zhang and Minghua Liu and Chao Xu and Xinyue Wei and Linghao Chen and Chong Zeng and Hao Su},
  journal={ArXiv},
  year={2023},
  volume={abs/2310.15110},
  url={https://api.semanticscholar.org/CorpusID:264436559}
}

@article{Szymanowicz2023ViewsetD,
  title={Viewset Diffusion: (0-)Image-Conditioned 3D Generative Models from 2D Data},
  author={Stanislaw Szymanowicz and C. Rupprecht and Andrea Vedaldi},
  journal={2023 IEEE/CVF International Conference on Computer Vision (ICCV)},
  year={2023},
  pages={8829-8839},
  url={https://api.semanticscholar.org/CorpusID:259144886}
}

@article{Shi2023MVDreamMD,
  title={MVDream: Multi-view Diffusion for 3D Generation},
  author={Yichun Shi and Peng Wang and Jianglong Ye and Mai Long and Kejie Li and X. Yang},
  journal={ArXiv},
  year={2023},
  volume={abs/2308.16512},
  url={https://api.semanticscholar.org/CorpusID:261395233}
}

@article{Liang2023LucidDreamerTH,
  title={LucidDreamer: Towards High-Fidelity Text-to-3D Generation via Interval Score Matching},
  author={Yixun Liang and Xin Yang and Jiantao Lin and Haodong Li and Xiaogang Xu and Yingcong Chen},
  journal={2024 IEEE/CVF Conference on Computer Vision and Pattern Recognition (CVPR)},
  year={2023},
  pages={6517-6526},
  url={https://api.semanticscholar.org/CorpusID:265295106}
}

@article{Lin2022Magic3DHT,
  title={Magic3D: High-Resolution Text-to-3D Content Creation},
  author={Chen-Hsuan Lin and Jun Gao and Luming Tang and Towaki Takikawa and Xiaohui Zeng and Xun Huang and Karsten Kreis and Sanja Fidler and Ming-Yu Liu and Tsung-Yi Lin},
  journal={2023 IEEE/CVF Conference on Computer Vision and Pattern Recognition (CVPR)},
  year={2022},
  pages={300-309},
  url={https://api.semanticscholar.org/CorpusID:253708074}
}

@article{Tang2023MakeIt3DH3,
  title={Make-It-3D: High-Fidelity 3D Creation from A Single Image with Diffusion Prior},
  author={Junshu Tang and Tengfei Wang and Bo Zhang and Ting Zhang and Ran Yi and Lizhuang Ma and Dong Chen},
  journal={2023 IEEE/CVF International Conference on Computer Vision (ICCV)},
  year={2023},
  pages={22762-22772},
  url={https://api.semanticscholar.org/CorpusID:257757320}
}

@article{Tang2023DreamGaussianGG,
  title={DreamGaussian: Generative Gaussian Splatting for Efficient 3D Content Creation},
  author={Jiaxiang Tang and Jiawei Ren and Hang Zhou and Ziwei Liu and Gang Zeng},
  journal={ArXiv},
  year={2023},
  volume={abs/2309.16653},
  url={https://api.semanticscholar.org/CorpusID:263131552}
}

@article{Wang2023ProlificDreamerHA,
  title={ProlificDreamer: High-Fidelity and Diverse Text-to-3D Generation with Variational Score Distillation},
  author={Zhengyi Wang and Cheng Lu and Yikai Wang and Fan Bao and Chongxuan Li and Hang Su and Jun Zhu},
  journal={ArXiv},
  year={2023},
  volume={abs/2305.16213},
  url={https://api.semanticscholar.org/CorpusID:258887357}
}

@article{Xiang2024Structured3L,
  title={Structured 3D Latents for Scalable and Versatile 3D Generation},
  author={Jianfeng Xiang and Zelong Lv and Sicheng Xu and Yu Deng and Ruicheng Wang and Bowen Zhang and Dong Chen and Xin Tong and Jiaolong Yang},
  journal={ArXiv},
  year={2024},
  volume={abs/2412.01506},
  url={https://api.semanticscholar.org/CorpusID:274436286}
}

@article{Zhao2025Hunyuan3D2S,
  title={Hunyuan3D 2.0: Scaling Diffusion Models for High Resolution Textured 3D Assets Generation},
  author={Zibo Zhao and Zeqiang Lai and Qin Lin and Yunfei Zhao and Haolin Liu and Shuhui Yang and Yifei Feng and Mingxin Yang and Sheng Zhang and Xianghui Yang and Huiwen Shi and Si-Ya Liu and Junta Wu and Yihang Lian and Fan Yang and Ruining Tang and Ze-Bao He and Xinzhou Wang and Jian Liu and Xuhui Zuo and Zhuo Chen and Biwen Lei and Haohan Weng and Jing Xu and Yi Zhu and Xinhai Liu and Lixin Xu and Chang-Ping Hu and Tianyu Huang and Lifu Wang and Jihong Zhang and Mengya Chen and Liang Dong and Yi-yong Jia and Yu-Xin Cai and Jiaao Yu and Yi Jun Tang and Hao Zhang and Zhengfeng Ye and Peng He and Runzhou Wu and Chao Zhang and Yonghao Tan and Jie Xiao and Yang-Dan Tao and Jian-Xi Zhu and Ji Xue and Kai Liu and Chongqing Zhao and Xinming Wu and Zhi-wei Hu and Lei Qin and Jian-Yong Peng and Zhan Li and Minghui Chen and Xipeng Zhang and Lin Niu and Paige Wang and Yingkai Wang and Hao Kuang and Zhongyi Fan and Xu Zheng and Weihao Zhuang and Yin-Yin He and Tian-Hai Liu and Yong Yang and Di Wang and Yuhong Liu and Jie Jiang and Jingwei Huang and Chunchao Guo},
  journal={ArXiv},
  year={2025},
  volume={abs/2501.12202},
  url={https://api.semanticscholar.org/CorpusID:275788857}
}

@article{Deitke2022ObjaverseAU,
  title={Objaverse: A Universe of Annotated 3D Objects},
  author={Matt Deitke and Dustin Schwenk and Jordi Salvador and Luca Weihs and Oscar Michel and Eli VanderBilt and Ludwig Schmidt and Kiana Ehsani and Aniruddha Kembhavi and Ali Farhadi},
  journal={2023 IEEE/CVF Conference on Computer Vision and Pattern Recognition (CVPR)},
  year={2022},
  pages={13142-13153},
  url={https://api.semanticscholar.org/CorpusID:254685588}
}

@article{Wen2023FoundationPoseU6,
  title={FoundationPose: Unified 6D Pose Estimation and Tracking of Novel Objects},
  author={Bowen Wen and Wei Yang and Jan Kautz and Stanley T. Birchfield},
  journal={2024 IEEE/CVF Conference on Computer Vision and Pattern Recognition (CVPR)},
  year={2023},
  pages={17868-17879},
  url={https://api.semanticscholar.org/CorpusID:266191252}
}

@article{Peng2018PVNetPV,
  title={PVNet: Pixel-Wise Voting Network for 6DoF Pose Estimation},
  author={Sida Peng and Yuan Liu and Qi-Xing Huang and Hujun Bao and Xiaowei Zhou},
  journal={2019 IEEE/CVF Conference on Computer Vision and Pattern Recognition (CVPR)},
  year={2018},
  pages={4556-4565},
  url={https://api.semanticscholar.org/CorpusID:57189382}
}

@article{Sun2022OnePoseOO,
  title={OnePose: One-Shot Object Pose Estimation without CAD Models},
  author={Jiaming Sun and Zihao Wang and Siyu Zhang and Xingyi He He and Hongcheng Zhao and Guofeng Zhang and Xiaowei Zhou},
  journal={2022 IEEE/CVF Conference on Computer Vision and Pattern Recognition (CVPR)},
  year={2022},
  pages={6815-6824},
  url={https://api.semanticscholar.org/CorpusID:249017678}
}

@article{Wang2019NormalizedOC,
  title={Normalized Object Coordinate Space for Category-Level 6D Object Pose and Size Estimation},
  author={He Wang and Srinath Sridhar and Jingwei Huang and Julien P. C. Valentin and Shuran Song and Leonidas J. Guibas},
  journal={2019 IEEE/CVF Conference on Computer Vision and Pattern Recognition (CVPR)},
  year={2019},
  pages={2637-2646},
  url={https://api.semanticscholar.org/CorpusID:57761160}
}

@inproceedings{Chen2024Learning3G,
  title={Learning 3D-Aware GANs from Unposed Images with Template Feature Field},
  author={Xinya Chen and Hanlei Guo and Yanrui Bin and Shangzhan Zhang and Yuanbo Yang and Yue Wang and Yujun Shen and Yiyi Liao},
  booktitle={European Conference on Computer Vision},
  year={2024},
  url={https://api.semanticscholar.org/CorpusID:269005219}
}

@inproceedings{Chen2020CategoryLO,
  title={Category Level Object Pose Estimation via Neural Analysis-by-Synthesis},
  author={Xu Chen and Zijian Dong and Jie Song and Andreas Geiger and Otmar Hilliges},
  booktitle={European Conference on Computer Vision},
  year={2020},
  url={https://api.semanticscholar.org/CorpusID:221068923}
}

@article{Chang2015ShapeNetAI,
  title={ShapeNet: An Information-Rich 3D Model Repository},
  author={Angel X. Chang and Thomas A. Funkhouser and Leonidas J. Guibas and Pat Hanrahan and Qi-Xing Huang and Zimo Li and Silvio Savarese and Manolis Savva and Shuran Song and Hao Su and Jianxiong Xiao and L. Yi and Fisher Yu},
  journal={ArXiv},
  year={2015},
  volume={abs/1512.03012},
  url={https://api.semanticscholar.org/CorpusID:2554264}
}

@article{Krishnan2024OmniNOCSAU,
  title={OmniNOCS: A unified NOCS dataset and model for 3D lifting of 2D objects},
  author={Akshay Krishnan and Abhijit Kundu and Kevis-Kokitsi Maninis and James Hays and Matthew Brown},
  journal={ArXiv},
  year={2024},
  volume={abs/2407.08711},
  url={https://api.semanticscholar.org/CorpusID:271098103}
}

@article{Ma2024ImageNet3DTG,
  title={ImageNet3D: Towards General-Purpose Object-Level 3D Understanding},
  author={Wufei Ma and Guanning Zeng and Guofeng Zhang and Qihao Liu and Letian Zhang and Adam Kortylewski and Yaoyao Liu and Alan L. Yuille},
  journal={ArXiv},
  year={2024},
  volume={abs/2406.09613},
  url={https://api.semanticscholar.org/CorpusID:270521689}
}

@article{Wang2024OrientAL,
  title={Orient Anything: Learning Robust Object Orientation Estimation from Rendering 3D Models},
  author={Zehan Wang and Ziang Zhang and Tianyu Pang and Chao Du and Hengshuang Zhao and Zhou Zhao},
  journal={ArXiv},
  year={2024},
  volume={abs/2412.18605},
  url={https://api.semanticscholar.org/CorpusID:274992375}
}

@inproceedings{Tang2024LGMLM,
  title={LGM: Large Multi-View Gaussian Model for High-Resolution 3D Content Creation},
  author={Jiaxiang Tang and Zhaoxi Chen and Xiaokang Chen and Tengfei Wang and Gang Zeng and Ziwei Liu},
  booktitle={European Conference on Computer Vision},
  year={2024},
  url={https://api.semanticscholar.org/CorpusID:267523413}
}

@inproceedings{Xu2024GRMLG,
  title={GRM: Large Gaussian Reconstruction Model for Efficient 3D Reconstruction and Generation},
  author={Yinghao Xu and Zifan Shi and Wang Yifan and Hansheng Chen and Ceyuan Yang and Sida Peng and Yujun Shen and Gordon Wetzstein},
  booktitle={European Conference on Computer Vision},
  year={2024},
  url={https://api.semanticscholar.org/CorpusID:268554137}
}

@article{Downs2022GoogleSO,
  title={Google Scanned Objects: A High-Quality Dataset of 3D Scanned Household Items},
  author={Laura Downs and Anthony Francis and Nate Koenig and Brandon Kinman and Ryan Michael Hickman and Krista Reymann and Thomas Barlow McHugh and Vincent Vanhoucke},
  journal={2022 International Conference on Robotics and Automation (ICRA)},
  year={2022},
  pages={2553-2560},
  url={https://api.semanticscholar.org/CorpusID:248392390}
}

@article{Stojanov2021UsingST,
  title={Using Shape to Categorize: Low-Shot Learning with an Explicit Shape Bias},
  author={Stefan Stojanov and Anh Thai and James M. Rehg},
  journal={2021 IEEE/CVF Conference on Computer Vision and Pattern Recognition (CVPR)},
  year={2021},
  pages={1798-1808},
  url={https://api.semanticscholar.org/CorpusID:231639354}
}

@article{Xiao2020FewShotOD,
  title={Few-Shot Object Detection and Viewpoint Estimation for Objects in the Wild},
  author={Yang Xiao and Vincent Lepetit and Renaud Marlet},
  journal={IEEE Transactions on Pattern Analysis and Machine Intelligence},
  year={2020},
  volume={45},
  pages={3090-3106},
  url={https://api.semanticscholar.org/CorpusID:220713477}
}

@article{Murray2002ShapePR,
  title={Shape perception reduces activity in human primary visual cortex},
  author={Scott O. Murray and Daniel J. Kersten and Bruno A. Olshausen and Paul Schrater and David L. Woods},
  journal={Proceedings of the National Academy of Sciences of the United States of America},
  year={2002},
  volume={99},
  pages={15164 - 15169},
  url={https://api.semanticscholar.org/CorpusID:1622719}
}

@misc{gemini,
  title = {Gemini-2.0},
  author={Sundar Pichai and Demis Hassabis and Koray Kavukcuoglu},
  howpublished={\url{https://blog.google/technology/google-deepmind/google-gemini-ai-update-december-2024}},
  note = {Accessed: 2025-05-14}
}

@article{Rombach2021HighResolutionIS,
  title={High-Resolution Image Synthesis with Latent Diffusion Models},
  author={Robin Rombach and A. Blattmann and Dominik Lorenz and Patrick Esser and Bj{\"o}rn Ommer},
  journal={2022 IEEE/CVF Conference on Computer Vision and Pattern Recognition (CVPR)},
  year={2021},
  pages={10674-10685},
  url={https://api.semanticscholar.org/CorpusID:245335280}
}

@article{Wang2021NeuSLN,
  title={NeuS: Learning Neural Implicit Surfaces by Volume Rendering for Multi-view Reconstruction},
  author={Peng Wang and Lingjie Liu and Yuan Liu and Christian Theobalt and Taku Komura and Wenping Wang},
  journal={ArXiv},
  year={2021},
  volume={abs/2106.10689},
  url={https://api.semanticscholar.org/CorpusID:235490453}
}

@article{Hu2021LoRALA,
  title={LoRA: Low-Rank Adaptation of Large Language Models},
  author={J. Edward Hu and Yelong Shen and Phillip Wallis and Zeyuan Allen-Zhu and Yuanzhi Li and Shean Wang and Weizhu Chen},
  journal={ArXiv},
  year={2021},
  volume={abs/2106.09685},
  url={https://api.semanticscholar.org/CorpusID:235458009}
}

@article{Lee2025Any6DM6,
  title={Any6D: Model-free 6D Pose Estimation of Novel Objects},
  author={Taeyeop Lee and Bowen Wen and Minjun Kang and Gyuree Kang and In-So Kweon and Kuk-Jin Yoon},
  journal={ArXiv},
  year={2025},
  volume={abs/2503.18673},
  url={https://api.semanticscholar.org/CorpusID:277271765}
}

@article{Liu2025HIPPoHI,
  title={HIPPo: Harnessing Image-to-3D Priors for Model-free Zero-shot 6D Pose Estimation},
  author={Yibo Liu and Zhaodong Jiang and Binbin Xu and Guile Wu and Yuan Ren and Tongtong Cao and Bingbing Liu and Rui Heng Yang and Amir Rasouli and Jinjun Shan},
  journal={ArXiv},
  year={2025},
  volume={abs/2502.10606},
  url={https://api.semanticscholar.org/CorpusID:276408804}
}

@article{Piccinelli2024UniDepthUM,
  title={UniDepth: Universal Monocular Metric Depth Estimation},
  author={Luigi Piccinelli and Yung-Hsu Yang and Christos Sakaridis and Mattia Segu and Siyuan Li and Luc van Gool and Fisher Yu},
  journal={2024 IEEE/CVF Conference on Computer Vision and Pattern Recognition (CVPR)},
  year={2024},
  pages={10106-10116},
  url={https://api.semanticscholar.org/CorpusID:268732706}
}

@inproceedings{Guo2022AVN,
  title={A Visual Navigation Perspective for Category-Level Object Pose Estimation},
  author={Jiaxin Guo and Fangxun Zhong and Rong Xiong and Yunhui Liu and Yue Wang and Yiyi Liao},
  booktitle={European Conference on Computer Vision},
  year={2022},
  url={https://api.semanticscholar.org/CorpusID:247749068}
}

@article{Zhang2024CLAYAC,
  title={CLAY: A Controllable Large-scale Generative Model for Creating High-quality 3D Assets},
  author={Longwen Zhang and Ziyu Wang and Qixuan Zhang and Qiwei Qiu and Anqi Pang and Haoran Jiang and Wei Yang and Lan Xu and Jingyi Yu},
  journal={ACM Transactions on Graphics (TOG)},
  year={2024},
  volume={43},
  pages={1 - 20},
  url={https://api.semanticscholar.org/CorpusID:270619933}
}

@article{Xu2023DMV3DDM,
  title={DMV3D: Denoising Multi-View Diffusion using 3D Large Reconstruction Model},
  author={Yinghao Xu and Hao Tan and Fujun Luan and Sai Bi and Peng Wang and Jiahao Li and Zifan Shi and Kalyan Sunkavalli and Gordon Wetzstein and Zexiang Xu and Kai Zhang},
  journal={ArXiv},
  year={2023},
  volume={abs/2311.09217},
  url={https://api.semanticscholar.org/CorpusID:265213192}
}

@article{Meng2024LT3SDLT,
  title={LT3SD: Latent Trees for 3D Scene Diffusion},
  author={Quan Meng and Lei Li and Matthias Nie{\ss}ner and Angela Dai},
  journal={ArXiv},
  year={2024},
  volume={abs/2409.08215},
  url={https://api.semanticscholar.org/CorpusID:272600456}
}

@article{Deitke2023ObjaverseXLAU,
  title={Objaverse-XL: A Universe of 10M+ 3D Objects},
  author={Matt Deitke and Ruoshi Liu and Matthew Wallingford and Huong Ngo and Oscar Michel and Aditya Kusupati and Alan Fan and Christian Laforte and Vikram S. Voleti and Samir Yitzhak Gadre and Eli VanderBilt and Aniruddha Kembhavi and Carl Vondrick and Georgia Gkioxari and Kiana Ehsani and Ludwig Schmidt and Ali Farhadi},
  journal={ArXiv},
  year={2023},
  volume={abs/2307.05663},
  url={https://api.semanticscholar.org/CorpusID:259836993}
}

@inproceedings{Yu2025BoxDreamerDB,
  title={BoxDreamer: Dreaming Box Corners for Generalizable Object Pose Estimation},
  author={Yuanhong Yu and Xingyi He He and Chen Zhao and Junhao Yu and Jiaqi Yang and Ruizhen Hu and Yujun Shen and Xing Zhu and Xiaowei Zhou and Sida Peng},
  year={2025},
  url={https://api.semanticscholar.org/CorpusID:277667750}
}

@article{Wei2023RGBbasedCO,
  title={RGB-based Category-level Object Pose Estimation via Decoupled Metric Scale Recovery},
  author={Jiaxin Wei and Xibin Song and Weizhe Liu and Laurent Kneip and Hongdong Li and Pan Ji},
  journal={2024 IEEE International Conference on Robotics and Automation (ICRA)},
  year={2023},
  pages={2036-2042},
  url={https://api.semanticscholar.org/CorpusID:262053929}
}

@inproceedings{Lin2022CategoryLevel6O,
  title={Category-Level 6D Object Pose and Size Estimation using Self-Supervised Deep Prior Deformation Networks},
  author={Jiehong Lin and Zewei Wei and Changxing Ding and Kui Jia},
  booktitle={European Conference on Computer Vision},
  year={2022},
  url={https://api.semanticscholar.org/CorpusID:250451453}
}

@inproceedings{Fan2022ObjectLD,
  title={Object Level Depth Reconstruction for Category Level 6D Object Pose Estimation From Monocular RGB Image},
  author={Zhaoxin Fan and Zhenbo Song and Jian Xu and Zhicheng Wang and Kejian Wu and Hongyan Liu and Jun He},
  booktitle={European Conference on Computer Vision},
  year={2022},
  url={https://api.semanticscholar.org/CorpusID:247939616}
}

@article{Tian2020ShapePD,
  title={Shape Prior Deformation for Categorical 6D Object Pose and Size Estimation},
  author={Meng Tian and Marcelo H. Ang and Gim Hee Lee},
  journal={ArXiv},
  year={2020},
  volume={abs/2007.08454},
  url={https://api.semanticscholar.org/CorpusID:220546332}
}

@article{Lawson1999AchievingVO,
  title={Achieving visual object constancy across plane rotation and depth rotation.},
  author={Rebecca Lawson},
  journal={Acta psychologica},
  year={1999},
  volume={102 2-3},
  pages={
          221-45
        },
  url={https://api.semanticscholar.org/CorpusID:5024709}
}

@article{Kirillov2023SegmentA,
  title={Segment Anything},
  author={Alexander Kirillov and Eric Mintun and Nikhila Ravi and Hanzi Mao and Chlo{\'e} Rolland and Laura Gustafson and Tete Xiao and Spencer Whitehead and Alexander C. Berg and Wan-Yen Lo and Piotr Doll{\'a}r and Ross B. Girshick},
  journal={2023 IEEE/CVF International Conference on Computer Vision (ICCV)},
  year={2023},
  pages={3992-4003},
  url={https://api.semanticscholar.org/CorpusID:257952310}
}

@article{Zhang2018TheUE,
  title={The Unreasonable Effectiveness of Deep Features as a Perceptual Metric},
  author={Richard Zhang and Phillip Isola and Alexei A. Efros and Eli Shechtman and Oliver Wang},
  journal={2018 IEEE/CVF Conference on Computer Vision and Pattern Recognition},
  year={2018},
  pages={586-595},
  url={https://api.semanticscholar.org/CorpusID:4766599}
}

@article{Phongthawee2023DiffusionLightLP,
  title={DiffusionLight: Light Probes for Free by Painting a Chrome Ball},
  author={Pakkapon Phongthawee and Worameth Chinchuthakun and Nontaphat Sinsunthithet and Amit Raj and Varun Jampani and Pramook Khungurn and Supasorn Suwajanakorn},
  journal={2024 IEEE/CVF Conference on Computer Vision and Pattern Recognition (CVPR)},
  year={2023},
  pages={98-108},
  url={https://api.semanticscholar.org/CorpusID:266210050}
}

@article{Wu2025Amodal3RA3,
  title={Amodal3R: Amodal 3D Reconstruction from Occluded 2D Images},
  author={Tianhao Wu and Chuanxia Zheng and Frank Guan and Andrea Vedaldi and Tat-Jen Cham},
  journal={ArXiv},
  year={2025},
  volume={abs/2503.13439},
  url={https://api.semanticscholar.org/CorpusID:277104133}
}

@article{Oquab2023DINOv2LR,
  title={DINOv2: Learning Robust Visual Features without Supervision},
  author={Maxime Oquab and Timoth{\'e}e Darcet and Th{\'e}o Moutakanni and Huy Q. Vo and Marc Szafraniec and Vasil Khalidov and Pierre Fernandez and Daniel Haziza and Francisco Massa and Alaaeldin El-Nouby and Mahmoud Assran and Nicolas Ballas and Wojciech Galuba and Russ Howes and Po-Yao (Bernie) Huang and Shang-Wen Li and Ishan Misra and Michael G. Rabbat and Vasu Sharma and Gabriel Synnaeve and Huijiao Xu and Herv{\'e} J{\'e}gou and Julien Mairal and Patrick Labatut and Armand Joulin and Piotr Bojanowski},
  journal={ArXiv},
  year={2023},
  volume={abs/2304.07193},
  url={https://api.semanticscholar.org/CorpusID:258170077}
}

@article{Jin2025Oneshot3O,
  title={One-shot 3D Object Canonicalization based on Geometric and Semantic Consistency},
  author={Li Jin and Yujie Wang and Wenzheng Chen and Qiyu Dai and Qingzhe Gao and Xueying Qin and Baoquan Chen},
  journal={2025 IEEE/CVF Conference on Computer Vision and Pattern Recognition (CVPR)},
  year={2025},
  pages={16850-16859},
  url={https://api.semanticscholar.org/CorpusID:280058793}
}

@STRING{CVPR = {Proc. IEEE Conf. on Computer Vision and Pattern	Recognition (CVPR)}}

@STRING{ICCV = {Proc. of the IEEE International Conf. on Computer Vision (ICCV)}}

@STRING{ICML = {Proc. of the International Conf. on Machine learning (ICML)}}

@STRING{TOG = {ACM Trans. on Graphics}}

@STRING{ICRA = {Proc. IEEE International Conf. on Robotics and Automation (ICRA)}}

@STRING{SS = {Statistical Science}}

@STRING{COMPUTER = {IEEE Computer}}

@STRING{ACM = {Communications of the ACM}}

@STRING{ARXIV = {arXiv.org}}

@STRING{NETWORKS = {Networks}}
}

\newpage
\appendix
\noindent{\Large{\textbf{Appendix}}}

In this appendix, we detail our dataset curation, experiment implementations, and method implementations in \secref{sec:implentation_details}. Besides, we provide more results of our efficient arrow-based object manipulation, orientation-aligned object generation, orientation estimation, and failure cases in \secref{sec:more_results}.

\section{Implementation Details} \label{sec:implentation_details}

\subsection{Dataset Curation} \label{sec:dataset_curation}

\noindent \textbf{Data Processing Pipeline. } We first apply Vision Language Model (VLM) recognition across the entire dataset, followed by manual identification and correction of failure cases. Specifically, four images were rendered for each 3D object using an orthogonal camera setup. The front view was identified using the VLM (Gemini 2.0~\cite{gemini}), and the objects were subsequently rotated to their canonical poses based on the recognition results. Note that objects recognized as without front-view orientation are automatically excluded from the dataset. After this initial step, the orientations of the processed objects were manually refined using Blender software. 

\begin{figure}[htbp]
    \centering
    \includegraphics[width=\linewidth]{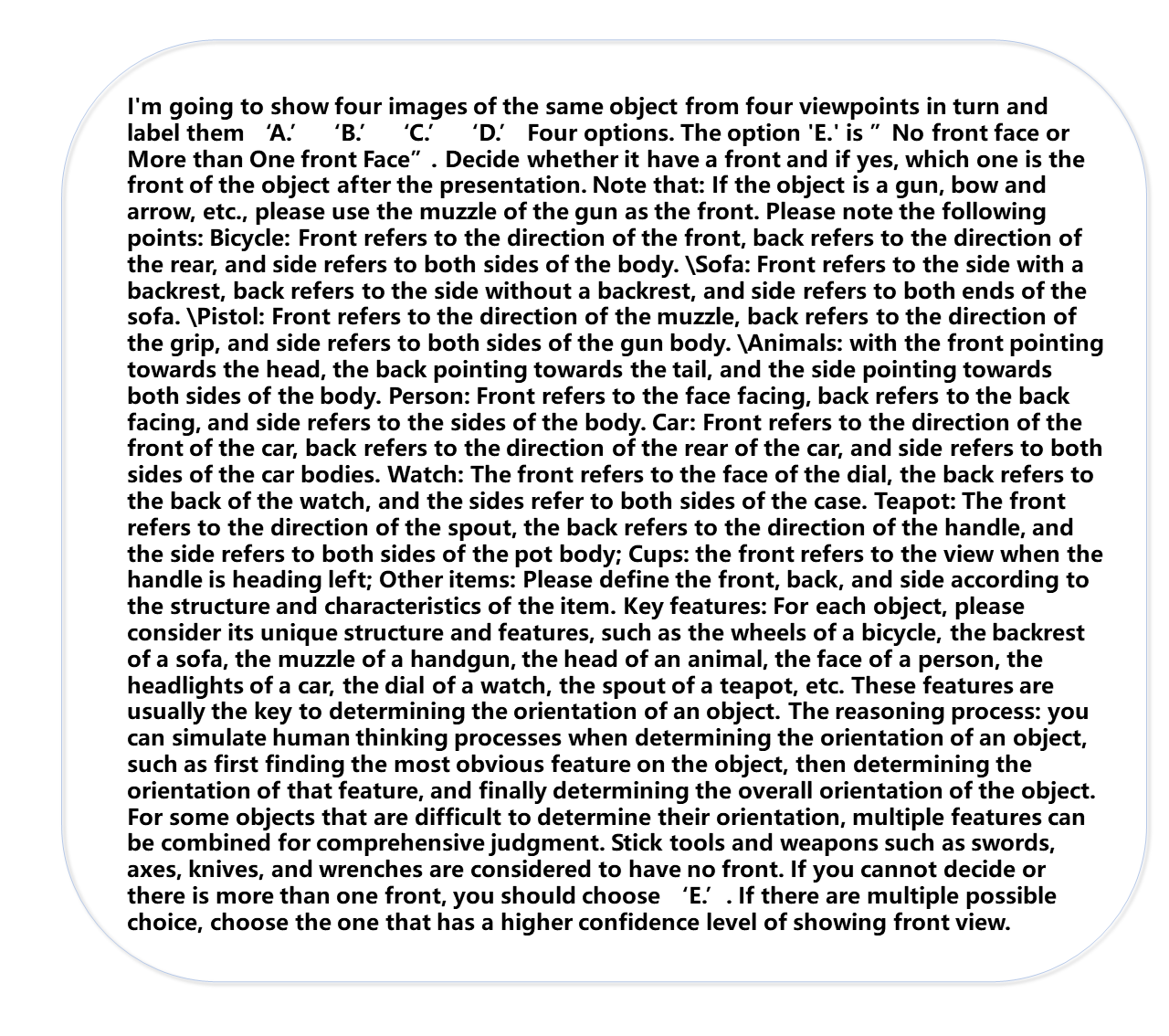}
    \caption{Text prompt for the Vision Language Model.}
    \label{fig:vlm_prompt}
\end{figure}

\noindent \textbf{VLM Pre-processing. } Different from Orient Anything, we further enhance the text prompt by adding more recognition rules for different kinds of objects, as shown in \figref{fig:vlm_prompt}. We find this operation can further improve the robustness of VLM via in-context learning.

\begin{figure}[h]
    \centering
    \includegraphics[width=0.95\linewidth]{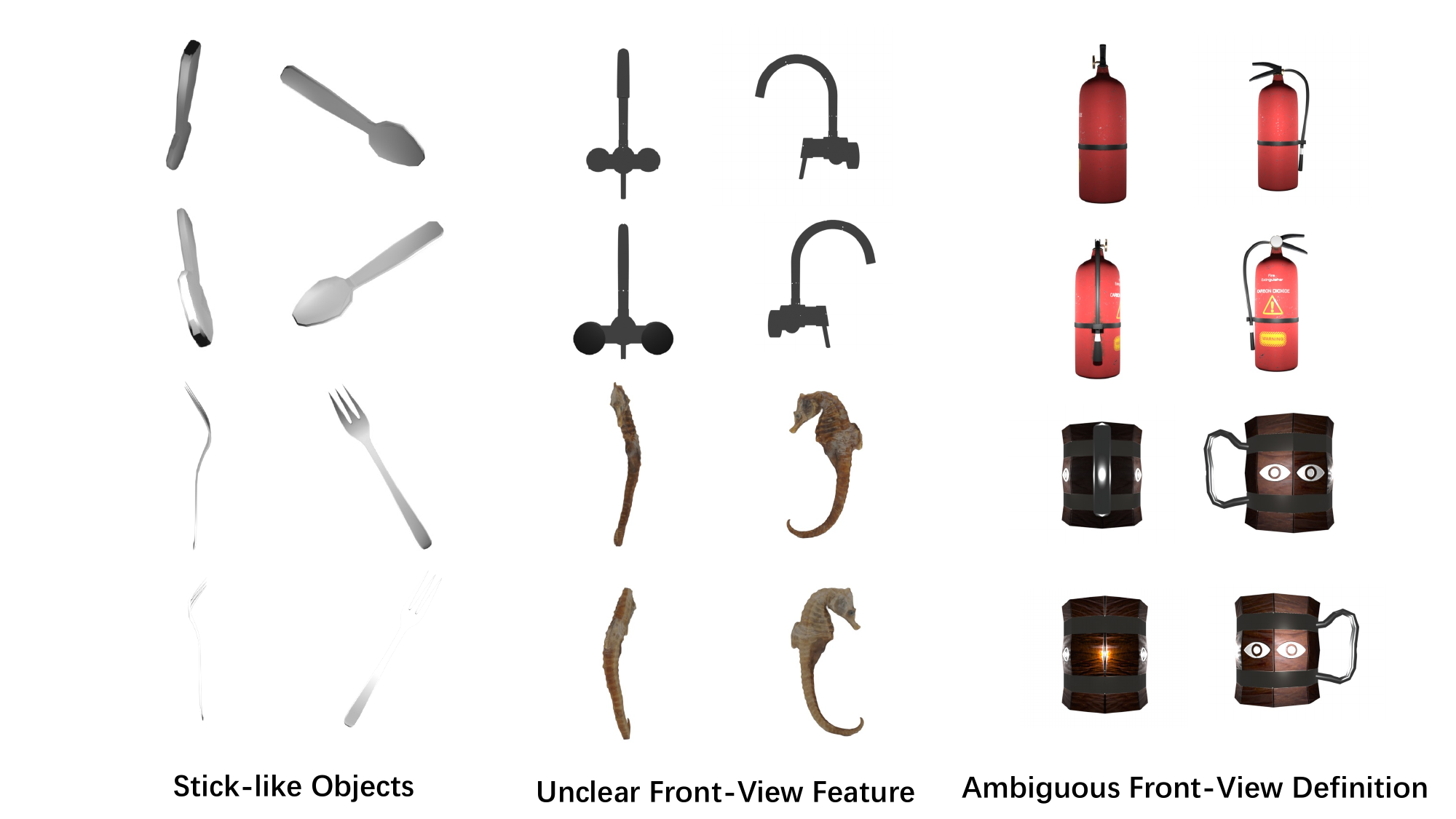}
    \caption{Failure cases in VLM-based front-view recognition. The four images are rendered from four orthogonal cameras following our VLM-based recognition method.}
    \label{fig:vlm_failure}
\end{figure}

\begin{figure}[htbp]
    \centering
    \centering
    \begin{minipage}[b]{0.8\linewidth}
        \centering
        \includegraphics[width=\linewidth]{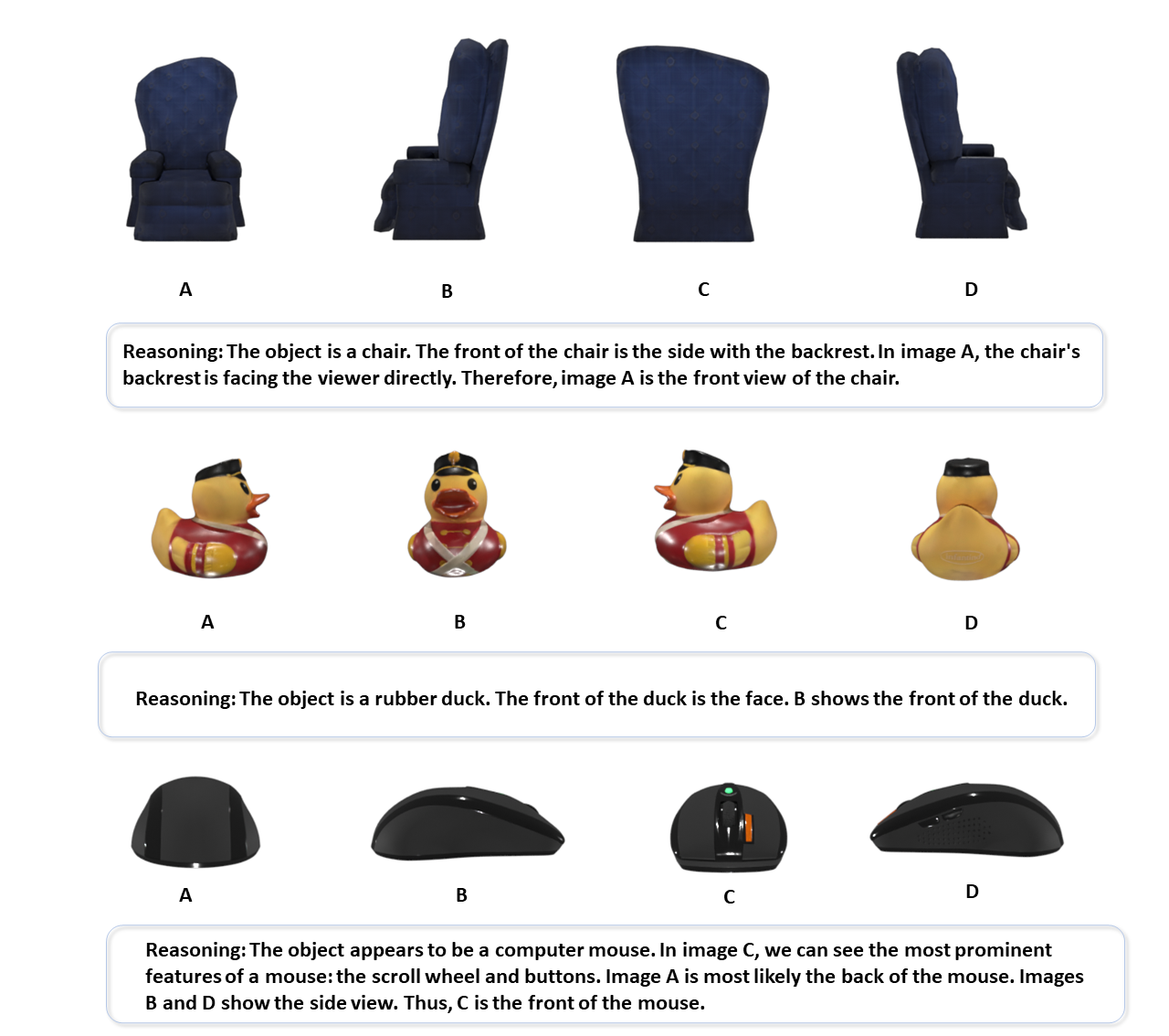}
        \caption{Correct VLM recognition results.}
        \label{fig:vlm_correct}
    \end{minipage}
    \vspace{1em}
    \begin{minipage}[b]{0.8\linewidth}
        \centering
        \includegraphics[width=\linewidth]{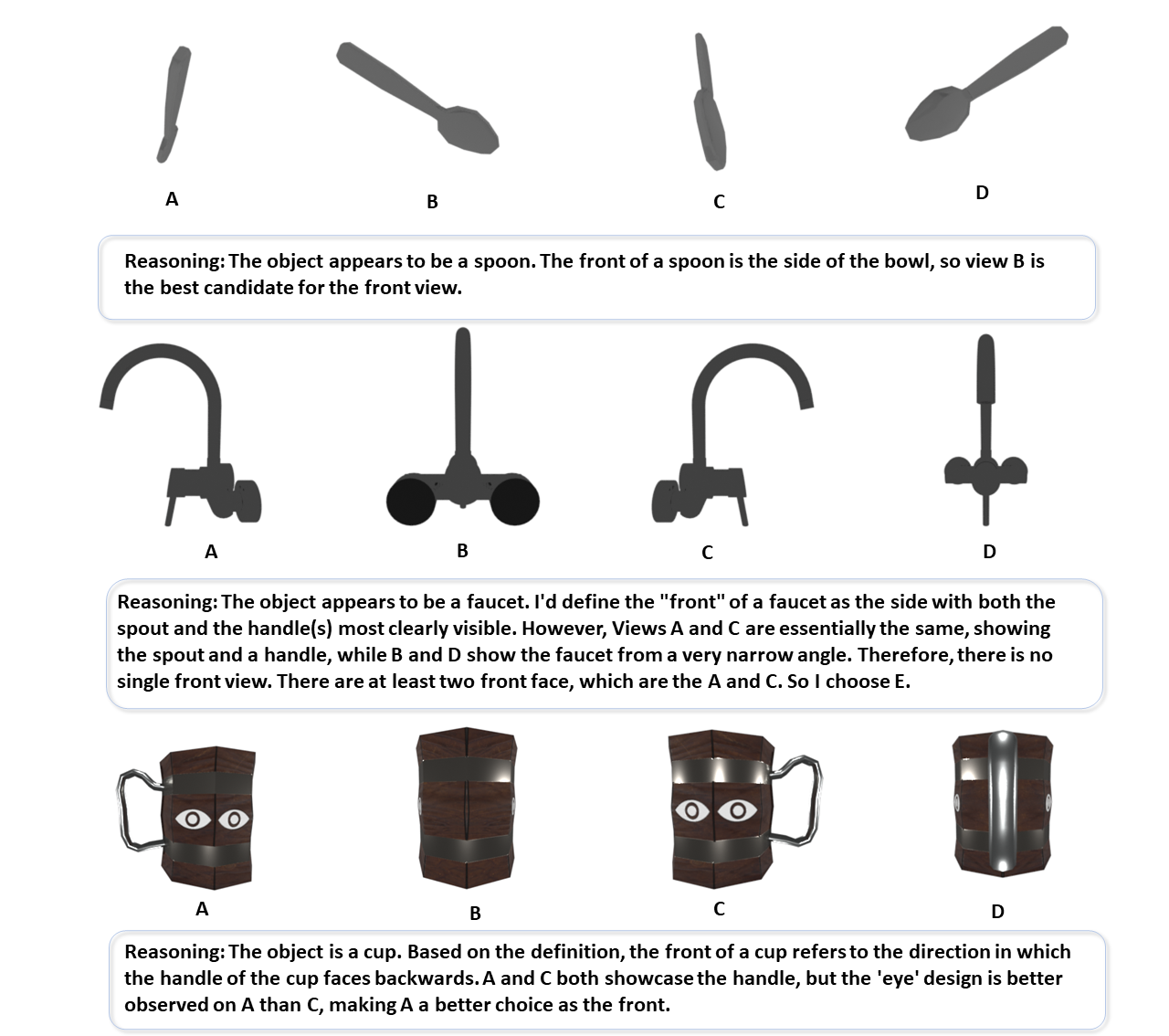}
        \caption{Wrong VLM recognition results.}
        \label{fig:vlm_wrong}
    \end{minipage}
\end{figure}

\noindent \textbf{Manual Corrrection Details. } Despite advancements, VLMs still face challenges in accurately recognizing objects' orientations due to a lack of training. Three particular object categories are prone to errors, as shown in \figref{fig:vlm_failure}: (1) stick-like objects (e.g., forks, spoons), (2) objects lacking distinctive front-view features (e.g., seahorses, water faucets), and (3) objects with inherently ambiguous front-view definitions (e.g., mugs, fire extinguishers). VLMs struggle with stick-like objects since they are not aligned in roll and pitch angles. For objects without clear front-view features, distinguishing the front from the back often requires side-view inference, which hasn't been achieved by existing VLMs. Ambiguity in defining a canonical front view further complicates recognition for certain object types. To address these issues, we implemented the following corrections: stick-like objects were aligned parallel to the global X-axis, positioning the handle toward the negative X-axis and the functional end toward the positive X-axis. Objects without distinct front-view cues were manually rotated in Blender. For objects with ambiguous front-view definitions, we aligned the component typically associated with human interaction (e.g., handles) to the negative X-axis.

\noindent \textbf{VLM Error Analysis} After obtaining the orientation-aligned dataset, we conducted a comprehensive error analysis by comparing the VLM's predictions with our manually validated ground truth. For each 3D model, we computed the Chamfer Distance (CD) between the 3D models after VLM-based recognition and our manually corrected 3D models. Each model was uniformly sampled with 10,000 surface points to facilitate CD computation.
Since CD values are influenced by object geometry and do not perfectly reflect orientation accuracy, we adopted a threshold-based approach to quantify recognition accuracy. Through empirical analysis, we found that rotating a model by 90° around its principal axis typically resulted in a CD > 0.01. This threshold, $\gamma = 0.01$, was therefore used to flag recognition errors: any model with a CD exceeding this value between VLM and ground-truth orientations was considered wrongly recognized.
Notably, this threshold may produce wrong results for perfectly cylindrical or cubical shapes due to their inherent symmetry. However, such cases represent a negligible minority in our dataset and were therefore excluded from our error statistics to maintain analytical integrity.

\subsection{Experiment Implementations} \label{sec:exp_implementation}

\noindent \textbf{Evaluation Data. } We evaluate our method on three unseen datasets: GSO~\cite{Downs2022GoogleSO}, Toys4k~\cite{Stojanov2021UsingST}, and Imagenet3D~\cite{Ma2024ImageNet3DTG}. For the GSO dataset, we randomly collected 48 objects with clearly defined front views. Each object was rendered into four images, sampled from the upper hemisphere with azimuth angles uniformly distributed within [0$^\circ$, 360$^\circ$], polar angles within [0$^\circ$, 60$^\circ$], and rotation angles within [-30$^\circ$, 30$^\circ$]. For the Toys4k dataset, we randomly selected 439 objects across 47 categories. Each object was rendered into a single image using the same camera configuration as applied to the GSO dataset. For the Imagenet3D dataset, we focused on seven stick-like object categories, comprising a total of 204 objects. Please refer to \tabref{tab:obj_num_toys4k} and \tabref{tab:obj_num_imagenet3d} for details.

\begin{table}[h]
\centering
\begin{tabular}{c|c|c|c|c|c|c|c}
\toprule
airplane & bicycle & boat & bunny & bus & car & cat & chair \\
\midrule
9 & 10 & 9 & 9 & 5 & 5 & 9 & 20 \\
\midrule
chicken & cow & crab & deer moose & dinosaur & dog & dolphin & dragon \\
\midrule
8 & 6 & 10 & 10 & 10 & 9 & 10 & 10 \\
\midrule
elephant & fish & fox & frog & giraffe & guitar & helmet & helicopter \\
\midrule
10 & 10 & 9 & 10 & 10 & 10 & 7 & 8 \\
\midrule
horse & laptop & lion & lizard & monkey & motorcycle & mouse & panda \\
\midrule
10 & 6 & 10 & 10 & 10 & 8 & 10 & 7 \\
\midrule
PC mouse & penguin & piano & pig & radio & robot & shark & sheep \\
\midrule
3 & 16 & 6 & 10 & 2 & 8 & 10 & 8 \\
\midrule
shoe & sofa & tractor & train & truck & violin & whale &  \\
\midrule
10 & 20 & 16 & 4 & 10 & 10 & 9 &  \\
\bottomrule
\end{tabular}
\vspace{1em}
\captionof{table}{Object numbers of each category in Toys4k~\cite{Stojanov2021UsingST} evaluation data.}
\label{tab:obj_num_toys4k}
\end{table}

\begin{table}[h]
\centering
\begin{tabular}{c|c|c|c|c|c|c}
\toprule
fork & knife & pen & rifle & scissors & screwdriver & spoon\\
\midrule
22 & 24 & 30 & 31 & 33 & 29 & 38\\
\bottomrule
\end{tabular}
\vspace{1em}
\captionof{table}{Object numbers of each category in Imagenet3D~\cite{Ma2024ImageNet3DTG} evaluation data.}
\label{tab:obj_num_imagenet3d}
\end{table}

\noindent \textbf{Baselines. } To implement the PCA-based baseline, we first calculate the three principal axes by eigen-decomposing the 3D models' geometries. The object is then rotated to align these principal axes with the global coordinate system's x-, y-, and z-axes. For the VLM-based baseline, we follow the same method used in the dataset curation, utilizing Gemini-2.0~\cite{gemini} for orientation recognition. To implement the baseline based on Orient Anything~\cite{Wang2024OrientAL}, we use the official checkpoint based on the ViT-large architecture and adopt its data augmentation module. The object orientation is estimated from an image rendered using a fixed camera, and the object is subsequently rotated according to the predicted orientation. For the pre-trained 3D generative models based on multi-view diffusion and 3D-VAE, we utilize the official Wonder3D++ and Trellis checkpoints. For the orientation estimation baseline based on FSDetView~\cite{Xiao2020FewShotOD}, we restrict evaluation to the object categories supported by the method.

\noindent \textbf{Metrics. }For the calculation of Chamfer Distance, we don't perform PCA for both predicted 3D models and GT models (except the PCA baseline), in order to measure the pose canonicalization performance. As for the calculation of CLIP and LPIPS, for the 3D-VAE backbone, we render four orthogonal views with the camera elevation angle of 0 for both the generated and GT 3D models, and computed the LPIPS and CLIP scores based on these renderings at matched camera poses, while for the Multi-view Diffusion backbone, we randomly sample views on a unit sphere for evaluation to avoid unfairness, since the orthogonal views are aligned with the camera setting of our Wonder3D-OA, but not aligned with the Wonder3D baselines.

\subsection{Method Implementations} \label{sec:method_implementation}

\noindent \textbf{Efficient Arrow-based Object Rotation Manipulation. }
For the arrow-based object rotation manipulation in the augmented reality application, we define this arrow-based interaction using a 2D start point $\bm{\mathcal{P}}_{start}$ and a 2D end point $\bm{\mathcal{P}}_{end}$, forming a direction vector $\bm{\mathcal{{V}}}_{target} = \bm{\mathcal{P}}_{end} - \bm{\mathcal{P}}_{start}$. Then we use Unidepth~\cite{Piccinelli2024UniDepthUM} to estimate the camera intrinsic and depth map. With camera intrinsic, we calculate the 3D camera rays $\bm{\mathcal{R}}_{start}$ and $\bm{\mathcal{R}}_{end}$ for the corresponding 2D points $\bm{\mathcal{P}}_{start}$ and $\bm{\mathcal{P}}_{end}$. With the depth map, we calculate the plane $\bm{\mathcal{P}}$ via the least square method. After that, we calculate the 3D intersection points $\bm{\mathcal{P}}_{end}^{3d}$ and $\bm{\mathcal{P}}_{start}^{3d}$ between the corresponding camera rays $\bm{\mathcal{R}}_{start}$, $\bm{\mathcal{R}}_{end}$ and the extracted plane $\bm{\mathcal{P}}$. Finally, 3D direction vector $\bm{\mathcal{{V}}}_{target}^{3d}$ can be calculated: $\bm{\mathcal{{V}}}_{target}^{3d} = \bm{\mathcal{P}}_{end}^{3d} - \bm{\mathcal{P}}_{start}^{3d}$. Since each generated model is aligned to a known forward vector $\bm{\mathcal{V}}_{init}^{3d}$, we can directly compute and apply the rotation that aligns $\bm{\mathcal{V}}_{init}^{3d}$ with $\bm{\mathcal{{V}}}_{target}^{3d}$, allowing the object to be placed in the correct orientation without any manual adjustment. Note that the 3D start point $\bm{\mathcal{P}}_{start}^{3d}$ is also the 3D location of the inserted 3D object, the 3D models need to rotate along $\bm{\mathcal{{V}}}_{target}^{3d}$ to ensure verticality to the ground and the object size still needs to be set via VLM or user interaction, since our generated models share a normalized scale. Specifically, to automatically estimate the size of the object to insert, we can directly ask the VLM to estimate the common size of the object in the real world with the single-view image as input, since our method estimates metric depth with the scale aligned with the real world for insertion. Besides, to make the insertion results realistic, we set a plane as a shadow catcher and estimate the environment lighting using DiffusionLight~\cite{Phongthawee2023DiffusionLightLP}. For the arrow-based object rotation manipulation in the generic 3D software, we define this arrow-based interaction using a 3D start point $\bm{\mathcal{P}}_{start}^{3d}$ and a 3D end point $\bm{\mathcal{P}}_{end}^{3d}$, forming a 3D direction vector $\bm{\mathcal{{V}}}_{target}^{3d} = \bm{\mathcal{P}}_{end}^{3d} - \bm{\mathcal{P}}_{start}^{3d}$. The rotation transformation is calculated by aligning $\bm{\mathcal{V}}_{init}^{3d}$ with $\bm{\mathcal{{V}}}_{target}^{3d}$ using Rodrigues' formula and users can further rotate the objects along $\bm{\mathcal{{V}}}_{target}^{3d}$ if needed.

\section{More Results} \label{sec:more_results}

\subsection{Efficient Arrow-based Object Manipulation}

\begin{figure}
    \centering
    \includegraphics[width=0.98\linewidth]{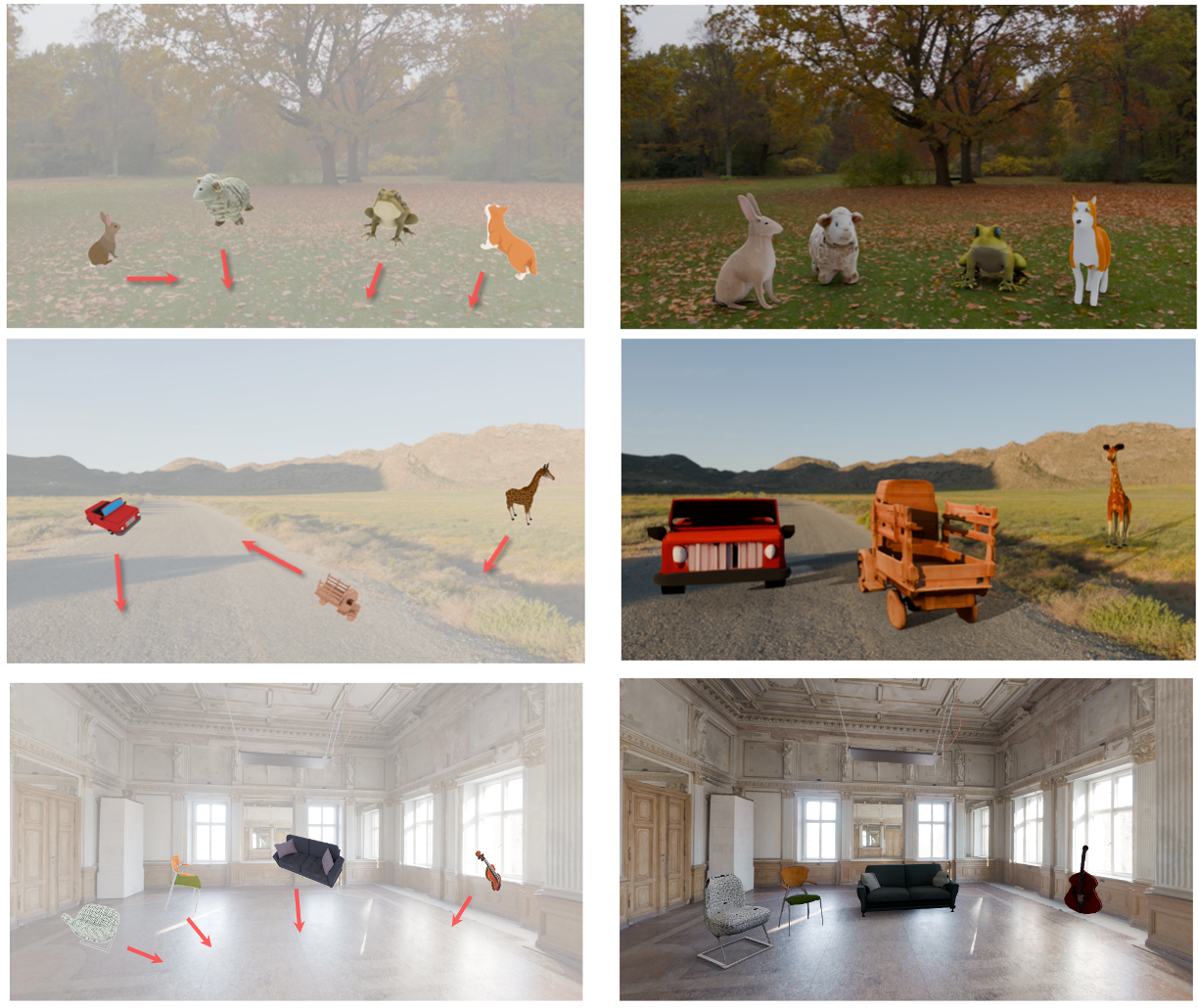}
    \caption{More results of our efficient arrow-based object manipulation method.}
    \label{fig:more_insertion}
\end{figure}

We present more results of our efficient arrow-based object manipulation method in the \figref{fig:more_insertion}.

\begin{figure}[htbp]
    \centering
    \includegraphics[width=\linewidth]{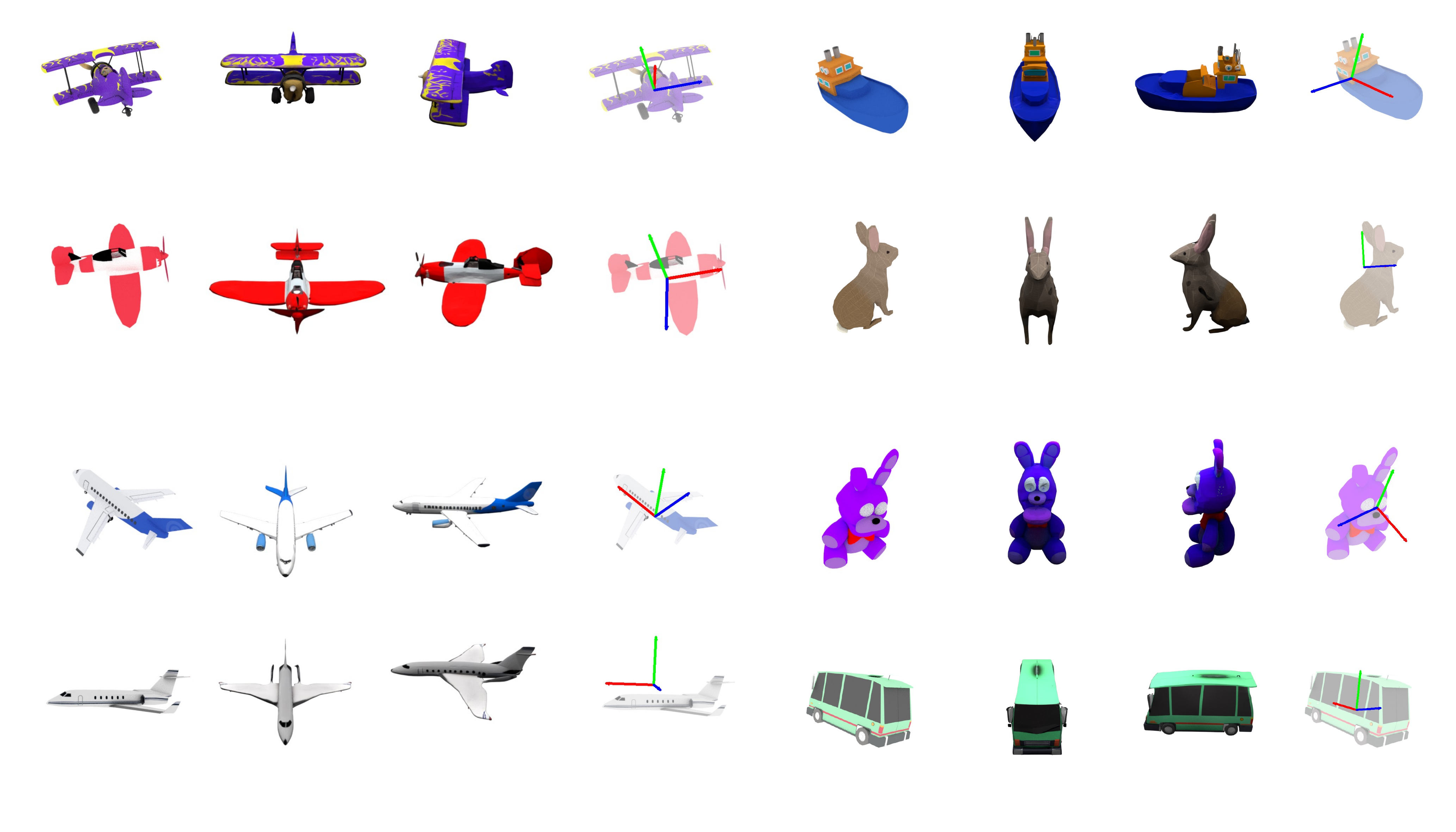}
    \caption{More qualitative results of our Trellis-OA and orientation estimation on Toys4k~\cite{Stojanov2021UsingST}. Note that the images correspond to the input image, two renderings of the generated 3D model, and the orientation estimation results, respectively.}
    \label{fig:more_results_1}
\end{figure}

\begin{figure}[htbp]
    \centering
    \includegraphics[width=0.95\linewidth]{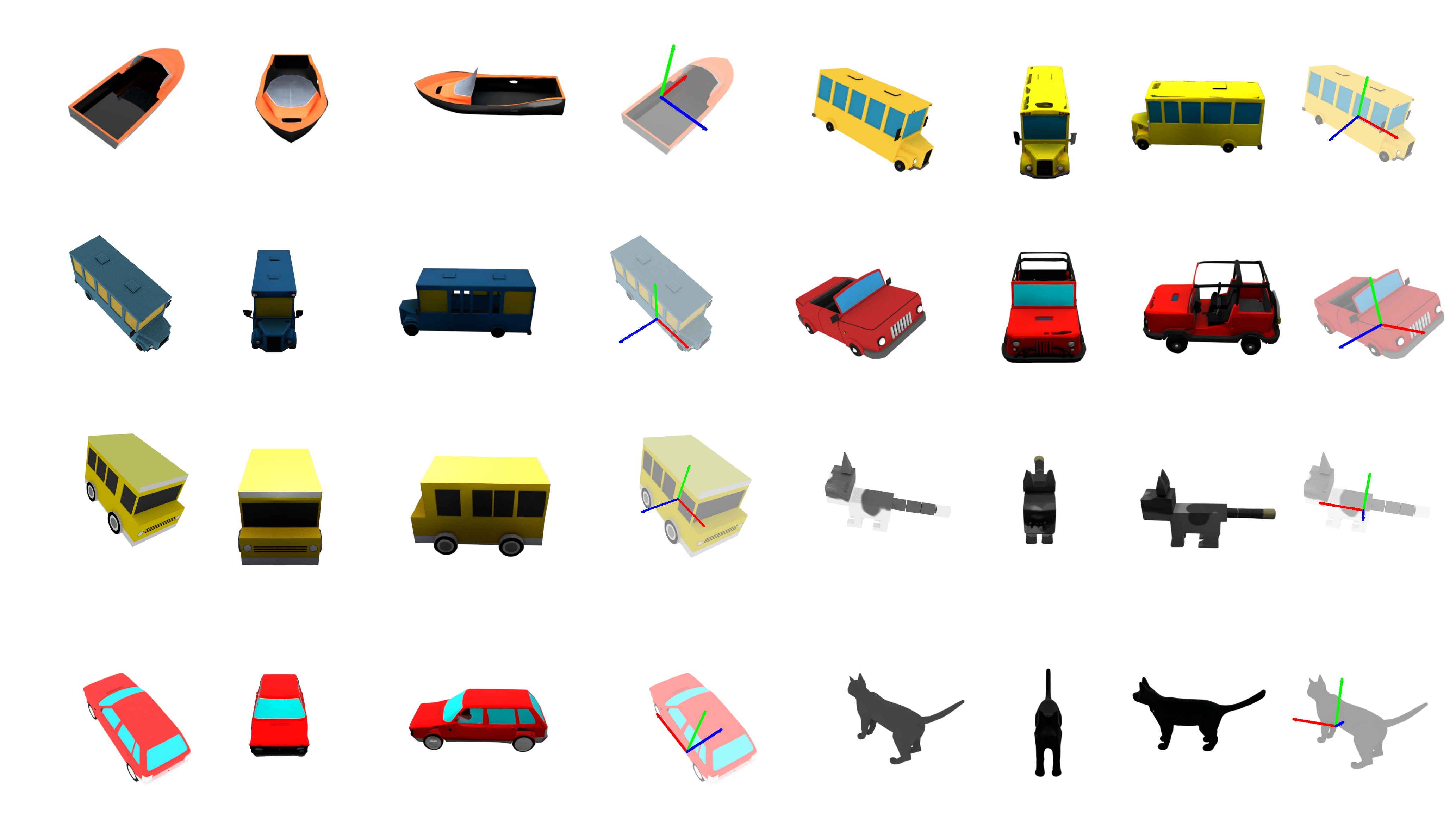}\\
    \vspace{-1em}
    \includegraphics[width=0.95\linewidth]{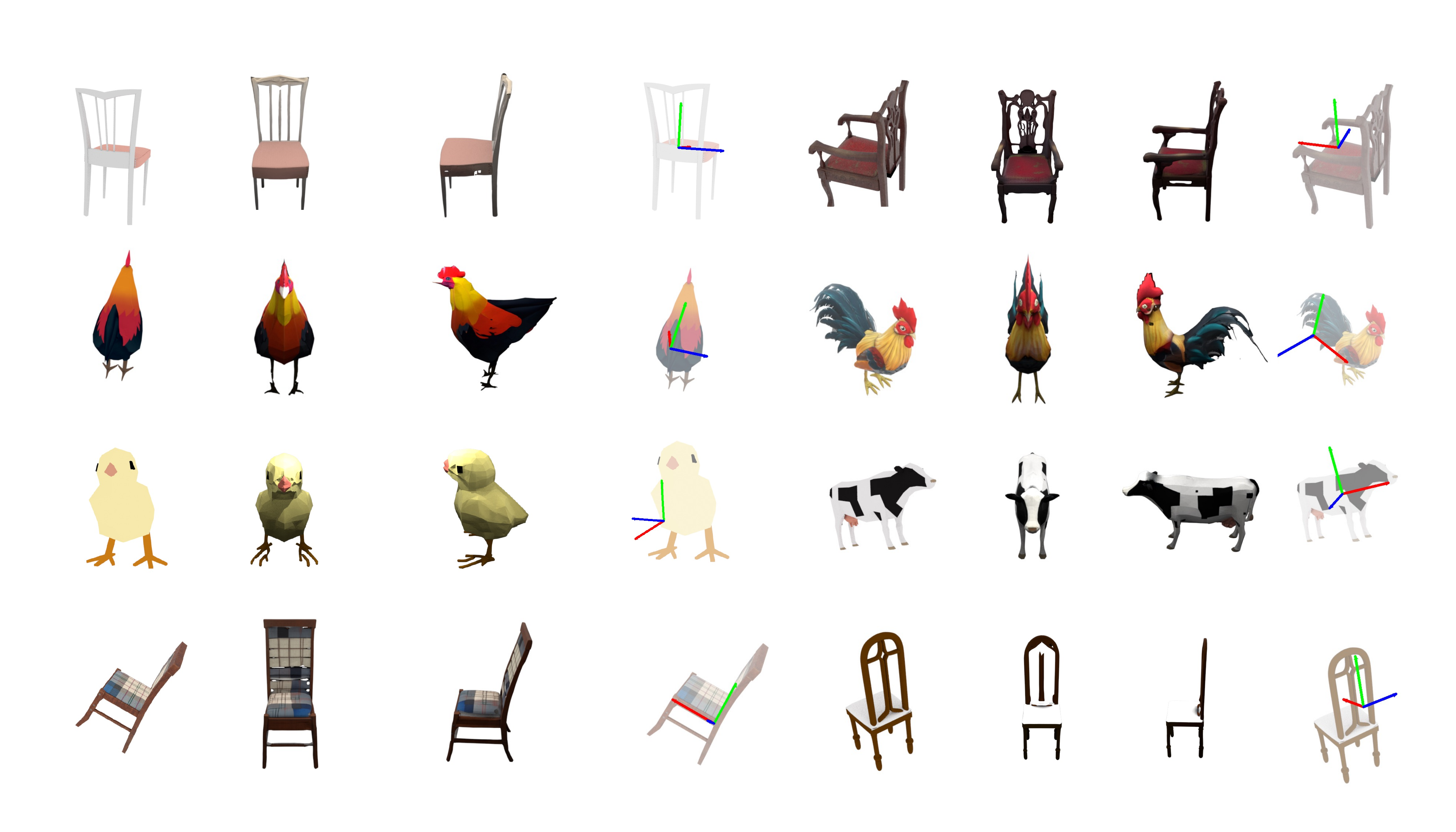} \\
    \vspace{-1em}
    \includegraphics[width=0.95\linewidth]{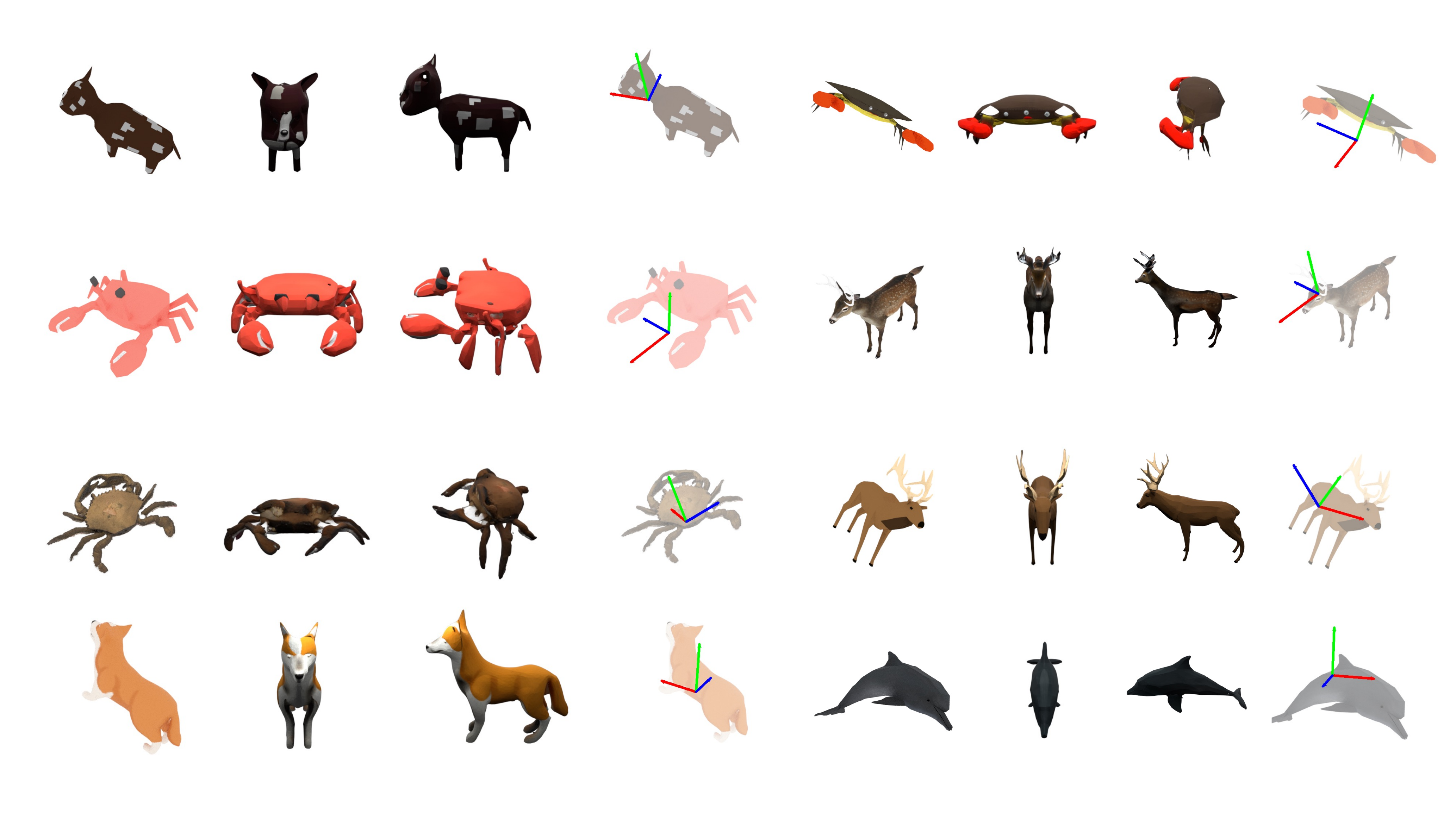}
    \caption{More qualitative results of our Trellis-OA and orientation estimation on Toys4k~\cite{Stojanov2021UsingST}.}
    \label{fig:more_results_2}
\end{figure}

\begin{figure}[htbp]
    \centering
    \includegraphics[width=0.95\linewidth]{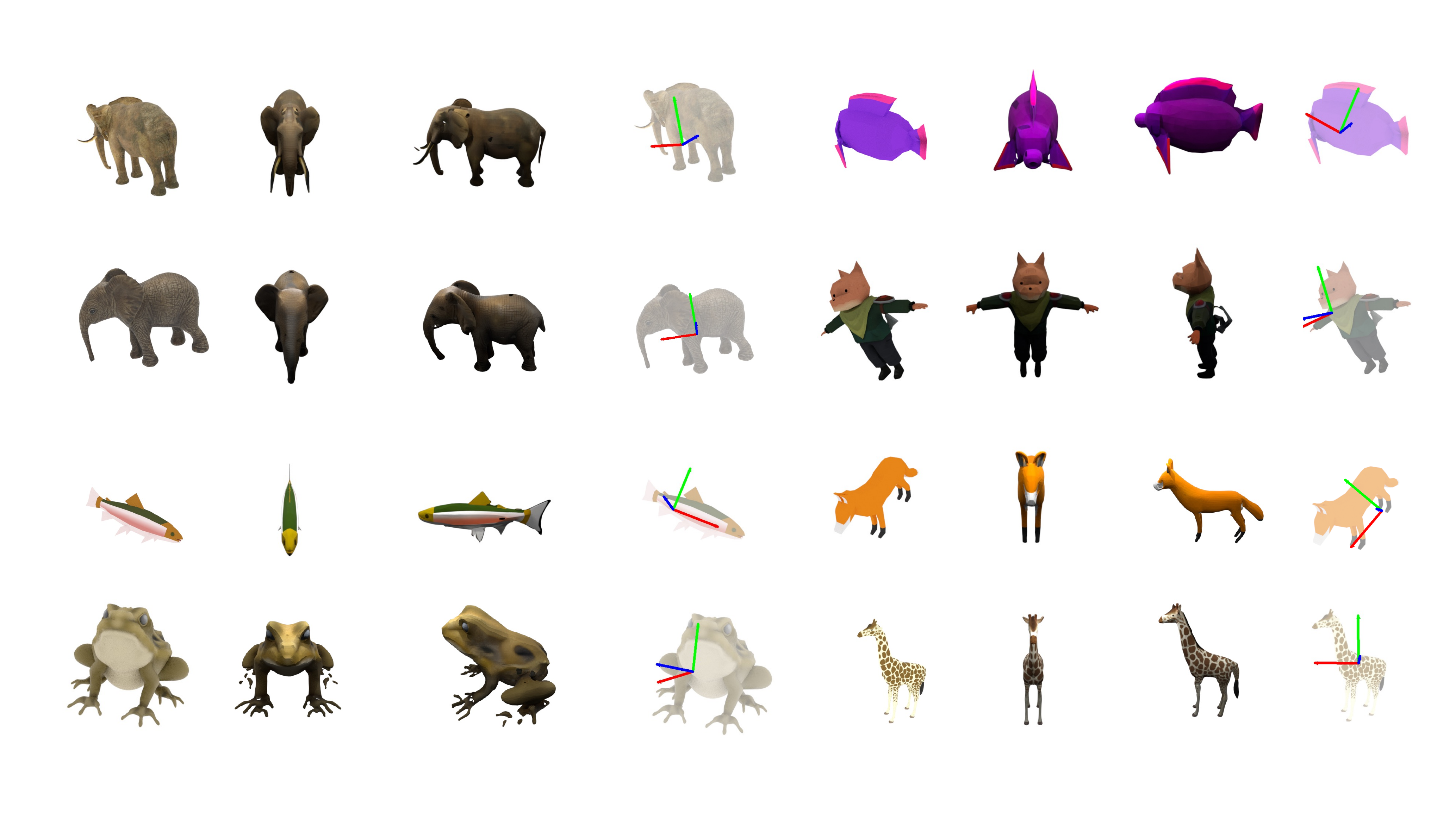}\\
    \vspace{-1em}
    \includegraphics[width=0.95\linewidth]{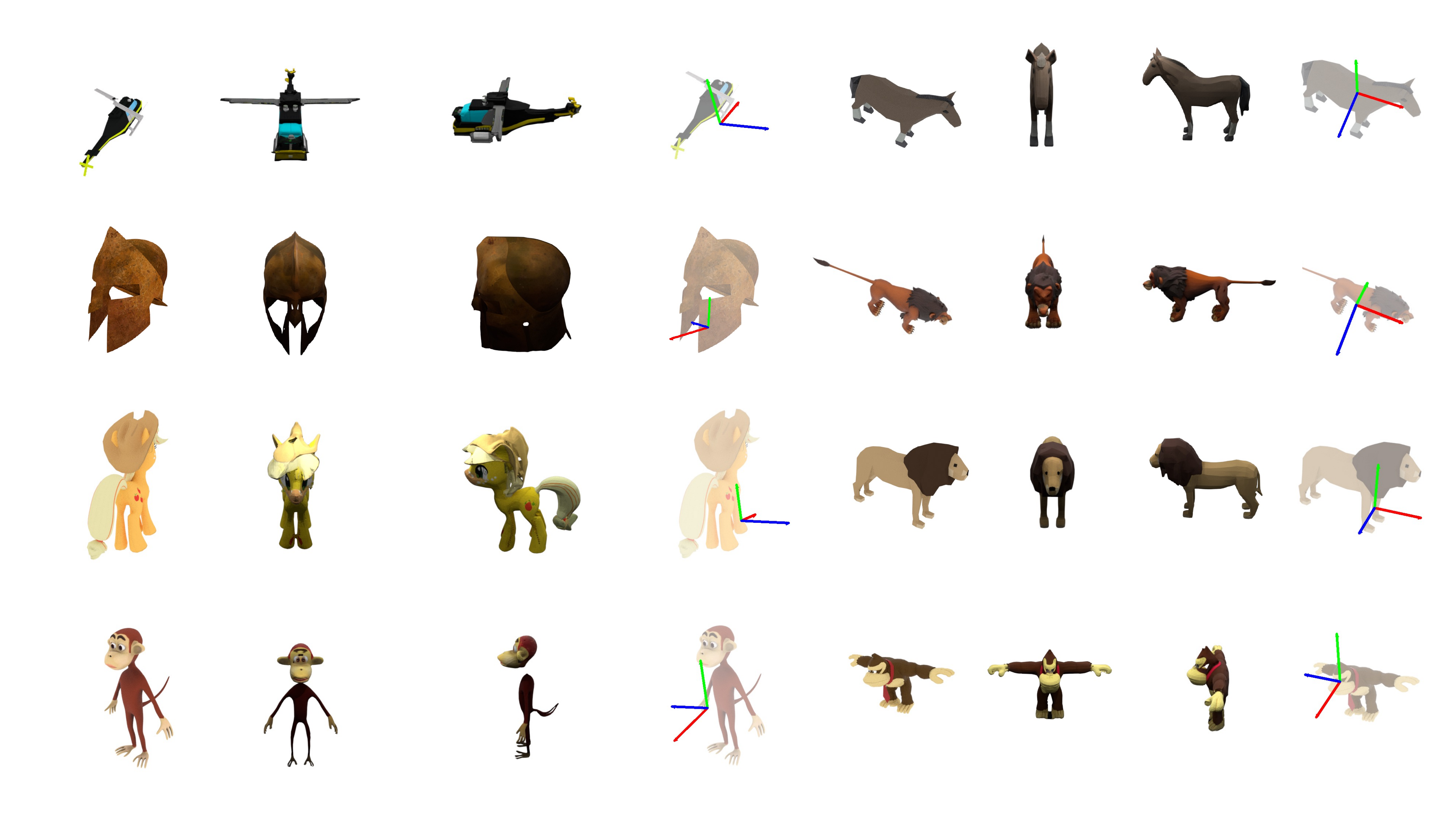} \\
    \vspace{-1em}
    \includegraphics[width=0.95\linewidth]{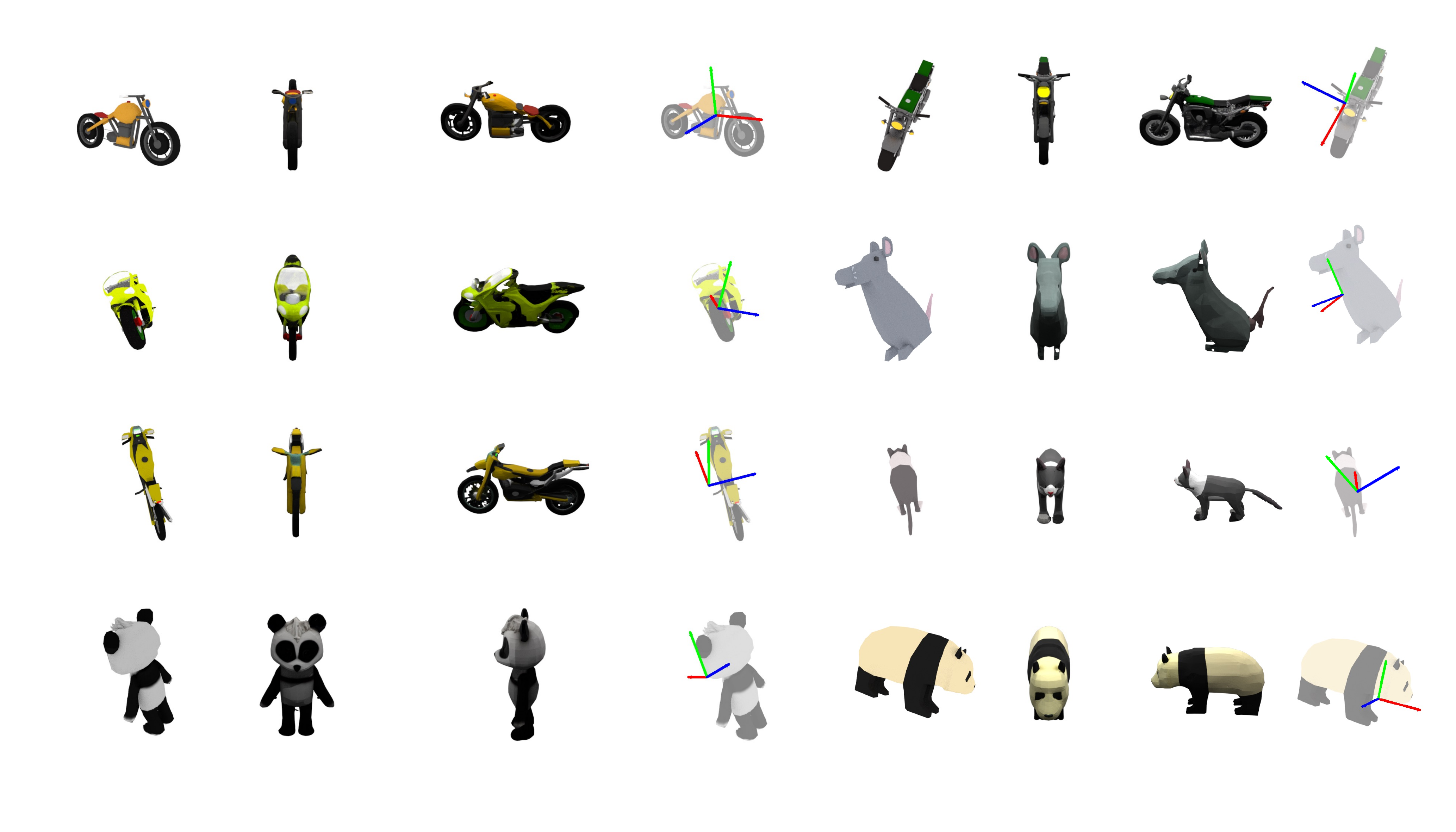}
    \caption{More qualitative results of our Trellis-OA and orientation estimation on Toys4k~\cite{Stojanov2021UsingST}.}
    \label{fig:more_results_3}
\end{figure}

\begin{figure}[htbp]
    \centering
    \includegraphics[width=0.95\linewidth]{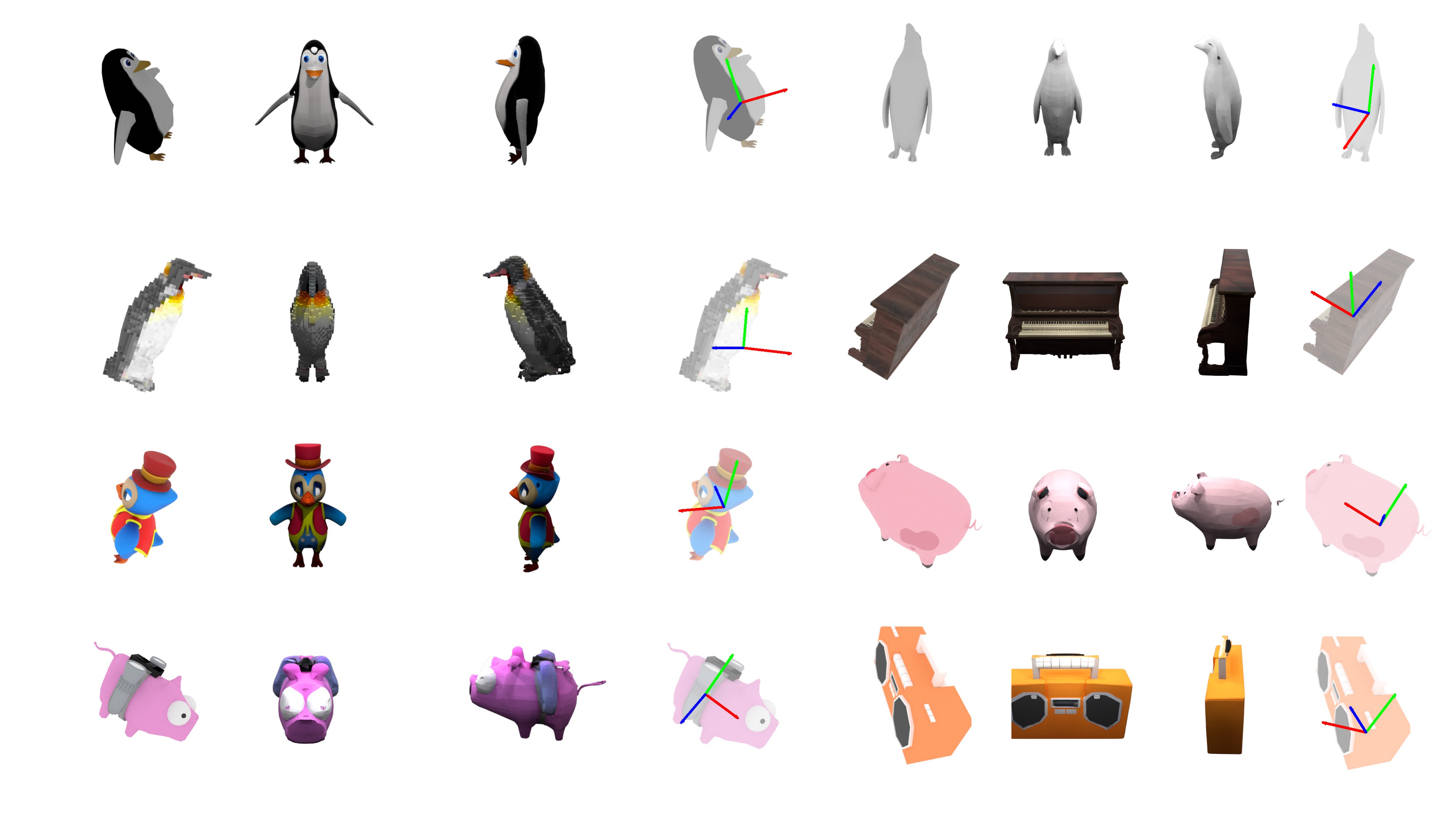}\\
    \vspace{-1em}
    \includegraphics[width=0.95\linewidth]{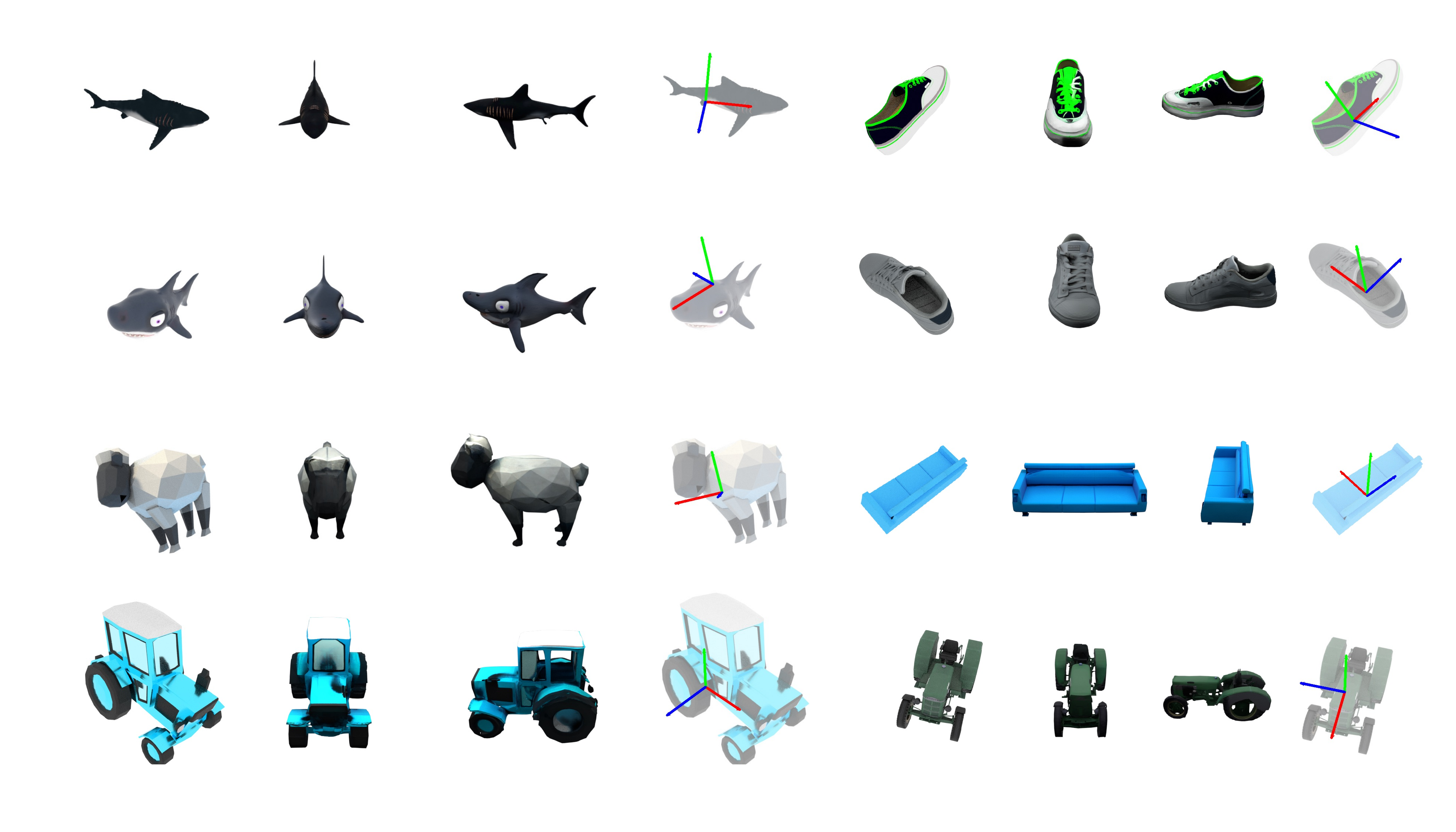} \\
    \vspace{-1em}
    \includegraphics[width=0.95\linewidth]{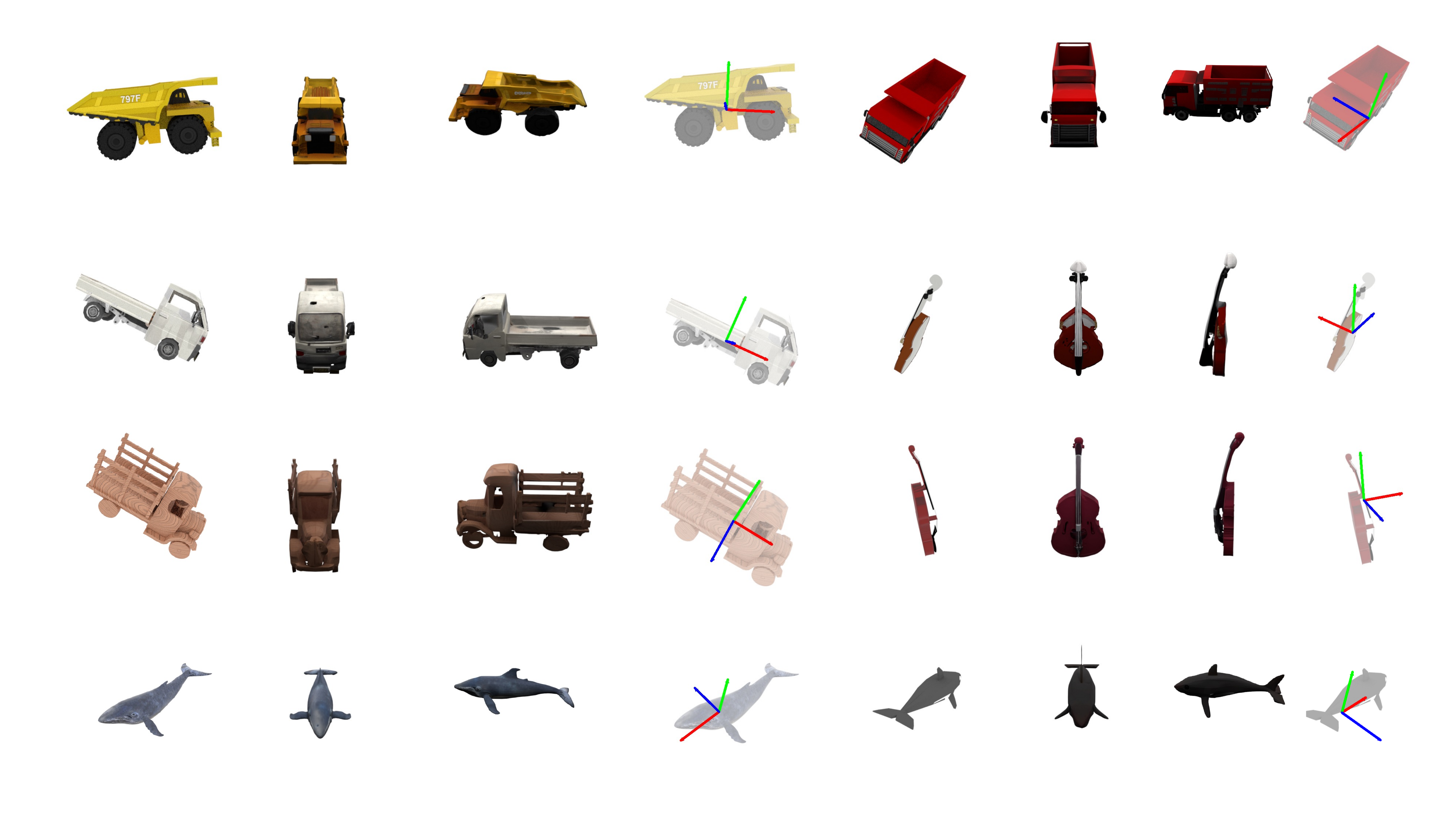}
    \caption{More qualitative results of our Trellis-OA and orientation estimation on Toys4k~\cite{Stojanov2021UsingST}.}
    \label{fig:more_results_4}
\end{figure}

\begin{figure}[htbp]
    \centering
    \includegraphics[width=0.95\linewidth]{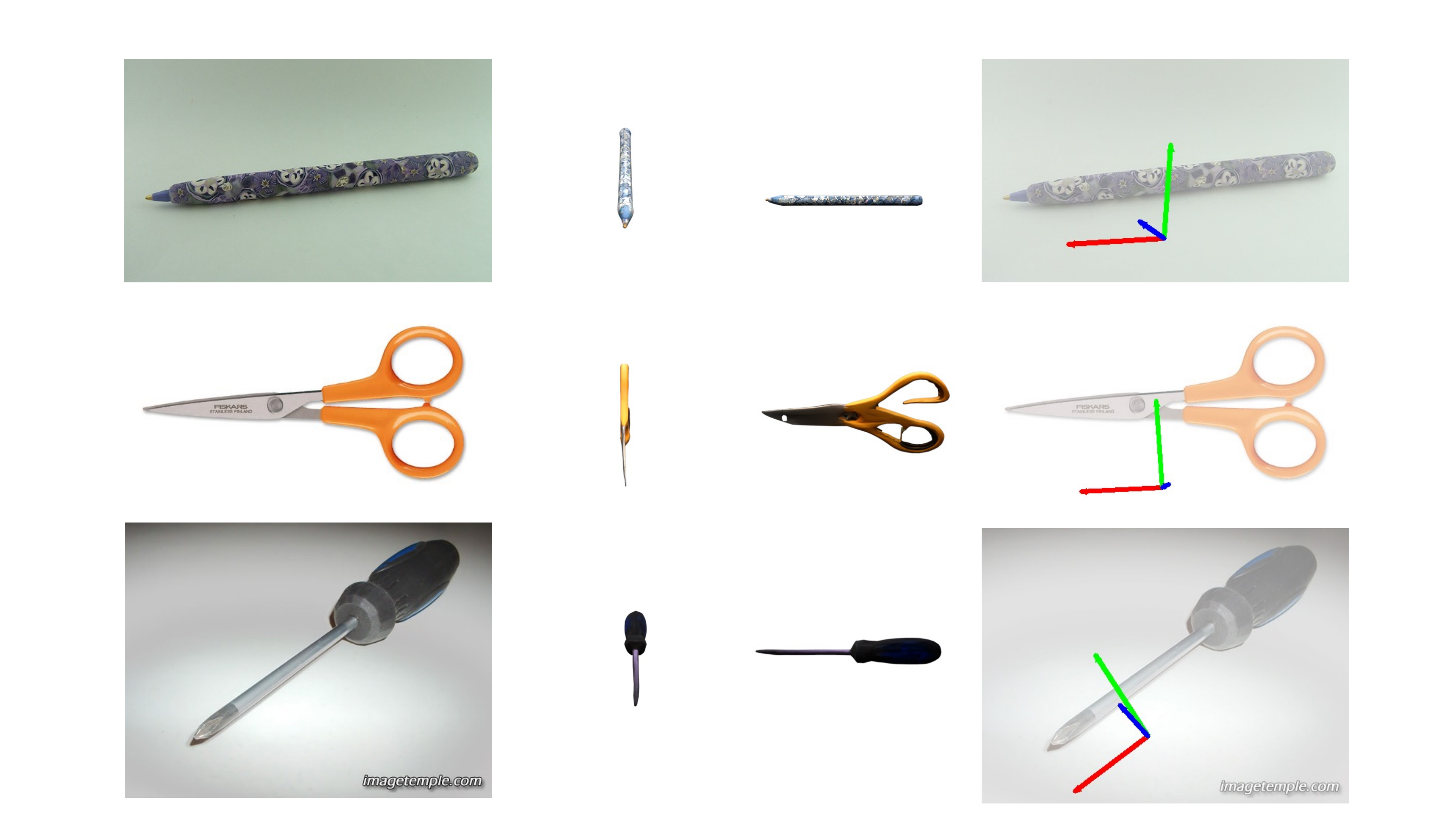}
    \caption{More qualitative results of our Trellis-OA and orientation estimation on Imagenet3D~\cite{Ma2024ImageNet3DTG} dataset.}
    \label{fig:more_results_5}
\end{figure}

\begin{figure}
    \centering
    \includegraphics[width=0.99\linewidth]{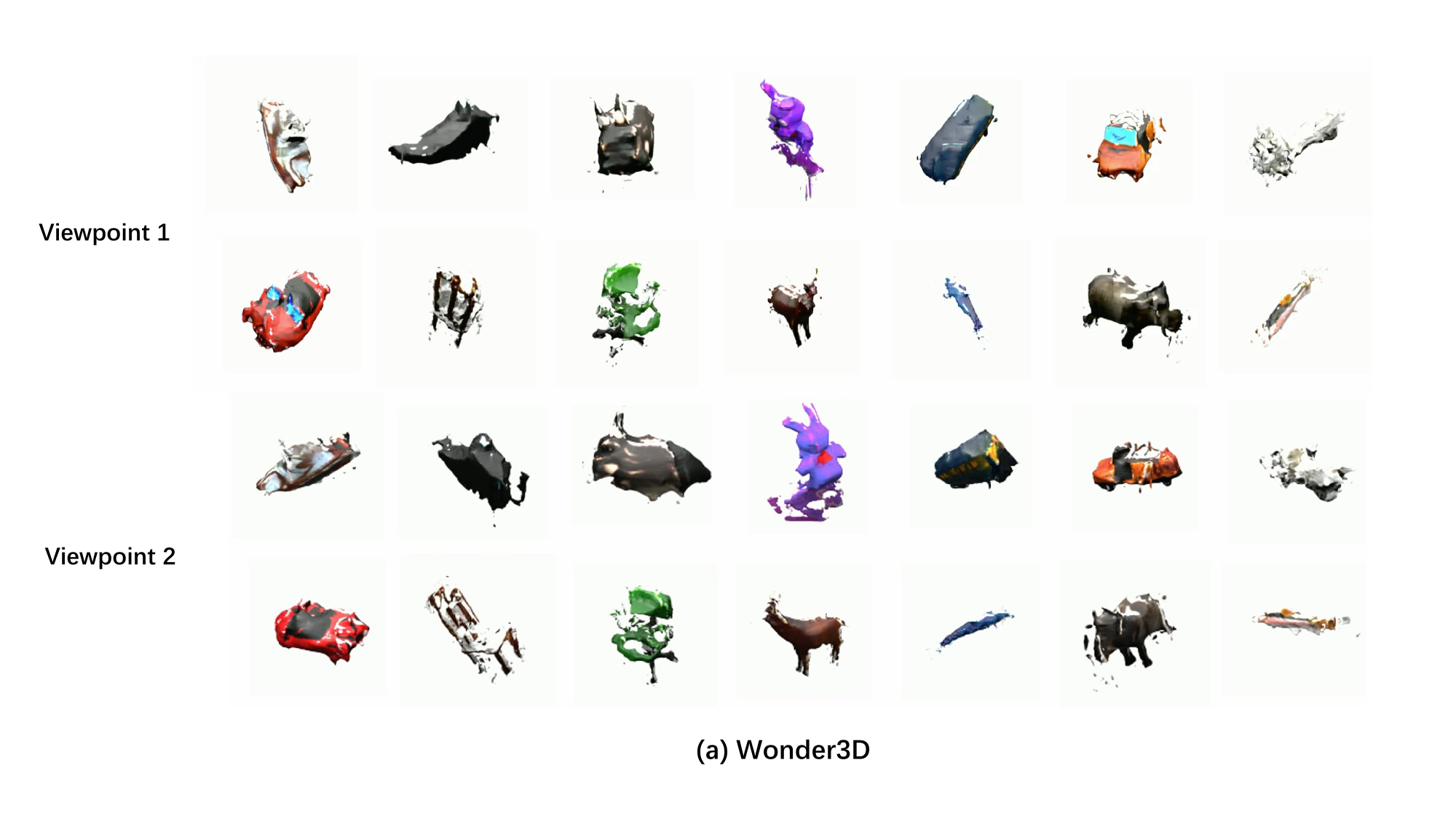}
    \includegraphics[width=0.99\linewidth]{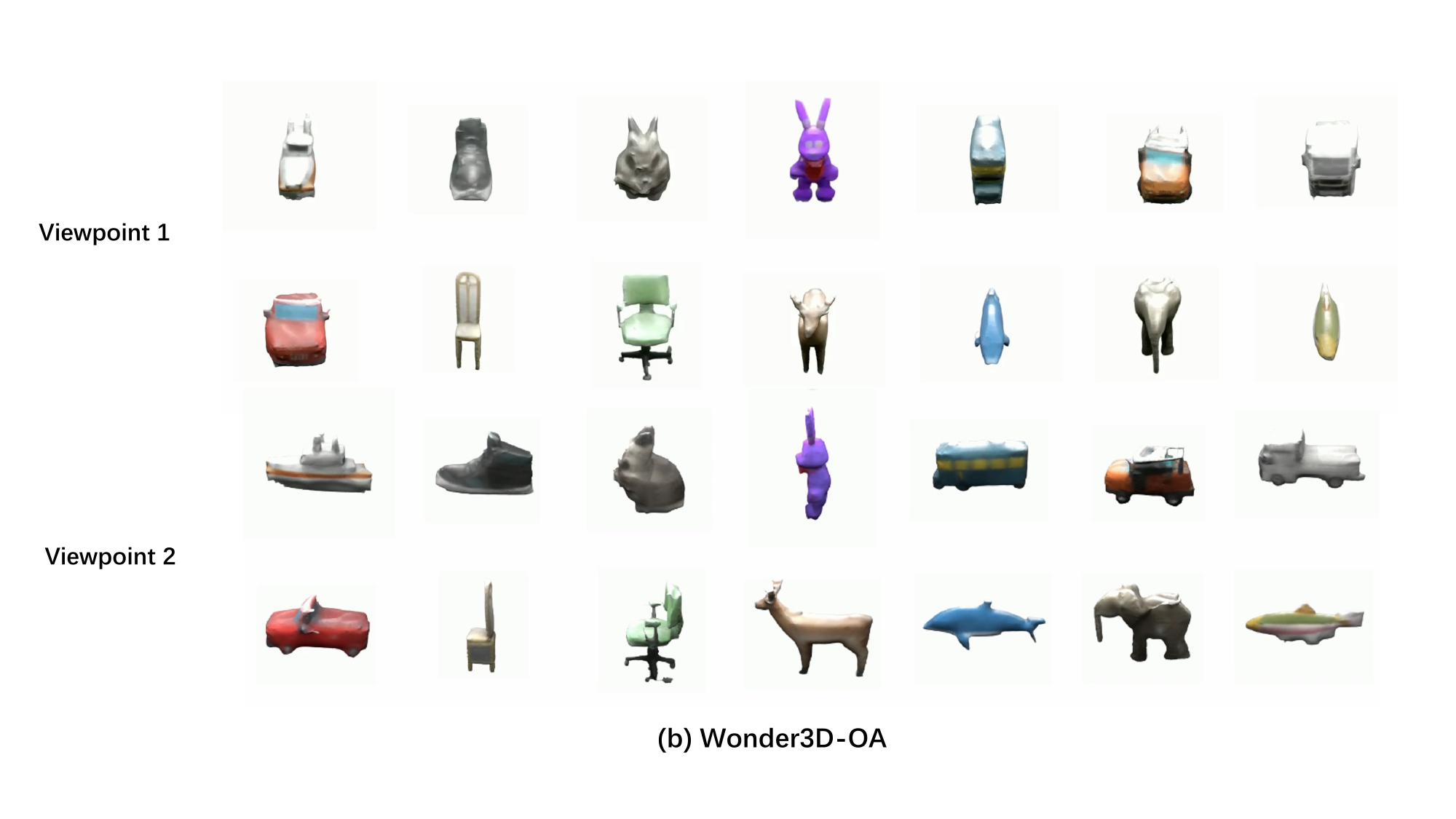}
    \caption{More qualitative comparison between (a) Wonder3D~\cite{Long2023Wonder3DSI} and (b) our Wonder3D-OA.}
    \label{fig:more_wonder3d_1}
\end{figure}

\begin{figure}
    \centering
    \includegraphics[width=0.99\linewidth]{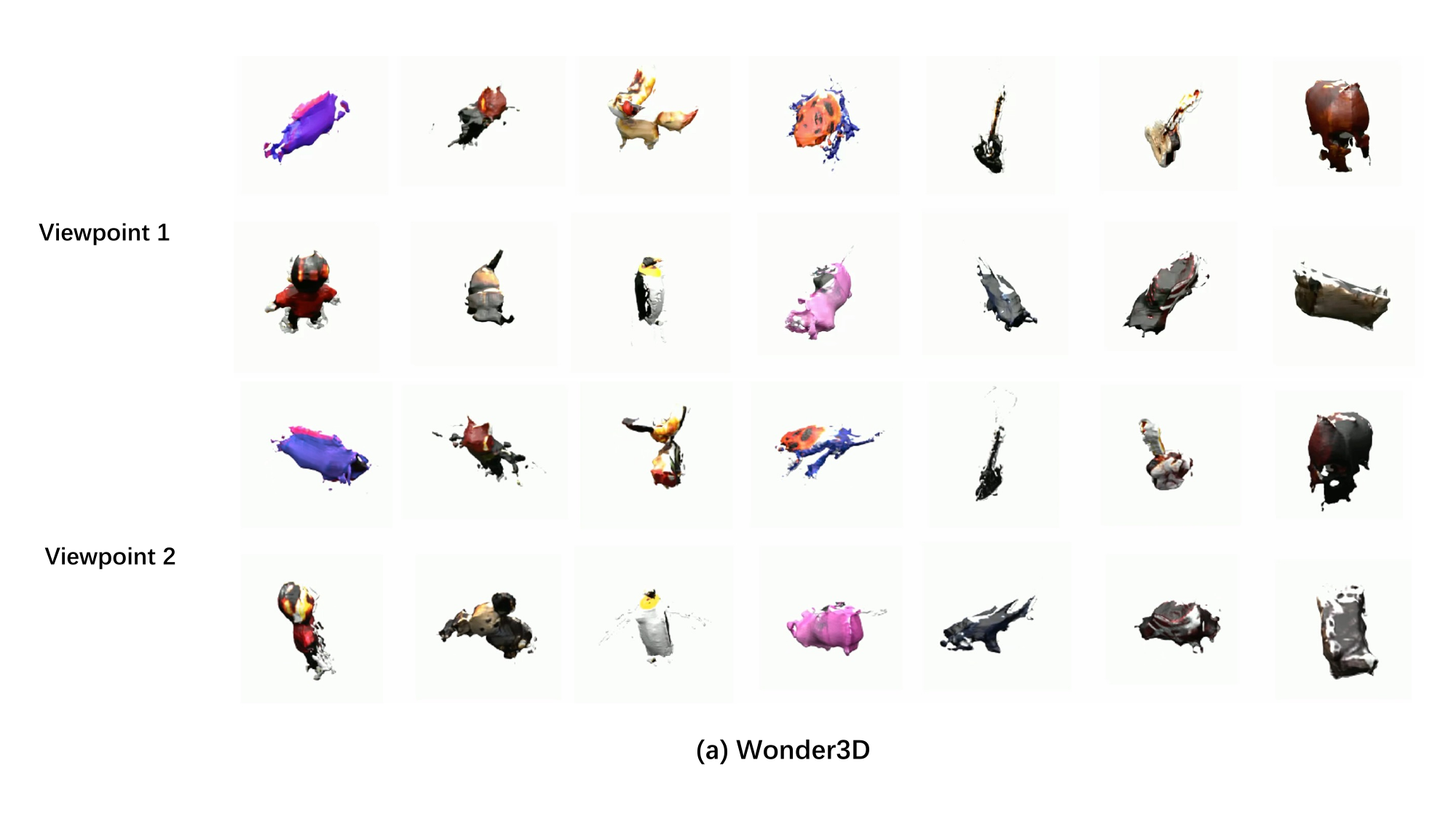}
    \includegraphics[width=0.99\linewidth]{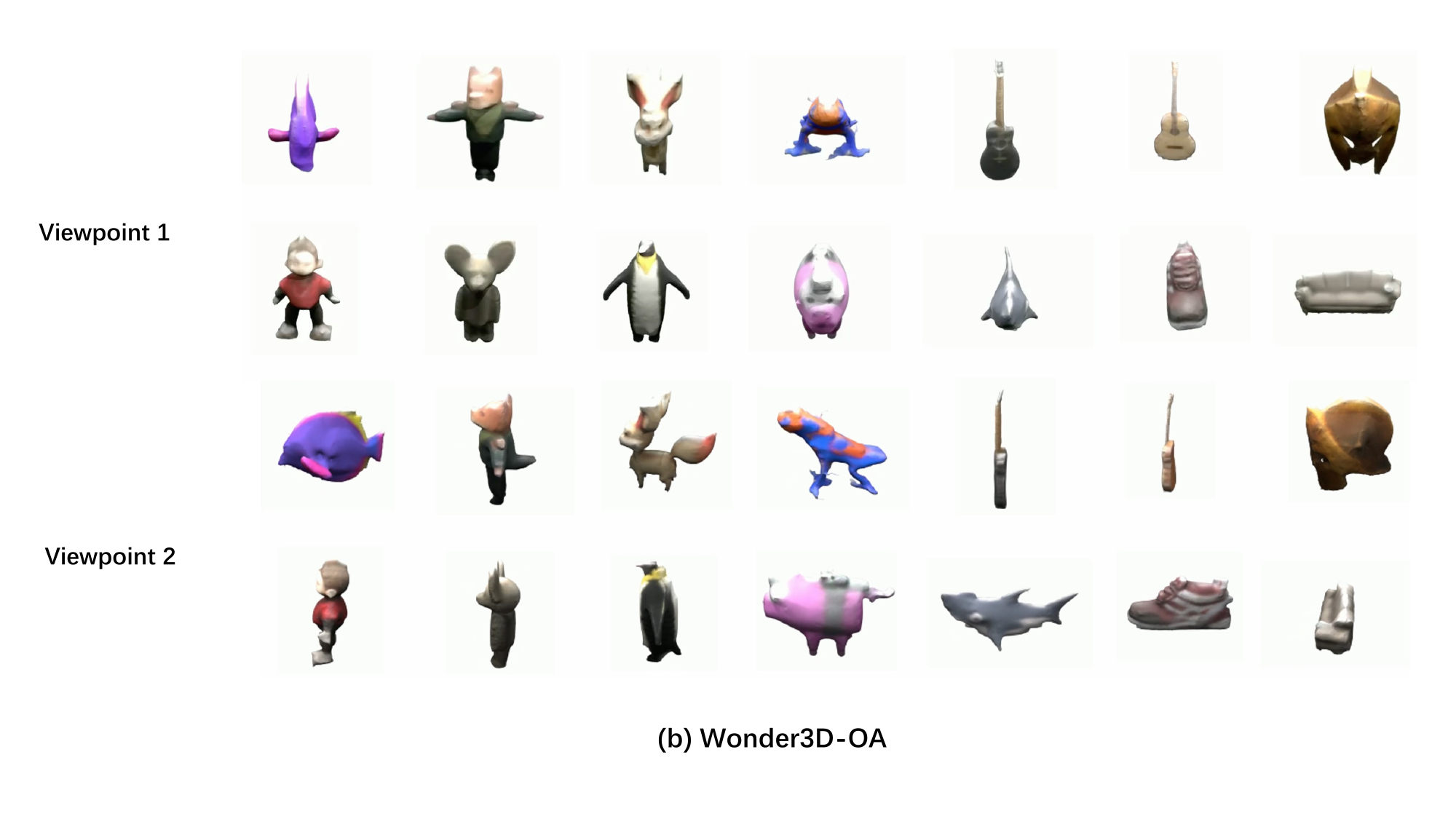}
    \caption{More qualitative comparison between (a) Wonder3D~\cite{Long2023Wonder3DSI} and (b) our Wonder3D-OA.}
    \label{fig:more_wonder3d_2}
\end{figure}

\subsection{Orientation-aligned Object Generation}

We present more qualitative results of our Trellis-OA and orientation estimation method in \figref{fig:more_results_1}, \figref{fig:more_results_2}, \figref{fig:more_results_3}, \figref{fig:more_results_4}, and \figref{fig:more_results_5}. We also present more qualitative comparisons between Wonder3D and our Wonder3D-OA in \figref{fig:more_wonder3d_1} and \figref{fig:more_wonder3d_2}. Besides, we present more results of Trellis-OA and Wonder3D-OA in our demo video. Please check the video for details.

\newcolumntype{C}[1]{>{\centering\arraybackslash}p{#1}}

\begin{table}[]
    \centering
    \resizebox{.9\textwidth}{!} 
    {
    \begin{tabular}{C{4cm}|C{3cm}|C{2.5cm}|C{2.5cm}}
    \toprule
         & Chamfer Distance $\downarrow$ & LPIPS $\downarrow$ & CLIP $\uparrow$ \\
    \midrule
    Trellis (Manually corrected) & 0.0377 & 0.1723 & 88.33 \\
    \midrule
    Trellis-OA (Ours) & \textbf{0.0280} & \textbf{0.1574} & \textbf{88.77} \\
    \bottomrule
    \end{tabular}
    }
    \vspace{1em}
    \caption{Quantitative comparison of geometry and appearance quality between Trellis with manually corrected orientations and Trellis-OA (Ours). The results demonstrate that our orientation-alignment fine-tuning can enhance the performance of the pre-trained checkpoint, likely attributable to the high quality of our Objaverse-OA dataset.}
    \label{tab:more_quality}
\end{table}

To investigate whether our orientation alignment fine-tuning affects the performance of the original pre-trained checkpoint, we randomly select 46 objects across 46 categories in the Toys4k~\cite{Stojanov2021UsingST} dataset. As shown in \tabref{tab:more_quality}, our method not only maintains performance but may even enhance the quality of the pre-trained 3D generative models. This improvement is likely due to the high fidelity of our manually corrected Objaverse-OA dataset. 

\subsection{Orientation Estimation}

\begin{table}[htbp]
\centering
\resizebox{.95\textwidth}{!} %
{
\begin{tabular}{l|cccccc} %
\toprule
& airplane & bicycle & boat & bunny & bus & car \\
\midrule
FSDetView~\cite{Xiao2020FewShotOD} & - / - & 40.00 / 58.52 & 11.11 / 114.89 & - / - & 20.00 / 75.71 & 40.00 / 58.07 \\
\midrule
Orient Anything~\cite{Wang2024OrientAL} (Vit-S) & 30.00 / 70.48 & 20.00 / 63.91 & 44.44 / 73.69 & 33.33 / 47.79 & 100.00 / 11.64 & 80.00 / 29.49 \\
\midrule
Orient Anything~\cite{Wang2024OrientAL} (Vit-L) & 80.00 / 39.03 & 40.00 / 74.73 & 44.44 / 87.91 & 88.89 / 15.63 & 80.00 / 20.32 & 80.00 / 18.73 \\
\midrule
Ours (ViT-S) & 66.67 / 55.99 & 80.00 / 40.28 & 55.56 / 66.21 & 44.44 / 50.77 & 100.00 / 9.52 & 60.00 / 74.66 \\
\midrule
Ours (ViT-L) & 77.78 / 34.96 & 80.00 / 18.76 & 44.44 / 58.47 & 44.44 / 45.00 & 100.00 / 10.55 & 100.00 / 17.15 \\
\midrule
& cat & chair & chicken & cow & crab & deer moose \\
\midrule
FSDetView~\cite{Xiao2020FewShotOD} & - / - & 10.00 / 96.30 & - / - & - / - & - / - & - / - \\
\midrule
Orient Anything~\cite{Wang2024OrientAL} (Vit-S) & 55.56 / 35.79 & 55.00 / 39.50 & 25.00 / 59.75 & 33.33 / 61.73 & 10.00 / 75.44 & 50.00 / 29.89 \\
\midrule
Orient Anything~\cite{Wang2024OrientAL} (Vit-L) & 77.78 / 19.83 & 90.00 / 15.53 & 50.00 / 34.62 & 50.00 / 40.90 & 50.00 / 42.89 & 100.00 / 17.54 \\
\midrule
Ours (ViT-S) & 22.22 / 54.68 & 45.00 / 59.01 & 37.50 / 73.32 & 33.33 / 41.42 & 40.00 / 69.11 & 50.00 / 30.04 \\
\midrule
Ours (ViT-L) & 33.33 / 65.48 & 40.00 / 49.99 & 75.00 / 28.66 & 33.33 / 63.09 & 60.00 / 46.70 & 60.00 / 34.98 \\
\midrule
& dinosaur & dog & dolphin & dragon & elephant & fish \\
\midrule
FSDetView~\cite{Xiao2020FewShotOD} & - / - & - / - & - / - & - / - & - / - & - / - \\
\midrule
Orient Anything~\cite{Wang2024OrientAL} (Vit-S) & 50.00 / 32.32 & 22.22 / 53.95 & 30.00 / 86.84 & 20.00 / 69.35 & 50.00 / 45.98 & 20.00 / 59.59 \\
\midrule
Orient Anything~\cite{Wang2024OrientAL} (Vit-L) & 70.00 / 22.11 & 88.89 / 16.71 & 30.00 / 50.37 & 50.00 / 48.06 & 70.00 / 28.76 & 40.00 / 47.36 \\
\midrule
Ours (ViT-S) & 10.00 / 54.34 & 55.56 / 47.04 & 70.00 / 34.27 & 10.00 / 76.13 & 50.00 / 56.19 & 20.00 / 82.77 \\
\midrule
Ours (ViT-L) & 40.00 / 38.01 & 44.44 / 46.88 & 60.00 / 34.52 & 20.00 / 59.41 & 70.00 / 35.56 & 20.00 / 89.96 \\
\midrule
& fox & frog & giraffe & guitar & helicopter & helmet \\
\midrule
FSDetView~\cite{Xiao2020FewShotOD} & - / - & - / - & - / - & 0.00 / 110.48 & - / - & 28.57 / 49.01 \\
\midrule
Orient Anything~\cite{Wang2024OrientAL} (Vit-S) & 77.78 / 31.87 & 50.00 / 48.89 & 0.00 / 51.62 & 30.00 / 67.83 & 0.00 / 101.22 & 14.29 / 53.71 \\
\midrule
Orient Anything~\cite{Wang2024OrientAL} (Vit-L) & 66.67 / 22.15 & 90.00 / 21.53 & 80.00 / 24.69 & 50.00 / 35.85 & 50.00 / 74.92 & 14.29 / 54.06 \\
\midrule
Ours (ViT-S) & 55.56 / 47.99 & 50.00 / 63.22 & 50.00 / 32.03 & 40.00 / 39.72 & 25.00 / 111.61 & 57.14 / 70.63 \\
\midrule
Ours (ViT-L) & 44.44 / 39.19 & 60.00 / 64.88 & 60.00 / 32.85 & 30.00 / 64.50 & 25.00 / 67.30 & 42.86 / 62.73 \\
\midrule
& horse & laptop & lion & lizard & monkey & motorcycle \\
\midrule
FSDetView~\cite{Xiao2020FewShotOD} & - / - & 62.50 / 42.29 & - / - & - / - & - / - & - / - \\
\midrule
Orient Anything~\cite{Wang2024OrientAL} (Vit-S) & 20.00 / 73.80 & 75.00 / 19.83 & 50.00 / 36.90 & 10.00 / 56.94 & 90.00 / 19.91 & 25.00 / 101.14 \\
\midrule
Orient Anything~\cite{Wang2024OrientAL} (Vit-L) & 80.00 / 36.15 & 50.00 / 70.09 & 70.00 / 28.20 & 40.00 / 38.71 & 70.00 / 21.11 & 50.00 / 51.71 \\
\midrule
Ours (ViT-S) & 70.00 / 31.55 & 50.00 / 72.41 & 30.00 / 65.78 & 50.00 / 45.07 & 60.00 / 40.70 & 50.00 / 72.01 \\
\midrule
Ours (ViT-L) & 60.00 / 48.74 & 50.00 / 32.11 & 40.00 / 33.39 & 50.00 / 36.63 & 60.00 / 49.30 & 62.50 / 68.29 \\
\midrule
& mouse & panda & PC mouse & penguin & piano & pig \\
\midrule
FSDetView~\cite{Xiao2020FewShotOD} & 10.00 / 80.55 & - / - & - / - & - / - & 33.33 / 102.13 & - / - \\
\midrule
Orient Anything~\cite{Wang2024OrientAL} (Vit-S) & 20.00 / 43.97 & 57.14 / 30.07 & 0.00 / 116.94 & 50.00 / 49.04 & 50.00 / 64.67 & 30.00 / 50.62 \\
\midrule
Orient Anything~\cite{Wang2024OrientAL} (Vit-L) & 40.00 / 29.71 & 71.43 / 14.63 & 0.00 / 89.58 & 56.25 / 33.90 & 50.00 / 61.28 & 80.00 / 30.68 \\
\midrule
Ours (ViT-S) & 20.00 / 73.00 & 57.14 / 79.84 & 33.33 / 69.42 & 56.25 / 58.98 & 50.00 / 68.89 & 20.00 / 72.01 \\
\midrule
Ours (ViT-L) & 40.00 / 51.85 & 57.14 / 55.52 & 0.00 / 74.86 & 31.25 / 57.58 & 50.00 / 56.13 & 40.00 / 67.66 \\
\midrule
& radio & robot & shark & sheep & shoe & sofa \\
\midrule
FSDetView~\cite{Xiao2020FewShotOD} & - / - & - / - & - / - & - / - & 40.00 / 98.08 & 15.00 / 107.20 \\
\midrule
Orient Anything~\cite{Wang2024OrientAL} (Vit-S) & 50.00 / 66.47 & 62.50 / 63.46 & 20.00 / 50.26 & 75.00 / 34.54 & 50.00 / 62.30 & 75.00 / 33.12 \\
\midrule
Orient Anything~\cite{Wang2024OrientAL} (Vit-L) & 50.00 / 23.66 & 87.50 / 32.48 & 40.00 / 45.30 & 100.00 / 16.06 & 50.00 / 32.59 & 65.00 / 39.84 \\
\midrule
Ours (ViT-S) & 50.00 / 26.10 & 37.50 / 48.90 & 60.00 / 46.41 & 75.00 / 23.73 & 60.00 / 39.92 & 85.00 / 33.35 \\
\midrule
Ours (ViT-L) & 0.00 / 94.76 & 37.50 / 75.79 & 70.00 / 24.69 & 100.00 / 18.32 & 80.00 / 34.58 & 85.00 / 33.35 \\
\midrule
& tractor & train & truck & violin & whale & \\
\midrule
FSDetView~\cite{Xiao2020FewShotOD} & - / - & 0.00 / 93.15 & - / - & - / - & - / - & \\
\midrule
Orient Anything~\cite{Wang2024OrientAL} (Vit-S) & 75.00 / 32.67 & 0.00 / 117.58 & 80.00 / 23.95 & 0.00 / 95.34 & 44.44 / 47.62 & \\
\midrule
Orient Anything~\cite{Wang2024OrientAL} (Vit-L) & 75.00 / 37.13 & 50.00 / 70.34 & 100.00 / 11.94 & 30.00 / 45.89 & 44.44 / 39.14 & \\
\midrule
Ours (ViT-S) & 87.50 / 31.52 & 25.00 / 119.20 & 90.00 / 23.30 & 50.00 / 69.62 & 44.44 / 57.50 & \\
\midrule
Ours (ViT-L) & 68.75 / 34.80 & 50.00 / 76.18 & 100.00 / 14.03 & 30.00 / 65.17 & 44.44 / 49.80 & \\
\bottomrule
\end{tabular}
}
\vspace{1em}
\captionof{table}{Category level quantitative results of orientation estimation on Toys4k~\cite{Stojanov2021UsingST} dataset in terms of Acc@30 $\uparrow$ and Abs $\downarrow$.}
\label{tab:orientation_est_toys4k}
\end{table}

\begin{table}[htbp]
\centering
\resizebox{.9\textwidth}{!} 
{
\begin{tabular}{c|c|c|c|c|c}
\toprule
& fork & knife & pen & rifle & scissors \\
\midrule
FSDetView~\cite{Xiao2020FewShotOD} & 9.09 / 90.67 & 4.76 / 81.40 & 0.00 / 81.72 & 54.84 / 40.24 & 0.00 / 93.31 \\
\midrule
Orient Anything~\cite{Wang2024OrientAL} (Vit-S) & 4.55 / 83.23 & 0.00 / 78.44 & 6.67 / 74.79 & 3.23 / 87.56 & 3.03 / 74.11 \\
\midrule
Orient Anything~\cite{Wang2024OrientAL} (Vit-L) & 9.09 / 78.51 & 0.00 / 78.95 & 13.33 / 76.79 & 6.45 / 81.73 & 3.03 / 85.85 \\
\midrule
Ours (ViT-S) & 54.55 / 37.43 & 37.50 / 56.61 & 50.00 / 59.02 & 61.29 / 33.89 & 48.48 / 47.15 \\
\midrule
Ours (ViT-L) & 54.55 / 34.98 & 54.17 / 29.82 & 56.67 / 52.50 & 54.84 / 27.90 & 57.58 / 35.68 \\
\midrule
& screwdriver & spoon &  &  &  \\
\midrule
FSDetView~\cite{Xiao2020FewShotOD} & 0.00 / 96.13 & 0.00 / 103.08 &  &  &  \\
\midrule
Orient Anything~\cite{Wang2024OrientAL} (Vit-S) & 0.00 / 90.53 & 2.63 / 76.98 &  &  &  \\
\midrule
Orient Anything~\cite{Wang2024OrientAL} (Vit-L) & 17.24 / 70.88 & 15.79 / 68.49 &  &  &  \\
\midrule
Ours (ViT-S) & 79.31 / 21.93 & 78.95 / 26.04 &  &  &  \\
\midrule
Ours (ViT-L) & 68.97 / 25.18 & 76.32 / 30.10 &  &  &  \\
\hline
\end{tabular}
}
\vspace{1em}
\captionof{table}{Category level quantitative results of orientation estimation on stick-like objects from Imagenet3D~\cite{Ma2024ImageNet3DTG} dataset in terms of Acc@30 $\uparrow$ and Abs $\downarrow$.}
\label{tab:orientation_est_imagenet3d}
\end{table}

We detail orientation estimation results of baselines and our method at the category level in \tabref{tab:orientation_est_toys4k} and \tabref{tab:orientation_est_imagenet3d}. 

\subsection{Impact of Occlusion}

\begin{figure}[h]
    \centering
    \includegraphics[width=0.95\linewidth]{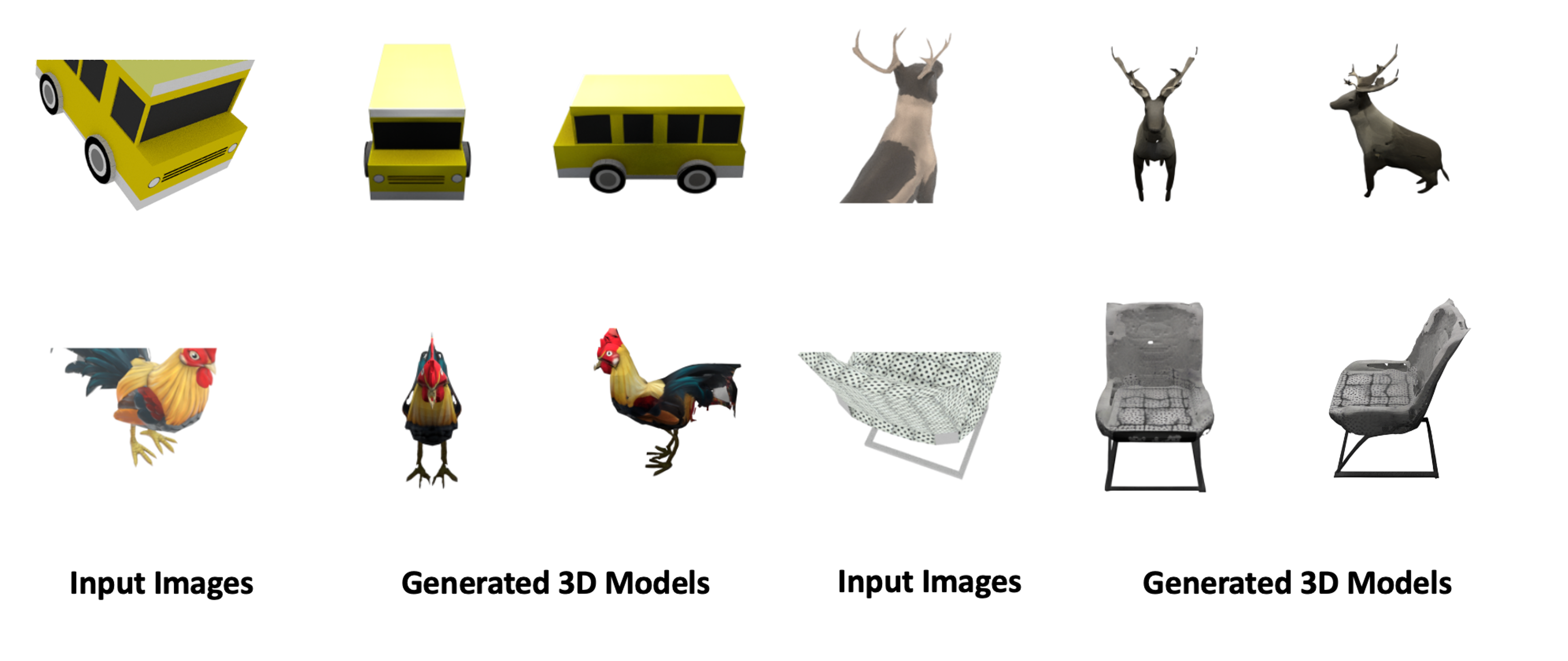}
    \caption{Impact of occlusion}
    \label{fig:occlusion}
\end{figure}

As illustrated in \figref{fig:occlusion}, our object generation module is capable of producing reasonable results under partial occlusion, with minimal performance degradation. However, in scenarios involving severe occlusion, we recommend using the method proposed in~\cite{Wu2025Amodal3RA3}, which is specifically designed for occlusion-robustness, as the pre-trained checkpoint prior to applying our orientation-alignment fine-tuning. It is also important to note that our orientation estimation approach has not been designed to handle occluded inputs. Further research is necessary to extend its applicability to occlusion-prone scenarios.

\subsection{Failure Cases}

\begin{figure}[h]
    \centering
    \includegraphics[width=0.7\linewidth]{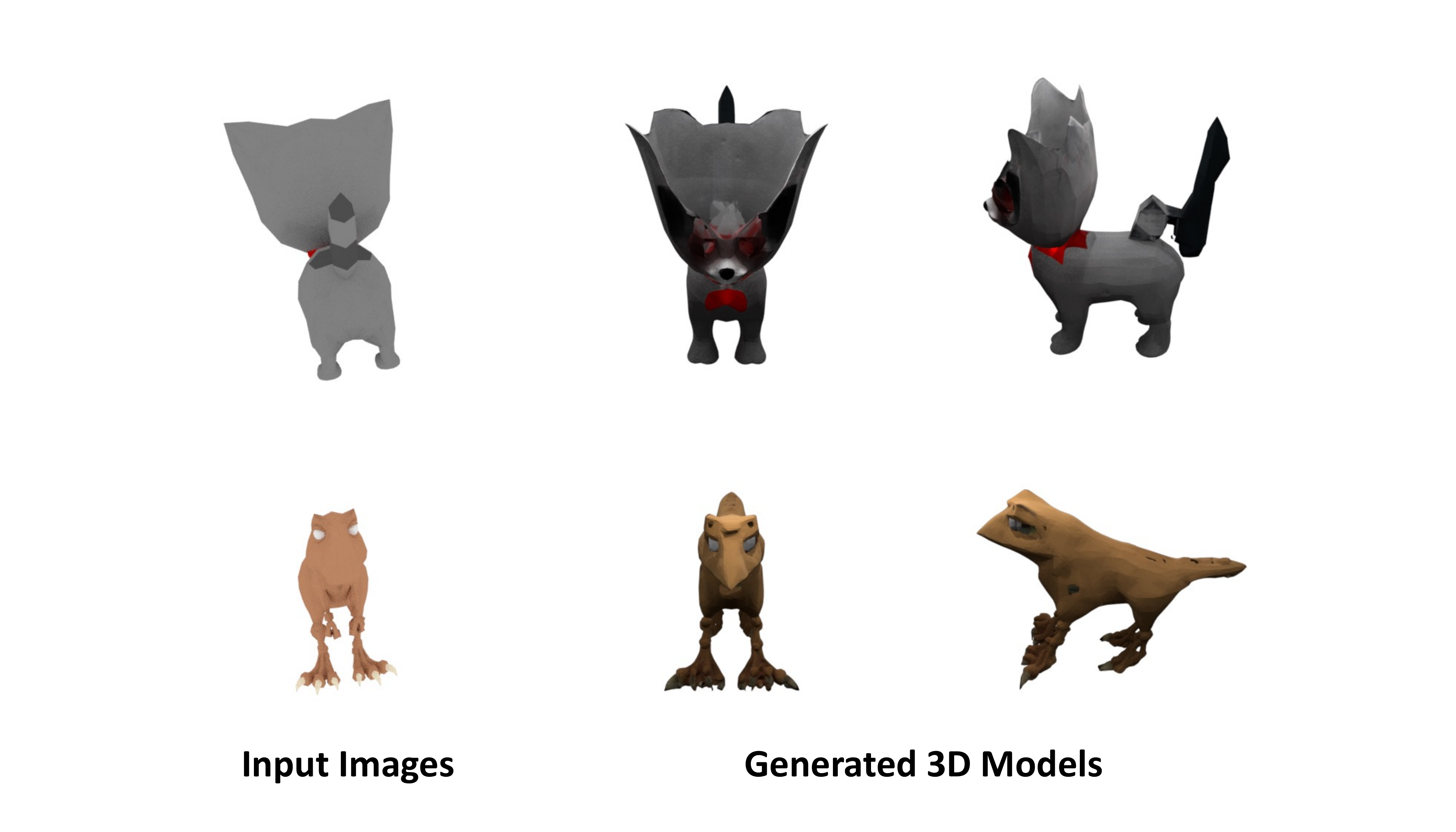}
    \caption{Failure cases.}
    \label{fig:failure_cases}
\end{figure}

Our method may fail in certain scenarios, as illustrated in \figref{fig:failure_cases}. For instance, when the input image is captured from the rear of an object, the synthesized 3D model may exhibit suboptimal results in the front view (first row). Additionally, when applied to unseen categories, the generated 3D model may incorrectly incorporate features from other categories present in the training dataset (second row).

\section{License and Border Impact}

\begin{table}[t]
\centering
\resizebox{.95\textwidth}{!} 
{
\begin{tabular}{ccc}

Assets & License & URL\\
\midrule

Blender 4.2.8 & GNU General Public License (GPL) & text{https://www.blender.org/} \\
Objaverse~\cite{Deitke2022ObjaverseAU} & ODC-By v1.0 license & text{https://objaverse.allenai.org/} \\

\midrule

Trellis~\cite{Xiang2024Structured3L} & MIT License & text{https://github.com/microsoft/TRELLIS} \\
Wonder3D~\cite{Long2023Wonder3DSI} & MIT License & text{https://github.com/xxlong0/Wonder3D/tree/main} \\
FoundationPose~\cite{Wen2023FoundationPoseU6} & NVIDIA Source Code License & text{https://github.com/NVlabs/FoundationPose} \\
LGM~\cite{Tang2024LGMLM} & MIT License & text{https://github.com/3DTopia/LGM} \\

\bottomrule
\end{tabular}
}
\vspace{1em}
\caption{License}
\label{tab:license}
\end{table}

The licensing details are provided in \tabref{tab:license}. Our method is suitable for generating orientation-aligned 3D objects for downstream applications in 3D perception and augmented reality, which may benefit both scientific research and commercial development. However, it is important to acknowledge potential risks: the method could be misused to generate hazardous 3D objects, such as weapons, which may lead to societal concerns.

\end{document}